\documentclass[10pt,journal]{IEEEtran}
\usepackage{cite}
\usepackage{amsmath,amssymb,amsfonts}
\usepackage{mathtools}
\usepackage{graphicx}
\let\labelindent\relax
\usepackage{enumitem}
\usepackage{siunitx}
\usepackage{float}
\usepackage{xcolor,soul}
\usepackage{dsfont}
\usepackage[capitalize,nosort]{cleveref}
\crefformat{equation}{(#2#1#3)}

\usepackage[linesnumbered,ruled,vlined]{algorithm2e}
\SetKwInput{KwInput}{Input}                
\SetKwInput{KwOutput}{Output}              
\SetKwComment{Comment}{$\triangleright$}{}
\SetKwFunction{KwgPCProjection}{gPCProjection}
\SetKwFunction{Kwlinearize}{\textbf{Linearize}}
\SetKwFunction{KwDiscretize}{\textbf{Discretize}}
\SetKwFunction{KwSolve}{\textbf{SCP}}
\SetKwFunction{Kwaorrt}{\textbf{AO-RRT}}

\SetKwBlock{blockgpcprojection}{gPC Projection}{}
\SetKwBlock{blockprojectback}{Compute_distribution_from_gPC_states}{}

\SetKwBlock{blockpcscp}{gPC-SCP$^\mathrm{PC}$}{}
\SetKwBlock{blockcovariance}{Covariance Prediction}{}
\SetKwBlock{block_correction_optimization}{Correction Program}{}

\usepackage{amsthm} 
\newtheorem{theorem}{Theorem}
\newtheorem{lemma}{Lemma}
\newtheorem{prop}{Proposition}
\newtheorem{defn}{Definition}
\newtheorem{rmrk}{Remark}
\newtheorem{asmp}{Assumption}
\newtheorem{problem}{Problem}
\newtheorem{corollary}{Corollary}
\newtheorem{example}{Example}

\Crefname{lemma}{Lemma}{Lemmas}
\Crefname{problem}{Problem}{Problem}
\Crefname{theorem}{Theorem}{Theorems}
\Crefname{defn}{Definition}{Definition}
\Crefname{rmrk}{Remark}{Remark}
\Crefname{asmp}{Assumption}{Assumptions}
\Crefname{prop}{Proposition}{Proposition}
\Crefname{corollary}{Corollary}{Corollary}
\crefname{example}{Example}{Example}
\crefname{equation}{Eq.\!}{Eqs.\!}
\crefname{figure}{Fig.\!}{Figs.\!}

\definecolor{babyblueeyes}{rgb}{0.63, 0.79, 0.95}
\sethlcolor{babyblueeyes}

\newcommand{\setX}{\mathcal{X}}
\newcommand{\setU}{\mathcal{U}}

\newcommand{\setN}{\mathcal{N}}
\newcommand{\Fnorm}{\mathrm{F}}
\newcommand{\setF}{\mathcal{F}}

\newcommand{\real}{\mathbb{R}}

\newcommand{\gpcX}{\mathrm{X}}

\newcommand{\baru}{\bar{u}}
\newcommand{\barp}{\bar{p}}

\newcommand{\vecx}{\mathbf{x}}
\newcommand{\vecv}{\mathbf{v}}
\newcommand{\vecw}{\boldsymbol\omega}
\newcommand{\vecu}{\mathbf{u}}
\newcommand{\vect}{\boldsymbol\tau}

\newcommand{\setI}{I}

\newcommand{\vecq}{\mathbf{q}}
\newcommand{\vecp}{\mathbf{p}}

\newcommand{\identity}{\mathrm{I}}

\newcommand{\matC}{\mathbf{C}}

\newcommand{\gpc}{\mathrm{gPC}}
\newcommand{\obs}{\mathrm{obs}}
\newcommand{\rob}{\mathrm{rob}}
\newcommand{\col}{\mathrm{col}}
\newcommand{\nom}{\mathrm{nom}}

\newcommand{\safe}{\mathrm{safe}}
\newcommand{\free}{\mathrm{free}}

\newcommand{\sol}{\mathrm{sol}}

\newcommand{\Prob}{\mathrm{Pr}}
\newcommand{\Exp}{\mathbb{E}}
\newcommand{\Trace}{\mathrm{tr}}

\newcommand{\changed}[1]{\textcolor{black}{#1}}

\title{\huge Trajectory Optimization of Chance-Constrained Nonlinear Stochastic Systems for Motion Planning Under Uncertainty}

\author{Yashwanth Kumar Nakka, \IEEEmembership{Member, IEEE}, and Soon-Jo Chung, \IEEEmembership{Senior Member, IEEE} 
\thanks{This work supported in part by the Jet Propulsion Laboratory.}
\thanks{Yashwanth Kumar Nakka and Soon-Jo Chung are with the Division of Engineering and Applied Science of the California Institute of Technology.{{Email:}\{ynakka@caltech.edu, sjchung@caltech.edu\}}
}}

\begin{document}
\maketitle
\begin{abstract} 
 We present gPC-SCP: Generalized Polynomial Chaos-based Sequential Convex Programming to compute a sub-optimal solution for a continuous-time chance-constrained stochastic nonlinear optimal control (SNOC) problem. The approach enables motion planning for robotic systems under uncertainty. The gPC-SCP method involves two steps. The first step is to derive a surrogate problem of \emph{deterministic} nonlinear optimal control (DNOC) with convex constraints by using gPC expansion and the distributionally-robust convex subset of the chance constraints. The second step is to solve the DNOC problem using sequential convex programming for trajectory generation and control. We prove that in the unconstrained case, the optimal value of the DNOC converges to that of SNOC asymptotically and that any feasible solution of the constrained DNOC is a feasible solution of the chance-constrained SNOC. \changed{We also present the predictor-corrector extension (gPC-SCP$^\mathrm{PC}$) for real-time motion trajectory generation in the presence of stochastic uncertainty. In the gPC-SCP$^\mathrm{PC}$ method, we first predict the uncertainty using the gPC method and then optimize the motion plan to accommodate the uncertainty. We empirically demonstrate the efficacy of the gPC-SCP and the gPC-SCP$^\mathrm{PC}$ methods for the following two test cases: 1) collision checking under uncertainty in actuation and physical parameters and 2) collision checking with stochastic obstacle model for 3DOF and 6DOF robotic systems. We validate the effectiveness of the gPC-SCP method on the 3DOF robotic spacecraft testbed.}
\end{abstract}

\section{Introduction}
\IEEEPARstart{C}oNFIDENCE-based motion planning~\cite{nakka2019nsoc,blackmore2010probabilistic,blackmore2011chance,toit2012robot} and control algorithms~\cite{tsukamoto2020,zhu2019chance}, that incorporate uncertainties in the dynamic model and environment to guarantee safety and performance with high probability, enable safe operation of robots and autonomous systems in partially-known and dynamic environments. A probabilistic approach can allow for integration with a higher-level discrete decision-making algorithm for information gathering~\cite{kaelbling2013integrated,nakka2021information}, and for safe exploration~\cite{nakka2020chance,wabersich2021,Nakka2021SpacecraftLearning} to learn the interaction with an unknown environment. Examples of autonomous systems that require safety guarantees under uncertainty include spacecraft with thrusters as actuators during proximity operations~\cite{nakka2018six,nakka2021information,Nakka2021SpacecraftLearning}, powered descent on Mars~\cite{ridderhof2018uncertainty}, and quadrotors flying in turbulent winds~\cite{shi2018neural,shi2020neural}.
\begin{figure}
    \centering
    \includegraphics[width=0.9\columnwidth]{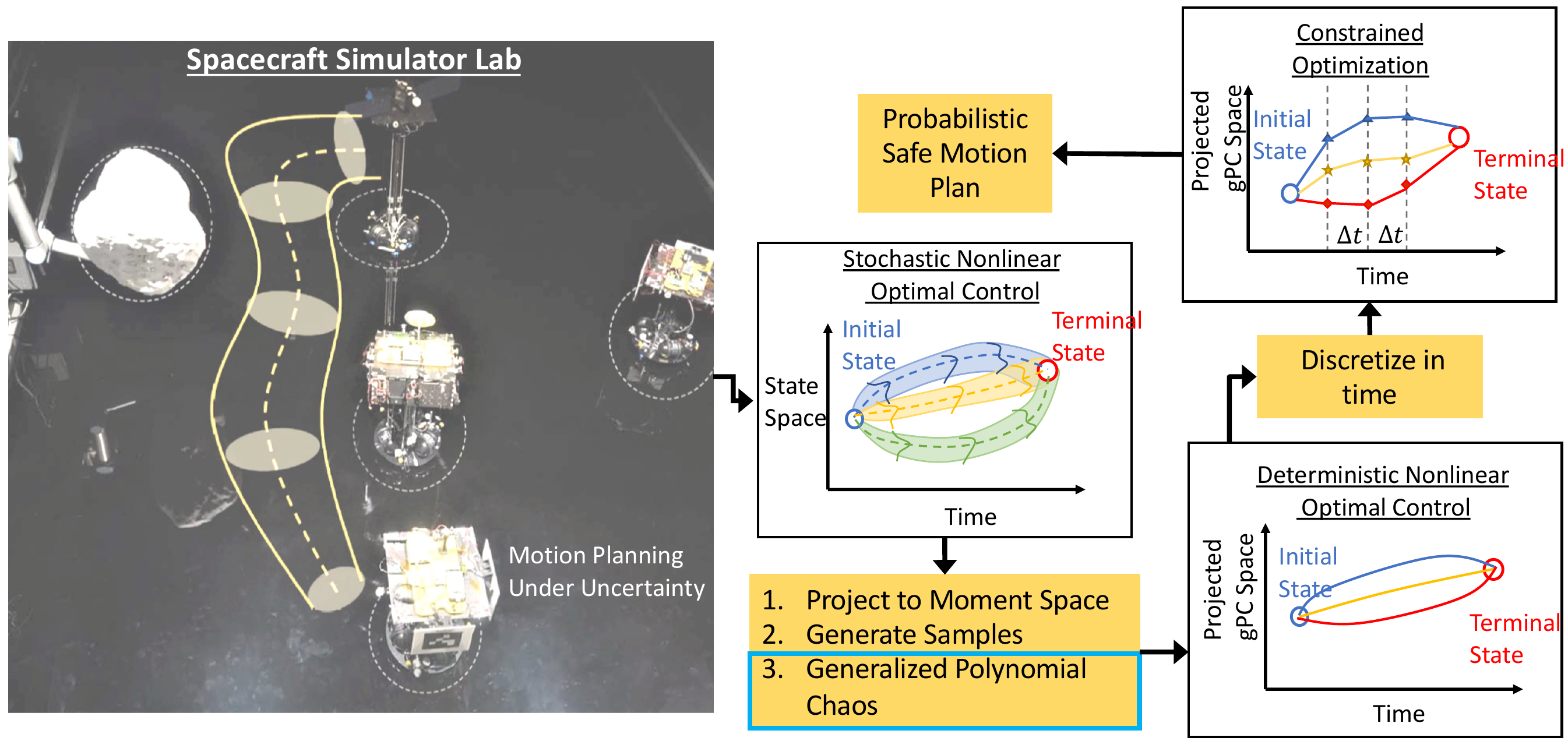}
    \caption{\changed{Caltech's M-STAR~\cite{nakka2018six} (Multi-Spacecraft Testbed for Autonomy Research) planning a safe trajectory to ensure safety under uncertainty in actuation during a proximity maneuver. The motion planning problem is formulated as a chance-constrained stochastic optimal control problem solved in two steps. Step 1: Project the stochastic problem to a deterministic problem by using generalized polynomial chaos approach and distributional robustness (the different colors correspond to trajectories with different values of the cost function); and Step 2: Use deterministic solvers to compute an optimal solution to the deterministic problem.}}\label{fig:space_lab_experiment}
\end{figure}

A motion planning problem considering safety in conjunction with optimality under uncertainty can be formulated as a continuous-time continuous-space stochastic nonlinear optimal control problem (SNOC) with chance constraints. In this paper, we present the generalized polynomial chaos-based sequential convex programming (gPC-SCP) method, as depicted in~\cref{fig:space_lab_experiment}, to solve a chance-constrained SNOC problem. The method involves deriving a deterministic nonlinear optimal control (DNOC) problem with convex constraints that are a surrogate to the SNOC problem with linear and quadratic chance constraints. We derive the DNOC problem by accounting for nonlinear stochastic dynamics using generalized polynomial chaos expansions (gPC)~\cite{xiu2002wiener,xiu2009fast,ghanem2003stochastic} and obtaining deterministic convex approximations of linear and quadratic chance constraints using distributional robustness~\cite{nemirovski2007convex,calafiore2006distributionally,zymler2013distributionally}. \changed{The DNOC problem is then solved using sequential convex programming (SCP)~\cite{morgan2014model,morgan2012spacecraft,morgan2016swarm} for trajectory optimization. The gPC-SCP method computes the full probability distribution of the state as a function of time. In order to compute only the mean of the state as a function of time, we derive a predictor-corrector gPC-SCP (gPC-SCP$^\mathrm{PC}$) method, where we decouple the prediction step (uncertainty propagation) and the optimization step (trajectory correction), thereby deriving an iterative algorithm for fast real-time planning.}

The main contributions of the paper are as follows: 
\begin{enumerate}[label=(\alph*)]
\item We present a systematic sequence of approximations for the chance-constrained SNOC problem to compute a convex-constrained DNOC problem using gPC projection. We analyze the gPC projection of the stochastic dynamics for existence and uniqueness~\cite{kushner1967stochastic,khasminskii2011stochastic} of a solution in the gPC space. We provide examples to study the effect of the projection on both the controllability of surrogate dynamics and the feasibility of the DNOC problem. We prove the convexity of the distributionally-robust linear and quadratic chance constraints in the gPC space.
\item To characterize the deterministic approximation obtained using gPC projection, we present a rigorous analysis on the convergence of the DNOC problem to the SNOC problem for the unconstrained case. Then, we prove that any feasible solution to the constrained DNOC problem is feasible for the chance-constrained SNOC problem with an appropriate gPC transformation step applied.
\item We derive convex surrogates for collision checking with deterministic and stochastic obstacle state models. We integrate this collision constraint with a sampling-based planning method~\cite{hauser2016,lavalle_2006} to derive an algorithm that computes safe and optimal motion plans under uncertainty.
\item \changed{We derive a predictor-corrector algorithm using the gPC-SCP method to compute a mean state trajectory that is safe and optimal under uncertainty. We compare gPC-SCP$^\mathrm{PC}$ with gPC-SCP and in terms of problem complexity, computation cost, and the ability to perform covariance control.}
\item We validate the convergence and the safety provided by the convex constraints in simulation on a three-degree-of-freedom robot dynamics and six-degree-of-freedom spacecraft simulator robot. We show empirically that the gPC-SCP and the gPC-SCP$^\mathrm{PC}$ methods compute a safer trajectory, thereby having a higher success rate in comparison to the Gaussian approximation~\cite{zhu2019chance,du2011probabilistic} of the collision chance constraints. We demonstrate the efficacy of the gPC-SCP and the gPC-SCP$^\mathrm{PC}$ methods by computing a safe trajectory for a spacecraft proximity maneuver under uncertainty in the environment (obstacles) on the robotic spacecraft dynamics simulator hardware~\cite{nakka2018six} and by executing the trajectory in real-time closed-loop experiments. 
\end{enumerate}

\subsection{Related Work} 
Existing methods to solve a chance-constrained stochastic optimal control problem use moment space propagation~\cite{todorov2005generalized,van2012motion,ridderhof2019nonlinear,zhu2019chance}, unscented transformation-based propagation~\cite{calafiore2013stochastic}, Monte Carlo sample propagation~\cite{janson2018monte,blackmore2010probabilistic,blackmore2011chance}, and scenario-based~\cite{calafiore2013,calafiore2013stochastic} approaches to construct a deterministic surrogate problem. Although these methods alleviate the curse of dimensionality, they do not provide asymptotic convergence guarantees for a DNOC problem. Monte Carlo methods provide asymptotic convergence guarantees but often require large samples to estimate the constraint satisfaction for nonlinear systems and use mixed-integer programming~\cite{blackmore2010probabilistic} solvers for computing a solution. We use gPC propagation~\cite{ghanem2003stochastic} to construct a DNOC problem that converges to the SNOC problem asymptotically. The gPC projection transforms the chance constraints from being a non-convex constraint in moment space to a convex constraint in the gPC space. This enables the use of sequential convex programming~\cite{morgan2014model,morgan2016swarm} method for computing a solution. Additionally, we study the existence and uniqueness~\cite{arnold1974stochastic} of a solution and the controllability of the deterministic surrogate dynamics of the stochastic dynamics.

Earlier work~\cite{castillo2020real,zhu2019chance,blackmore2011chance} uses a Gaussian approximation of the linear and the quadratic chance constraint for collision checking and for terminal constraint satisfaction. While this avoids multi-dimensional integration of chance constraints for feasibility checking, Gaussian approximation might not be an equivalent representation (or) even a subset of the feasible set in the presence of stochastic process noise in dynamics. We use distributional robustness~\cite{calafiore2006distributionally,zymler2013distributionally} property to propose a new deterministic second-order cone constraint and a quadratic constraint approximation of the linear and quadratic chance constraints. We prove that the deterministic approximations are a subset of the respective chance constraints.

In~\cite{blackmore2010probabilistic,blackmore2011chance}, linear chance constraints were considered for probabilistic optimal planning for linear systems. The literature on chance-constrained programming focuses on problems with deterministic decision variables and uncertain system parameters for both linear~\cite{calafiore2006distributionally} and nonlinear~\cite{zymler2013distributionally} cases. The results~\cite{calafiore2006distributionally,nemirovski2007convex} on distributional robust subset and convex approximations of the chance constraints can be readily transformed to the case with a random decision variable for an unknown measure. The quadratic chance constraint would lead to an inner semi-definite program~\cite{vandenberghe2007generalized} that adds complexity to the SNOC problem considered in this paper. The linear chance constraint for collision checking was first presented in~\cite{nakka2019nsoc}. In~\cite{tlew2020}, authors show that linearized chance constraint is a subset of the original nonlinear chance constraint for a Gaussian confidence-based constraint. Since the local Gaussian assumption might not be valid for nonlinear systems, we present proof for the distributionally robust convex constraint formulation that includes uncertainty in obstacle state for a nonlinear stochastic differential equation.

The gPC expansion approach was used for stability analysis and control design of uncertain systems~\cite{mesbah2016stochastic,mesbah2014stochastic,bavdekar2016stochastic,hover2006application,fisher2008stability,kim2013wiener}. For trajectory optimization, recent work focused on nonlinear systems with parametric uncertainty~\cite{boutselis2017stochastic,fisher2011optimal} with no constraints on the state or linear systems with linear chance-constraints that do not extend to the SNOC problem considered here and lack analysis on the deterministic approximation of the uncertain system. The gPC approach was used to compute a moment-space receding horizon approximation~\cite{buehler2017efficient}, which was solved using nonlinear programming methods. We extend prior work to incorporate nonlinear dynamics and include analysis on the deterministic approximation. We formulate convex constraints for linear and quadratic constraints in gPC space and use this formulation to design algorithms for motion planning and control of a nonlinear stochastic dynamic system. 

\changed{Furthermore, we present a new predictor-corrector formulation (gPC-SCP$^\mathrm{PC}$). The gPC-SCP$^\mathrm{PC}$ method decouples the stochastic propagation using gPC and planning using SCP to optimize a nominal trajectory and ensure safety under uncertainty. In~\cite{foust2020optimal}, a similar approach was used to correct for the linearization and discretization errors via nonlinear propagation in the SCP. To the best of our knowledge, no other work used such an approach for planning under uncertainty for nonlinear stochastic systems with chance constraints.}

We generalize and extend our prior conference paper~\cite{nakka2019nsoc} significantly as follows: i) we derive a sequence of approximations from a SNOC problem to the DNOC problem, which provides a modular architecture to understand trajectory optimization under uncertainty; ii) we include examples discussing the effect of gPC projection on the controllability of stochastic dynamics; iii) we design a motion planning algorithm to handle uncertainty in both dynamics and obstacle location, which takes advantage of the state-of-the-art sampling-based~\cite{hauser2016} planning algorithms; iv) \changed{we formulate a predictor-corrector extension of gPC-SCP and provide conditions for real-time trajectory generation; and v) we validate the gPC-SCP and the gPC-SCP$^\mathrm{PC}$ approach empirically on three and six-degree-of-freedom robot dynamics and the spacecraft simulator hardware testbed.}

\subsubsection*{Organization} We discuss the stochastic nonlinear optimal control (SNOC) problem with results on deterministic approximations of chance constraints along with preliminaries on gPC expansions in~\cref{sec:nsoc_prelim}. In~\cref{sec:det_opt_prob}, we present the deterministic surrogate of the SNOC problem in terms of the gPC coefficients and the SCP formulation of the DNOC problem. In~\cref{sec:motion_planning_gpc_scp}, we apply the gPC-SCP method under uncertainty in dynamics and constraints for motion planning using SNOC solutions. Furthermore, we derive the gPC-SCP$^\mathrm{PC}$ method and compare with the original gPC-SCP method. In~\cref{sec:simulations_and_experiments}, we validate the gPC-SCP method via simulations on a 3DOF and 6DOF spacecraft simulator robot and via experiments on the robotic spacecraft simulator testbed. Finally, we conclude the paper in~\cref{sec:conclusion} with a brief discussion on the approach and impact of the method.

\subsubsection*{Notation} 
For a random variable $x \in \real^{d_x}$, $\mu_x$ is the mean, $\Sigma_x$ is the covariance matrix, $\real$ is real line, $d_{x}$ is the dimension of $x$. $\Exp$ and $\Prob$ are the expectation operation and probability measure, respectively. We define a deterministic vector as $\bar{x}$. The $p-$norm of a vector $\baru \in \real^{d_u}$ is defined as $\|\baru\|_{p} = (\sum_{1}^{d_{u}}|\baru_i|^p)^{\frac{1}{p}}$. The risk measure for constraint violation is $\epsilon$. We define the gPC state using $\gpcX$ and the indicator function as $I$. We use $\mathbb{I}$ for an identity matrix and $\mathds{1}$ for a matrix with entries as $1$. The Kronecker's product of two matrices $A_{m \times n}$ and $B_{p \times q}$ is defined as follows:
\begin{equation}
 (A \otimes B)_{mp \times nq} = \begin{bmatrix} a_{11}B & \dots & a_{1n}B\\ \vdots & \ddots & \vdots \\ a_{m1}B & \dots & a_{mn}B\end{bmatrix}, \nonumber 
\end{equation}
where $a_{ij}$ is the element at $i^{\mathrm{th}}$ row and $j^{\mathrm{th}}$ column of A. For a matrix $A$, $\Trace (A)$ is the trace operation, $\lambda_{\min}(A)$ and $\lambda_{\max}(A)$ are the minimum and maximum eigenvalues of $A$. 

\section{Problem and Preliminaries} \label{sec:nsoc_prelim}
In this section, we present the stochastic optimal control problem formulation, preliminaries on the relaxations used for chance constraints, and the generalized polynomial chaos approach that forms a basis for constructing a surrogate deterministic optimal control problem.
\subsection{Stochastic Nonlinear Optimal Control Problem}\label{subsec:stoc_opt_prob}
We consider the finite-horizon stochastic nonlinear optimal control (SNOC) problem with joint chance constraints in continuous time and continuous space. The SNOC problem minimizes an expectation cost function, which is the sum of a quadratic function in the random state variable $x(t)$ and a convex norm of the control policy $\Bar{u}(t)$. The evolution of the stochastic process $x(t)$ for all sampled paths is defined by a stochastic differential equation. The joint chance constraints guarantee constraint feasibility with a probability of $1- \epsilon$, where $\epsilon >0$ and is chosen to be a small value (e.g. $\epsilon \in [0.001,0.05]$) for better constraint satisfaction. The following optimal control problem is considered with the state distribution and control as the decision variables. 
\begin{problem} 
\label{prob:cc_stoc_opt_cntrl} 
Chance-Constrained Stochastic Nonlinear Optimal Control. 
\begin{align}
J_{\mathrm{SNOC}}^{*} = & \underset{x(t),\Bar{u}(t)}{\min}
\scalebox{0.9}{$\Exp \left[\int_{t_{0}}^{t_{f}}J(x(t),\Bar{u}(t))dt + J_{f}(x(t_{f}))\right]$}\\
\text{s.t.} \quad
& \scalebox{0.9}{$dx = f(x(t),\Bar{u}(t))dt + g(x(t),\Bar{u}(t)) dw(t)$}\\
&\scalebox{0.9}{$\Prob (x(t) \in \setX_{\setF}) \geq 1-\epsilon \ \forall t \in [t_{0},t_{f}]$}\\
&\scalebox{0.9}{$\Bar{u}(t) \in \setU \quad \forall t \in [t_{0},t_{f}]$} \label{eq:control_limits}\\
&\scalebox{0.9}{$x(t_{0}) = x_{0} \quad x_{t_{f}} \in \setX_{f}$} \label{eq:init_term_conditions}
\end{align} 
\end{problem}
The cost functional $J$ and the terminal cost $J_{f}$ are:
\begin{align}
  J(x(t),\Bar{u}(t))&= x(t)^\top Qx(t)  + \|\bar{u}\|_{p},\ \text{where} \ p \in \{1,2,\infty\},\nonumber\\
    J_{f}(x(t_{f}))&=x(t_f)^\top Q_{f}x(t_f).
\label{eq:cost_stopt}
\end{align}
where $Q$ and $Q_{f}$ are positive definite matrices. The $p-$norm of a vector $\baru$ is defined as $\| \baru\|_{p} = (\sum_{1}^{d_{u}}|\baru_i|^p)^{\frac{1}{p}}$. The terminal set $\setX_{f}$ is the set of allowed realization of the state $x$ after propagation. We apply the terminal constraint with slackness in the terminal variance to ensure feasibility of~\cref{prob:cc_stoc_opt_cntrl}. 
In the following, we define each of the aforementioned elements of \cref{prob:cc_stoc_opt_cntrl} and discuss convex approximations of linear and quadratic chance constraints. 
\subsubsection{Stochastic Differential Equation (SDE)~\cite{arnold1974stochastic}}
The dynamics of the system is modeled as a controlled diffusion process with $\mathrm{It\hat{o}}$ assumptions. The random variable $x(t)$ is defined on a probability space $(\Omega,\mathbb{F},\Prob)$ where $\Omega$ is the sample space, $\mathbb{F}$ forms a $\sigma$-field with measure $\Prob$.
\begin{equation}
\begin{aligned}
    & dx(t) = f(x(t),\Bar{u}(t))dt + g(x(t),\Bar{u}(t))dw(t),\\&  
    \Prob(|x(t_{0}) - x_{0}| = 0) = 1, \quad \forall t_{0} \leq t \leq t_{f} < \infty,
\end{aligned}\label{eq:nl_stochastic_dynamics}
\end{equation} 
where: $f(.,.):\setX \times \setU \to \real^{d_{x}}$, $g(.,.): \setX \times \setU \to \real^{d_{x} \times d_{\xi}}$, and $w(t)$ is a $d_{\xi}$-dimensional Wiener process and the initial random variable $x_{0}$ is independent of $w(t)-w(t_{0})$ for $t \geq t_{0}$, and $dw(t) \sim \setN(0,dt\mathbb{I})$. The sets $\setX \subseteq \real^{d_{x}}$ and $\setU \subseteq \real^{d_{u}}$ are compact sets. We make the following assumptions to ensure the existence and uniqueness of a solution to the SDE.
\begin{asmp}The functions $f(x(t),\Bar{u}(t))$ and $g(x(t),\Bar{u}(t))$ are defined and measurable on $\setX \times \setU$. 
\end{asmp} 
\begin{asmp}Equation~\cref{eq:nl_stochastic_dynamics} has a unique solution $x(t)$, which is continuous with probability 1, and $\exists$ a $K  \in\real^{++}$ such that the following conditions are satisfied:\\
a) Lipschitz condition~\cite{arnold1974stochastic}: $\forall t \in [t_{0},t_{f}]$, $s_{1} \& s_{2} \in \setX \times \setU$,
\begin{align}
    &\|f(s_1) - f(s_2)\| + \|g(s_1)-g(s_2)\|_{\Fnorm} \leq K \|s_1-s_2\|,
\label{eq:lipshitz_stoc}\end{align}
\noindent b) Restriction on growth: $\forall t \in [t_{0},t_{f}]$, $s_1\in \setX \times \setU$
\begin{equation}
    \begin{aligned}
 \|f(s_1)\|^2 + \|g(s_1)\|_{\textit{F}}^2 \leq K^2 (1 + \|s_1\|^2).
    \end{aligned}\label{eq:growth_stoc}
\end{equation}
\end{asmp}
We use the following definition to study the controllability of the deterministic approximation of the SDE~\cref{eq:nl_stochastic_dynamics}. 
\begin{defn}\label{defn:stoc_controllability} The SDE~\cref{eq:nl_stochastic_dynamics} is $\epsilon_{c}$-controllable~\cite{kushner1967stochastic}. For any initial state $x_{0} \in \setX$, we can compute a sequence of control $\baru(t)$ $\forall t \in [t_{0}, t_{f}]$ such that $\Prob\left(\|x - x(t_f)\|^2 \geq \delta \mid x(t_0) = x_0 \right) \leq \epsilon_c$, where $x(t_f)$ is the terminal state, $\delta>0$ and $\epsilon_c>0$ are small, and $t_f$ is finite. 
\end{defn}

\textit{Control Policy.} We assume that the control policy $\Bar{u}(t) \in \setU \subseteq \real^{d_u}$ is deterministic and the set $\setU$ is a convex set. The deterministic control policy is motivated by a hardware implementation strategy, where a state-dependent Markov control policy defined on the compact set $\setX$ is sampled for a value with highest probability (or) for the mean.

\noindent \textbf{Note:} We use the gPC method to project the SDE to an Ordinary Differential Equation (ODE) in a higher dimensional space for propagating the dynamics.

\subsubsection{Chance Constraints~\cite{calafiore2006distributionally}}
In order to accommodate the unbounded uncertainty model in the dynamics, the feasible region $\setX_{\setF}$ defined as,
\begin{equation}
    \setX_{\setF} = \{x(t) \in \setX: h_i(x(t)) \leq  0 \ \forall \ i\  \in\  \{1,\dots, m \}\},
    \label{eq:const_set}
\end{equation} 
is relaxed to a chance constraint (CC) using the risk measure $\epsilon$,
\begin{equation}
    \setX_\mathrm{CC} = \{ x(t) \in \setX: \Prob(x(t) \in \setX_{\setF}) \geq 1- \epsilon\},
    \label{eq:chance_const}
\end{equation}
with a constraint satisfaction probability of $1-\epsilon$. The constraint set $\setX_{\setF}$ is assumed to be the polytope $\setX_{\setF} = \{x \in \mathcal{X}: \land_{i =1}^{m} a_{i}^\top x + b_{i} \leq 0\}$ with $m$ flat sides, or a quadratic constraint set $\setX_{\setF} = \{x \in \mathcal{X}: x^\top A x  \leq c\}$ for any realization $x$ of the state. The joint chance constraint formulation of the polytopic constraint is of the form, $\Prob(\land_{i =1}^{m} a_{i}^\top x + b_{i} \leq 0) \geq 1- \epsilon$. 

\begin{figure}
    \centering
    \includegraphics[width=\columnwidth]{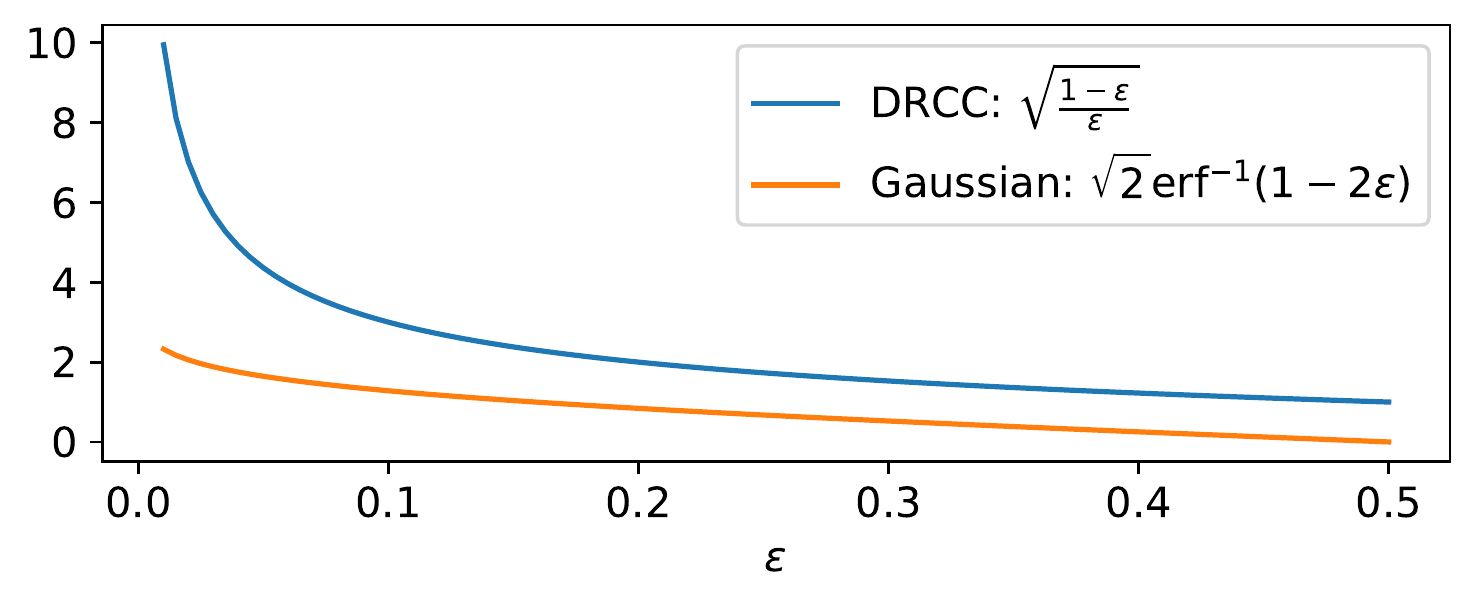}
    \caption{\changed{A comparision of the distributionally robust constraint tightening bound $\sqrt{\frac{1-\epsilon}{\epsilon}}$ used in~\cref{eq:lin_socc} and the Gaussian bound $\sqrt{2}\mathrm{erf}^{-1}(1-2\epsilon)$ used in~\cref{eq:gaussian_linear_constraint}.}}
    \label{fig:dr_vs_gaussian}
\end{figure}

A convex relaxation of the individual chance constraint for an arbitrary distribution of the state vector $x(t)$ due to the nonlinearity in the system is intractable, so an extension of the problem called Distributionally-Robust Chance Constraints (DRCC) given as follows,
\begin{equation}
    \setX_\mathrm{DRCC} = \{ x(t)\in \setX: \inf_{x(t) \sim (\mu_x,\Sigma_{x})} \Prob(x(t) \in \setX_{\setF}) \geq 1- \epsilon \},
    \label{eq:drcc}
\end{equation}
where the chance constraint is satisfied for all distributions with known mean and variance of the decision variable is used. The set defined by the DRCC in~\cref{eq:drcc} is  a conservative approximation ~\cite{zymler2013distributionally} of the chance constraint i.e., $\setX_\mathrm{DRCC} \subseteq \setX_\mathrm{CC}$.

\noindent a) \textit{Distributionally-Robust Linear Chance Constraint (DRLCC)}~\cite{calafiore2006distributionally}: Consider a single Linear Chance Constraint (LCC) with $a \in \real^{d_{x}}$ and $b \in \real$:
\begin{equation}
    \setX_\mathrm{LCC} = \{ x(t)\in \setX:\Prob(a^\top x(t)+ b \leq 0) \geq 1- \epsilon \}.
    \label{eq:cc_lin}
\end{equation}
The column vector $a$, real constant $b$, and risk measure $\epsilon$ are known a priori. The state vector $x$ is the decision variable. Assuming that the mean $\mu_x$ and the covariance $\Sigma_{x}$ of $x$ are known, a distributionally-robust constraint version of \cref{eq:cc_lin} is given as follows:
\begin{equation}
\setX_\mathrm{DRLCC}=\{x(t)\in\setX:\inf_{x(t)\sim(\mu_x,\Sigma_{x})}\Prob(a^\top x(t)+ b\leq 0)\geq 1-\epsilon\}.\label{eq:drcc_lin}
\end{equation}
Equivalently, \cref{eq:drcc_lin} can be rewritten in the following deterministic form, which will be used to derive a second-order cone constraint for the DNOC in~\cref{sec:finite_dimensional_approximation}.
\begin{equation}
  \setX_\mathrm{DRLCC} = \{ x(t)\in \setX:  a^\top \mu_x(t) + b+ \sqrt{\frac{1-\epsilon}{\epsilon}}\sqrt{a^\top \Sigma_{x}a} \leq 0 \}
    \label{eq:lin_socc}
\end{equation} 
\begin{lemma} \label{lemma:lin_consev_approx}
The set $\setX_\mathrm{DRLCC}$ in~\cref{eq:lin_socc} is a subset of $\setX_\mathrm{LCC}$ defined in \cref{eq:cc_lin}.
\end{lemma} 
\begin{proof}
See Theorem 3.1 in~\cite{calafiore2006distributionally}.  
\end{proof}
If the dynamics are linear, the constraint in~\cref{eq:lin_socc} is replaced with a tighter equivalent deterministic constraint given by the following inequality:
\begin{equation}
    a^\top \mu_x(t) + b+ \sqrt{2}\mathrm{erf}^{-1}(1-2\epsilon)\sqrt{a^\top \Sigma_{x}a} \leq 0,
    \label{eq:gaussian_linear_constraint}
\end{equation}
where the function $\mathrm{erf}(.)$ is defined as $\text{erf}(\delta)=\frac{2}{\sqrt{\pi}}\int_{0}^{\delta} e^{-t^2} dt$ and $ \forall \epsilon \in (0,0.5)$. The constraint set is transformed to a second-order cone constraint in the gPC variables. \changed{In~\cref{fig:dr_vs_gaussian}, we show a comparison of the robustness provided by the distributional robustness cosntraint~\cref{eq:lin_socc} and the Gaussian constraint~\cref{eq:gaussian_linear_constraint}.}
\begin{rmrk}
The risk measure $\epsilon$ in~\cref{eq:cc_lin} is assumed to be in the range $[0.001,0.5]$. For small values of $\epsilon$ ($< 0.001$), the value $\sqrt{\frac{1-\epsilon}{\epsilon}}$ increases dramatically, as shown in~\cref{fig:dr_vs_gaussian}. This decreases the feasible space defined by the set $\setX_\mathrm{LCC}$ drastically leading to numerical issues in the gPC-SCP method. For handling the risk of a very small value of $\epsilon$ (e.g., $1e-7$, as discussed in~\cite{cheng2021limits}), the uncertainty in the system needs to be modeled accurately such that $\Sigma_x$ is small or a newer deterministic surrogate method needs to be developed to overcome the numerical instability.
\end{rmrk}

\noindent b) \textit{Conservative Quadratic Chance Constraint (CQCC):} \cref{lemma:quad_const} presents a new conservative deterministic relaxation for the quadratic chance constraint that is used to bound the deviation of the random vector $x(t)$ from the mean $\mu_x(t)$.
\begin{lemma}
\label{lemma:quad_const}
The constraint set 
\begin{equation}
 \setX_\mathrm{CQCC} = \{ x(t)\in \setX: \frac{1}{c}\text{tr}(Q\Sigma_x) \leq \epsilon\}
 \label{eq:sdp_constraint}
\end{equation}
is a conservative approximation of the original Quadratic Chance Constraint (QCC) 
\begin{equation}
 \setX_\mathrm{QCC} = \{ x\in \setX:  \Prob((x-\mu_x)^\top Q (x-\mu_x) \geq c) \leq \epsilon\}
 \label{eq:quad_chnc_cons}
\end{equation}
i.e., $\setX_\mathrm{CQCC} \subseteq \setX_\mathrm{QCC}$, where $Q \in \real^{n\times n}$ is a positive definite matrix and $c \in \real^{++}$ and $\Sigma_x$ is the covariance of the random variable $x$
\end{lemma}
\changed{\begin{proof}
We prove that any random vector $x \in \setX$ that is in the set $\setX_\mathrm{CQCC}$ is also in the set $\setX_\mathrm{QCC}$ implying  $\setX_\mathrm{CQCC} \subseteq \setX_\mathrm{QCC}$. The proof follows from the approach taken to prove the multivariate Chebyschev's inequality~\cite{chen2007new}. Let $F(x)$ be the Cumulative Distribution Function (CDF) of the random variable $x$ and $v = x - \mu_x$.
\begin{equation*}
    \begin{aligned}
    \mathcal{G}  = \{ v \in \setX : v^\top Q v \geq c \} \implies \frac{1}{c} v^\top Q v \geq 1 \ \ \forall v \in \mathcal{G}.
    \end{aligned}
\end{equation*}
Using the definition of probability in terms of the CDF,
\begin{equation*}
    \begin{aligned}
    \mathrm{Pr}((x-\mu_x) \in \mathcal{G}) & \leq \frac{1}{c} \int_{v\in \mathcal{G}} v^\top Q v dF(v)\\ & \leq \frac{1}{c} \int_{v \in \mathcal{R}^n}v^\top Q v dF(v).
    \end{aligned}
\end{equation*}
Let $q_{ij}$ denote an element of matrix $Q$ in the $i^\mathrm{th}$ row and $j^\mathrm{th}$ column, and $v_{i}$ be the $i^\mathrm{th}$ element in the vector $v$.
Using the expansion $v^\top Q v = \sum_{i = 1}^{d_{x}}\sum_{j = 1}^{d_{x}} q_{ij}v_{i}v_{j}$ in the inequality above, the integral is simplified.
\begin{equation}
    \begin{aligned}
    \int_{v \in \mathbb{R}^n}v^\top Q v dF(v) & = \int_{v \in \mathbb{R}^{v}} \sum_{i = 1}^{d_{x}}\sum_{j = 1}^{d_{x}} q_{ij}v_{i}v_{j} dF(v) \\
    & = \sum_{i = 1}^{d_{x}}\sum_{j = 1}^{d_{x}} q_{ij} \int_{v \in \mathbb{R}^{d_{x}} }v_{i}v_{j} dF(v)\\
    & = \text{tr}(Q \Sigma_{x}).\\ 
    \end{aligned}
\end{equation}
The quadratic chance constraint  holds if~(\ref{eq:sdp_constraint}) is satisfied, as $\mathrm{Pr}((x-\mu_x)\in \mathcal{G}) \leq \frac{1}{c}\text{tr}(Q\Sigma_{x})$. Therefore,~(\ref{eq:sdp_constraint}) is a conservative deterministic approximation of the quadratic chance constraint $\mathrm{Pr}((x-\mu_x)^\top Q (x-\mu_\vecx) \geq c) \leq \epsilon$ i.e., $\setX_\mathrm{CQCC} \subseteq \setX_\mathrm{QCC}$. Note that if $\epsilon$ is a design variable, the approximation can be made tight by solving an inner semi-definite program following the approach in~\cite{vandenberghe2007generalized}. \end{proof}}

\begin{corollary}
The constraint set $\frac{1}{c}\text{tr}(A\Sigma_x) + \frac{1}{c}(\mu_x^\top A \mu_x )\leq \epsilon$ is a conservative approximation of the quadratic chance constraint $\Prob(x^\top A x \geq c) \leq \epsilon\}$, where $A \in \real^{n\times n}$ is a positive definite matrix and $c \in \real^{++}$ and $\Sigma_x$ is the co-variance of the random variable $x$.
\end{corollary}
\begin{proof}
\changed{The proof follows from~\cref{lemma:quad_const} by transforming $x$ to $x+ \mu_x$ in \Cref{eq:quad_chnc_cons}.}
\end{proof}
\noindent c) \textit{Joint Chance Constraints (JCC)}~\cite{zymler2013distributionally}: 
  The distributionally-robust joint chance constraint (DRJCC) for a polytope set is defined as
  $\inf_{x(t) \sim (\mu_x,\Sigma_{x})} \Prob(\land_{i = 1}^{m} a_{i}^\top x + b_{i} \leq 0) \geq 1- \epsilon$. The joint constraints are split into multiple single chance constraints using Bonferroni's inequality~\cite{zymler2013distributionally} method as follows:
\begin{equation}
\begin{aligned}
   & \inf_{x(t) \sim (\mu_x,\Sigma_{x})} \Prob(\land_{i = 1}^{m} a_{i}^\top x + b_{i} \leq 0) \geq 1- \epsilon \\ & \iff \sup_{x(t) \sim (\mu_x,\Sigma_{x})} \Prob( \lor_{i = 1}^{m} a_{i}^\top x + b_{i} \geq 0) \leq \epsilon \\ & \subseteq  \sum_{i = 1}^{m} \sup_{x(t) \sim (\mu_x,\Sigma_{x})}  \Prob( a_{i}^\top x + b_{i} \geq 0) \leq \epsilon.
\end{aligned} \label{eq:benf_inq}
\end{equation}
If the probability distribution of $x$ is Gaussian, then the JCC are split using Boole's inequality~\cite{blackmore2010probabilistic}. The total risk measure $\epsilon$ is allocated between each of the chance constraints  in the summation such the $\sum_{i = 1}^{m} \epsilon_{i}  = \epsilon$ leading to $m$ individual DRCC of the following form.
\begin{equation}
    \inf_{x(t) \sim (\mu_x,\Sigma_{x})}  \Prob( a_{i}^\top x + b_{i} \leq 0) \geq 1-\epsilon_{i}
    \label{eq:benf_inequality_indv}
\end{equation}
We follow a naive risk allocation approach by equally distributing the risk measure $\epsilon$ among the $m$ constraints such that $\epsilon_{i} = \frac{\epsilon}{m}$. Alternatively, optimal risk allocation~\cite{hiro2008} can be achieved using iterative optimization techniques. Using distributional robustness, \cref{prob:cc_stoc_opt_cntrl} is reformulated to the following~\cref{prob:dr_stoc_opt_cntrl}. 
\begin{problem}
\label{prob:dr_stoc_opt_cntrl}
Distributionally-Robust Chance-Constrained Stochastic Nonlinear Optimal Control.
\begin{equation*}
\begin{aligned}
J_{\mathrm{DR-SNOC}}^{*} =& \underset{x(t),\Bar{u}(t)}{\min}
& & \scalebox{0.9}{$\Exp \left[\int_{t_{0}}^{t_{f}}J(x(t),\Bar{u}(t))dt + J_{f}(x(t_{f}))\right]$} \\
& \text{s.t.} & & \cref{eq:nl_stochastic_dynamics},~\cref{eq:lin_socc},~\cref{eq:sdp_constraint},~\cref{eq:control_limits}, and~\cref{eq:init_term_conditions}.
\end{aligned} \label{eq:dr_stoptprob}
\end{equation*}
\end{problem}
\textbf{Note:} Given a risk measure $\epsilon$, the constraints in~\cref{prob:dr_stoc_opt_cntrl} are a function of mean $\mu_x$ and covariance matrix $\Sigma_x$ of the state at any time $t$. \changed{While this enables fast computation of chance constraints, it reduces the feasible space $\setX_{\setF}$. We compute a distributionally robust deterministic subset of the stochastic set $\setX_{\setF}$ that is convex in gPC space, as discussed in the following~\cref{subsec:gPC-SCP-scp}. We present empirical evidence that using the distributional robustness approach does not lead to infeasibility in practical scenarios. This approach should be integrated with system design and modelling to ensure feasibility. We transform the SNOC problem into a DNOC problem by applying the generalized polynomial chaos expansion. This approach transforms the infinite-dimensional SNOC problem in both the state (stochastic state) and time to a DNOC problem with infinite dimension only in time.}

\subsection{Generalized Polynomial Chaos} \label{subsec:gpc}
The generalized Polynomial Chaos (gPC)~\cite{xiu2002wiener,xiu2009fast,boutselis2017stochastic} expansion theory is used to model uncertainty with finite second-order moments as a series expansion of orthogonal polynomials. The polynomials are orthogonal with respect to a known density function $\rho(\cdot)$. Consider the random vector $\xi$ with independent identically distributed (iid) random variables $\{\xi_{i}\}_{i=1}^{d_{\xi}}$ as elements. Each $\xi_{i} \sim \setN(0,1)$ is normally distributed with zero mean and unit variance. The random vector $x(t)$, defined by the SDE in~\cref{eq:nl_stochastic_dynamics}, can be expressed as the following series\begin{equation}
    \scalebox{1}{$x_{i}(t) = \sum_{j = 0}^{\infty} x_{ij}(t) \phi_{j}(\xi)$},
    \label{eq:series_expan}
\end{equation}
where $x_{i}$ denote the $i^\mathrm{th}$ element in the vector $x \in \setX$ and $x_{ij}$ is the $j^\mathrm{th}$ coefficient in the series expansion. The dimension $d_{\xi}$ is the sum of the number of random inputs in the SDE~\cref{eq:nl_stochastic_dynamics} and the number of random initial conditions. The functions $\phi_{j}(\xi)$ are constructed using the Hermite polynomial~\cite{xiu2002wiener} basis functions. The functions $\phi_{j}(\xi)$ are orthogonal with respect to the joint probability density function $\boldsymbol{\rho}(\xi) = \varrho(\xi_1)\varrho(\xi_2)\cdots\varrho(\xi_{d_{\xi}})$, where $\varrho(\xi_{k}) = \tfrac{1}{\sqrt{2\pi}} e^{\tfrac{-\xi_{k}^2}{2}}$. The choice of the orthogonal polynomials depends on the uncertainty model effecting the dynamics. We refer to~\cite{ghanem2003stochastic} for details on type and construction of the polynomials for different standard uncertainty models such as uniform, beta and Poisson distributions. 
\begin{rmrk} \label{rmrk:P_gpc_and_ell}
The series expansion is truncated to a finite number $\ell+1$ as $x_{i} \approx \sum_{j = 0}^{\ell} x_{ij}(t) \phi_{j}(\xi)$ based on the maximum degree of the polynomials ($\mathrm{P}_{\gpc}$) required to represent the variable $x$. The minimum $\ell$ required to appropriately represent $x$ with uncertainty parameter $\xi \in \real^{d_{\xi}}$ is given by $\ell = \Big(\begin{smallmatrix} \mathrm{P}_{\gpc} + d_{\xi} \\ d_{\xi}\end{smallmatrix} \Big) - 1$.
\end{rmrk}
The coefficients $x_{ij}(t)$ are computed using the Galerkin projection given  by the following equation: 
\begin{equation}
    x_{ij}(t) = \frac{\int_{\mathbb{D}} \boldsymbol{\rho}(\xi) x_i (t) \phi_{j}(\xi) d\xi}{\langle \phi_j(\xi),\phi_j(\xi) \rangle},\label{eq:galerkin_projec}
\end{equation}
where $\langle\phi_i(\xi),\phi_j(\xi)\rangle=\int_{\mathbb{D}}\boldsymbol{\rho}(\xi)\phi_i (\xi)\phi_j(\xi) d\xi$. For non-polynomial functions, the Galerkin projection is computed using the Stochastic Collocation~\cite{xiu2009fast} method as follows:
\begin{equation}
    \int_{\mathbb{D}} \rho(\xi) x_i (t) \phi_{j}(\xi) d\xi \approx \sum_{k = 1}^{m} w_{k} x_i (t) \phi_{j}(n_{k}),
    \label{eq:stoch_coll}
\end{equation} 
where Gauss-Hermite quadrature is used to generate the nodes $n_{k}$ and the corresponding node weights $w_{k}$. In the following section, we derive an approximate nonlinear ordinary differential equation system for the SDE in~\cref{eq:nl_stochastic_dynamics} using gPC expansion and the Galerkin scheme. The DRCC are projected to the gPC coordinates $x_{ij}$ leading to convex constraints. \cref{lemma:conv_gpc,lemma:approximation_error} discuss the convergence of the gPC expansion to the true distribution and the error due to truncated polynomial approximation of a distribution.
\begin{figure}\begin{center}
\includegraphics[width=0.4\columnwidth]{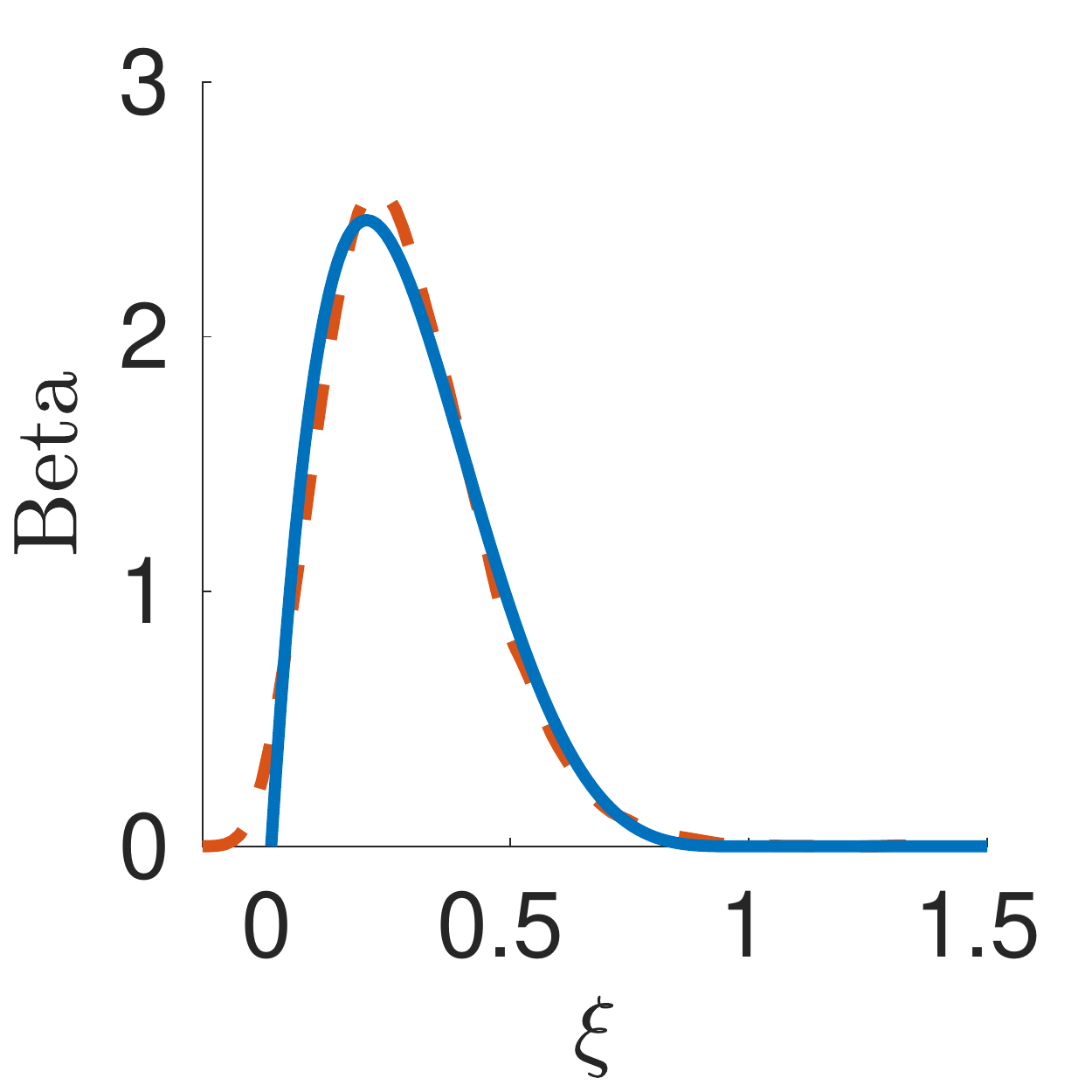}\includegraphics[width=0.4\columnwidth]{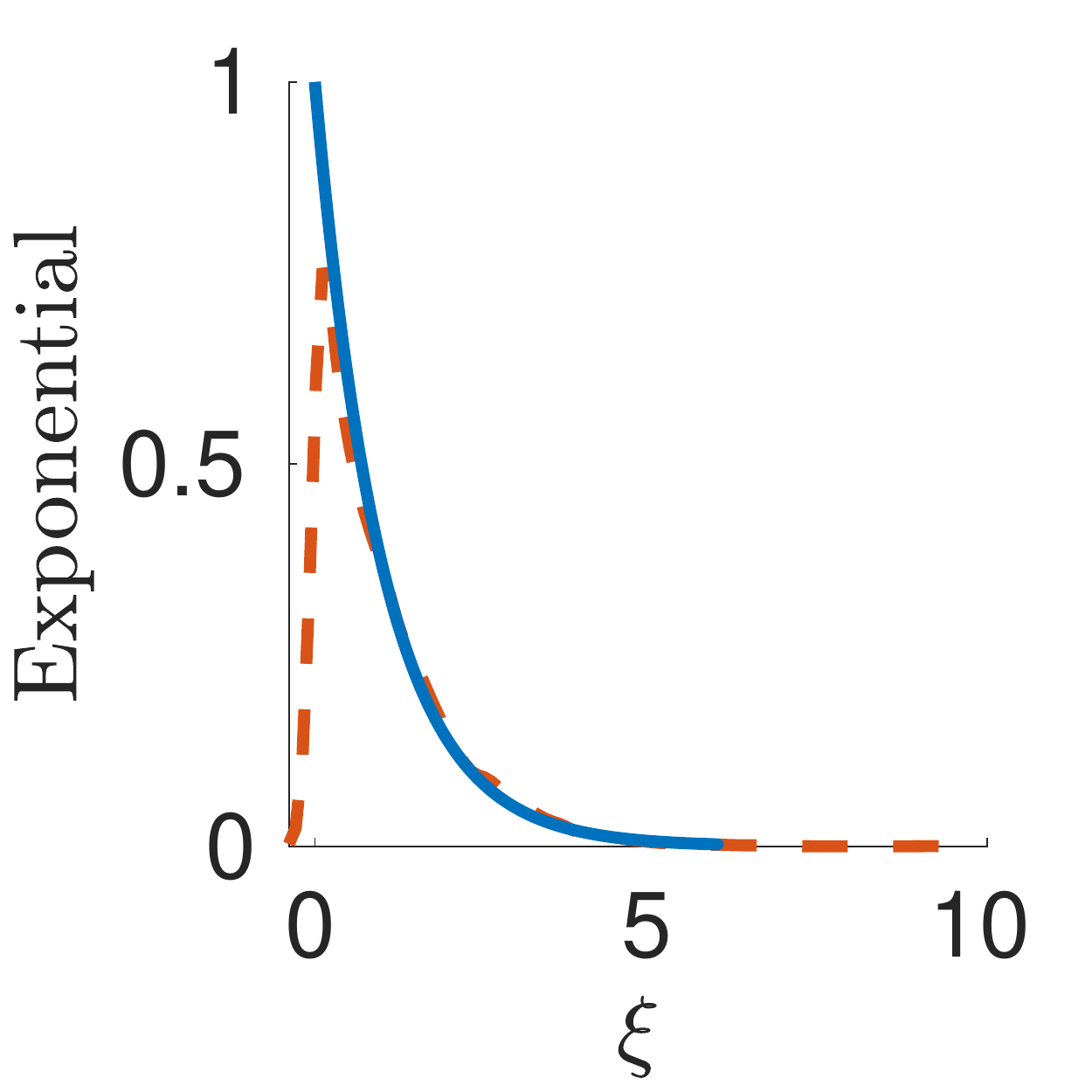} 
\includegraphics[width=0.4\columnwidth]{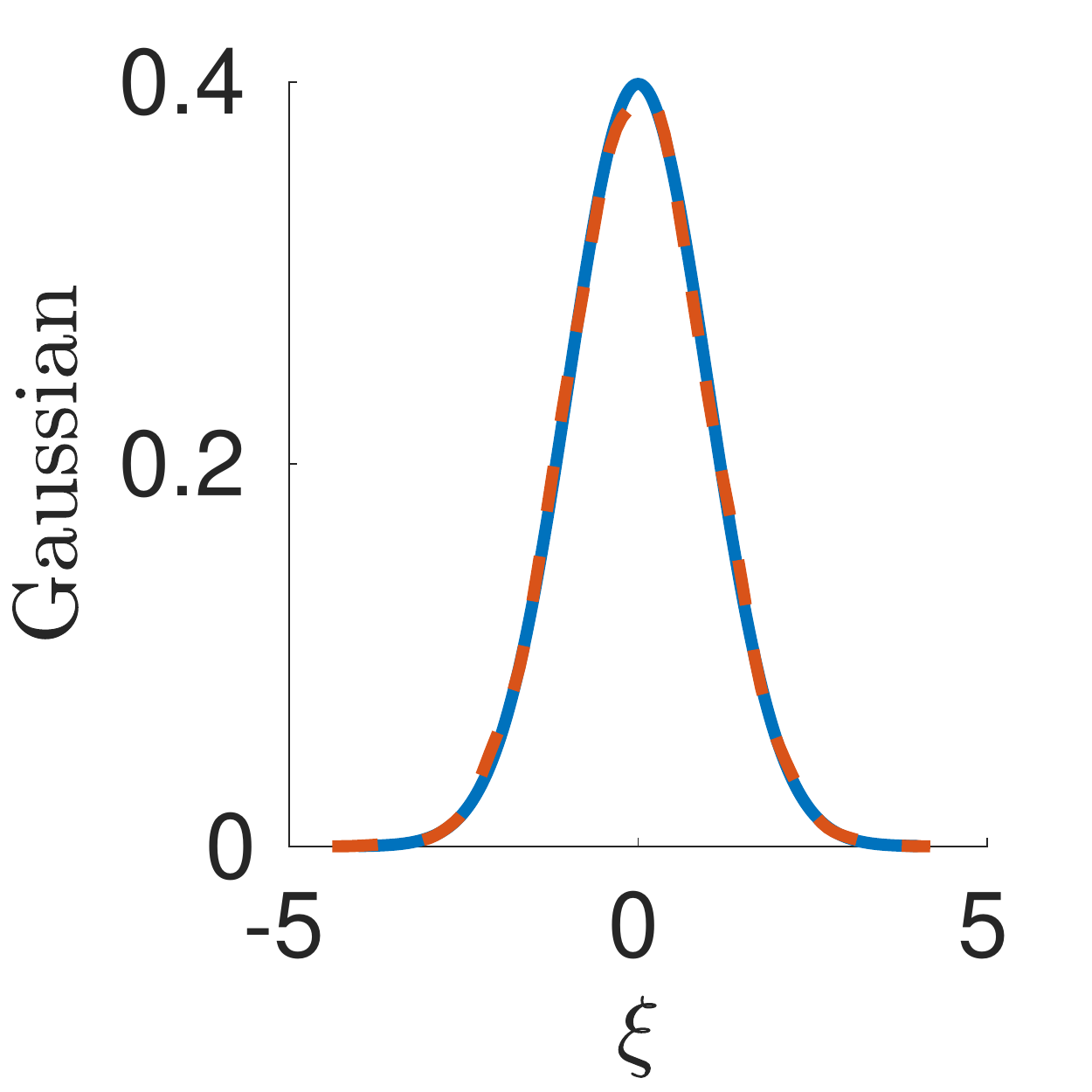}\includegraphics[width=0.4\columnwidth]{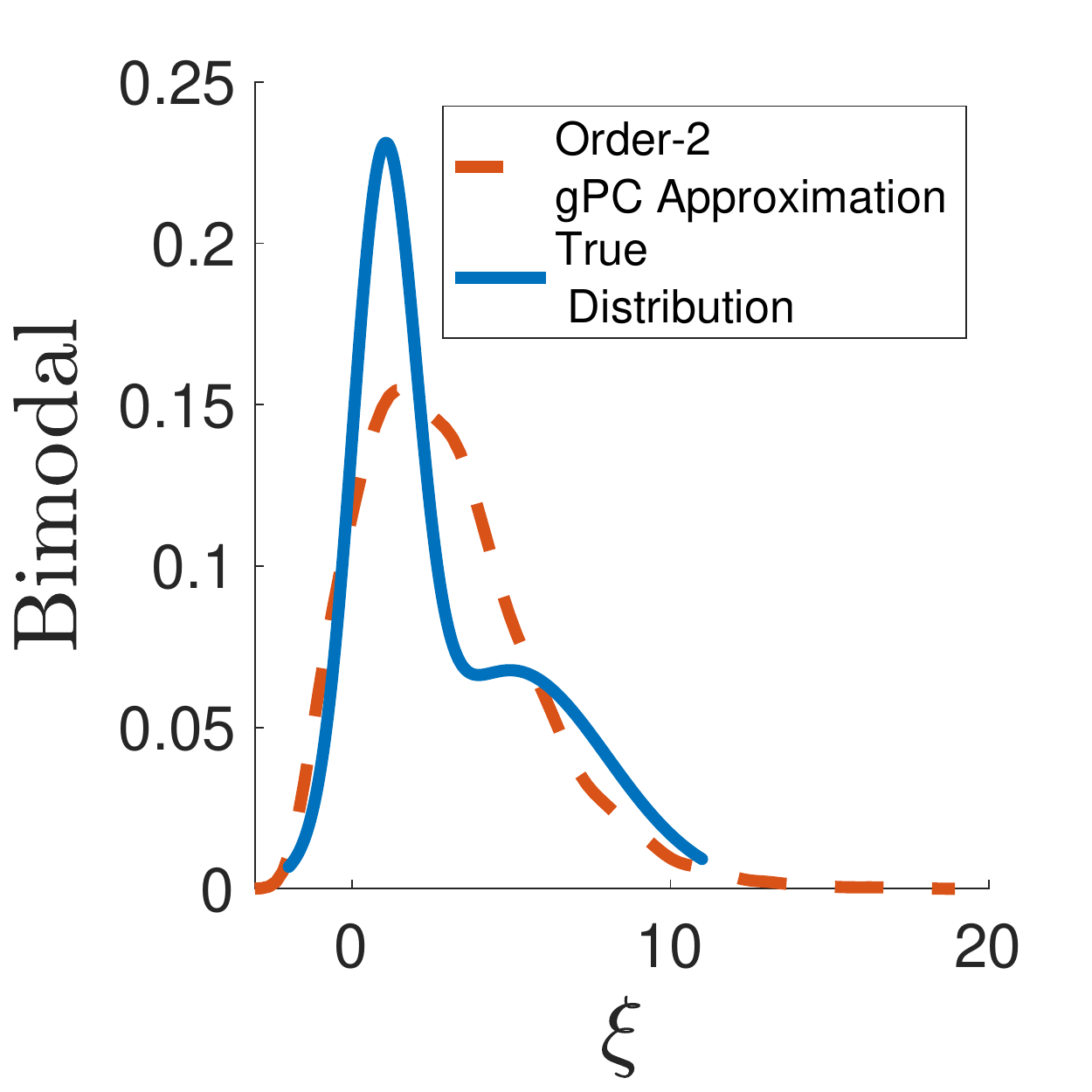}
\caption{Example gPC approximation of some standard probability distribution functions (PDF) using gPC expansion. For the beta and exponential distributions, gPC expansion represents the PDF well with just second order $P_{\gpc} = 2$ and $d_{\xi} = 1$ approximation. For a Gaussian distribution, the gPC representation is exact.}\label{fig:gpc_examples}\end{center}
\end{figure}
\begin{lemma}\label{lemma:conv_gpc}(Cameron-Martin Theorem~\cite{cameron1947orthogonal}) The gPC series approximation in~\cref{eq:series_expan} converges to the true value $x_{i} \in \mathcal{L}_2$. 
\begin{equation}
    \scalebox{1}{$\|x_{i}(t) - \sum_{j = 0}^{\ell} x_{ij}(t) \phi_{j}(\xi) \|_{\mathcal{L}_2}$} \to 0,\ \text{as} \ \ell \ \to \ \infty \ \forall \ t \in [t_{0},t_{f}]
\end{equation}
\end{lemma}
\begin{rmrk} \label{rmrk:exp_var_gpc} The expectation $\Exp(x_{i})$ and variance $\Sigma_{x_{i}}$ of the random variable $x_{i}$ can be expressed in terms of the coefficients of the expansion as follows:
\begin{equation}
    \Exp(x_{i}) = x_{i0},  \quad \Sigma_{x_{i}} \approx \sum_{j = 1}^{\ell} x_{ij}^2 \langle \phi_{j},\phi_{j} \rangle \ \text{as} \ \ell \to \infty.
    \label{eq:exp_var}
\end{equation}
\end{rmrk}
\begin{lemma}\label{lemma:approximation_error} (Truncation Error Theorem~\cite{muhlpfordt2017comments}) If an element $x_i$ of the random variable $x$ is represented using $\ell$ polynomials, then the approximation error is given as follows:
\begin{align}
\|x_i -\sum_{j = 0}^{\ell} x_{ij}(t) \phi_{j}(\xi)\|  = \|e_\ell\| \leq \sqrt{\sum_{j = \ell+1}^{\infty}x^2_{ij}\|\phi_{j}\|^2}. 
\end{align}
\end{lemma} 
\cref{lemma:conv_gpc,lemma:approximation_error}, and~\cref{rmrk:exp_var_gpc}  will be used in studying the convergence of the gPC approximation of the cost function, the SDE, and the chance constraints. Furthermore, the higher-order moments can be expressed as a polynomial function of the coefficients.
\subsubsection*{Curse of Dimensionality} The truncated polynomial expansion is a finite-dimensional approximation of the random variable. The number of polynomials $\ell$ grow exponentially large based on the degree of polynomial used to represent the state distribution. The large dimensionality can be reduced, inducing sparsity in the gPC expansion, by using  techniques like sparse gPC~\cite{blatman2011adaptive}, and data-driven gPC~\cite{oladyshkin2012data}. A cost-effective approach to estimate moments up to second order is to use gPC polynomials up to degree 2, i.e., $\mathrm{P}_{\gpc} = 2$~\cite{xu2018novel}. For a given $\mathrm{P}_{\gpc}$, we use~\cref{rmrk:P_gpc_and_ell} to compute $\ell$. The computationally complexity for $\mathrm{P}_{\gpc} = 2$ is equivalent to linear covariance propagation. Note that, unlike the linear covariance propagation method, the gPC method with $\mathrm{P}_{\gpc}=2$ accounts for the coupling between the state $x$ and the white-noise process $dw$.

\section{Deterministic surrogate of the SNOC Problem} \label{sec:det_opt_prob}
The stochastic nonlinear optimal control problem discussed in~\cref{subsec:stoc_opt_prob} is reformulated in terms of the coefficients of the gPC expansion, with decision variables as the gPC coefficients and the control $\Bar{u}$. In the following, we discuss the existence and uniqueness of a solution to the coupled Ordinary Differential Equations (ODE) obtained form gPC approximation of SDE, the cost function in the gPC space, and present the convex constraints for the gPC coefficients obtained from deterministic approximation of chance constraints. We present the convergence and feasibility theorem of the approximation at the end of this section. 
\subsection{Deterministic ODE Approximation of the SDE}
The gPC expansion in~\cref{eq:series_expan} is applied for all the elements in the vector $x \in \setX \subseteq \real^{d_{x}}$ and the matrix representation using Kronecker product is given in the following, where $\gpcX = \begin{bmatrix}x_{10} &\cdots &x_{1\ell} & \cdots & x_{d_{x}0} & \cdots& x_{d_{x}\ell}\end{bmatrix}^\top$ are gPC states.
\begin{equation}
    \Phi(\xi) = \begin{bmatrix} \phi_{0}(\xi) & \cdots & \phi_{\ell}(\xi)\end{bmatrix}^\top
    \label{eq:phi_gpc_matrix}
\end{equation}
\begin{equation}
    x \approx \Bar{\Phi} \gpcX ;\ \text{where}\ \Bar{\Phi} = \mathbb{I}_{d_{x}\times d_{x}} \otimes \Phi(\xi)^\top
    \label{eq:state_kronck_notation}
\end{equation}
Consider the following Ito's integral form of the SDE in~\cref{eq:nl_stochastic_dynamics}.
\begin{equation}
    x(t) = x(t_0) + \int_{t_0}^{t} f(x,\baru) dt + \int_{t_0}^{t} g(x, \bar{u})dw 
    \label{eq:ito_nl_sde}
\end{equation}
The gPC projection of the above SDE is given by the following ODE.
\begin{align}
    \hspace{-5pt}x_{ij}(t) & = x_{ij}(t_{0}) + \int_{t_{0}}^{t}\Bar{f}_{ij}(\gpcX,\Bar{u})dt + \int_{t_{0}}^{t}\Bar{g}_{ij}(\gpcX,\Bar{u})\sqrt{dt} \label{eq:continuous_ode_form}\\
    &\Bar{f}_{ij} = \frac{\int_{\mathbb{D}}\rho(\xi)\phi_{j}(\xi)f_{i}(\Bar{\Phi}\gpcX,\Bar{u})d\xi}{\langle \phi_j(\xi),\phi_j(\xi) \rangle}, \nonumber\\
    &\Bar{g}_{ij} = \frac{\int_{\mathbb{D}}\rho(\xi)\phi_{j}(\xi)g_{i}(\Bar{\Phi}\gpcX,\Bar{u}) \xi d\xi}{\langle \phi_j(\xi),\phi_j(\xi) \rangle} \nonumber
\end{align}
The dynamics of the coefficients $x_{ij}$ with the above notation is given in~\cref{eq:ode_form}, where: $f_{i}$ and $g_{i}$ are the $i^\mathrm{th}$ element of the vector $f$ and $i^\mathrm{th}$ row of the matrix $g$ respectively.  We use the Euler-Maruyama discretization method of the SDE for time integration. The discrete time stochastic dynamics is given as follows:
\begin{equation}
    x[k+1] = x[k] + f(x[k],\baru[k])\Delta t + g(x[k],\bar{u}[k]) \sqrt{\Delta t} \xi,
    \label{eq:discrete_sde}
\end{equation}
where $x[k]$, $\baru[k]$ are the states and controls at time step $k$, $\Delta t$ is the integration time interval, and $\xi$ is a multivariate Gaussian distribution $\setN(0,\mathbb{I})$. The discrete stochastic system is projected to a discrete deterministic system using the gPC method. 
\begin{align}
    x_{ij}[k+1] & = x_{ij}[k] + \Bar{f}_{ij}(\gpcX[k],\Bar{u}[k])\Delta t \nonumber\\&+ \Bar{g}_{ij}(\gpcX[k],\Bar{u}[k])\sqrt{\Delta t} \label{eq:ode_form}
\end{align}

\begin{figure}
    \centering
    \includegraphics[width=0.9\columnwidth]{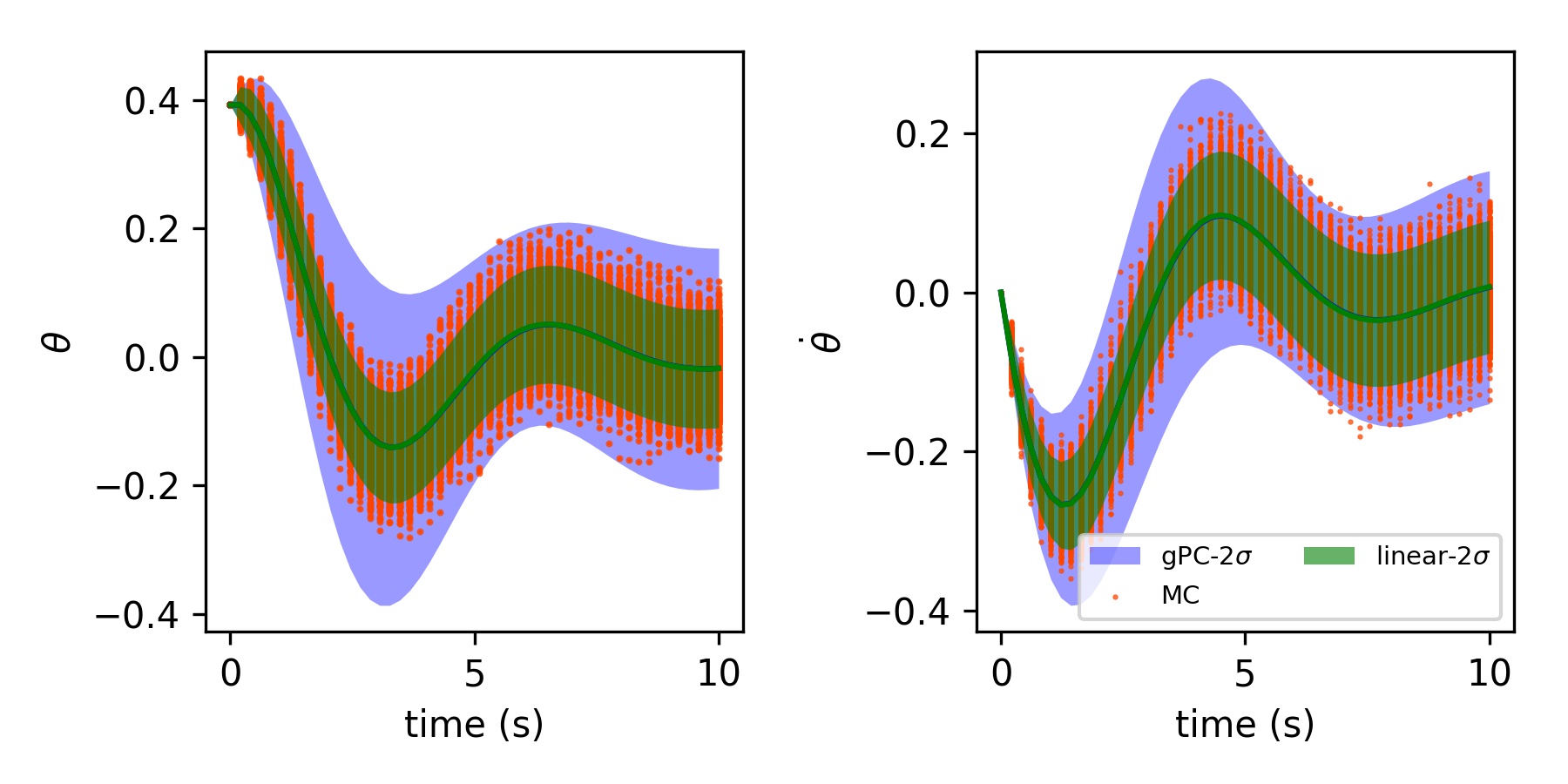}
    \caption{Example gPC propagation for a pendulum. The figure compares the mean and $2\sigma$ confidence computed using gPC Projection~\cref{eq:discrete_sde} ($P_{\gpc}=1$), linear covariance propagation, and Monte Carlo (MC) propagation of the simple pendulum dynamics $\ddot{\theta} = -\sin{\theta} -0.8\dot{\theta} + \sqrt{0.001}\xi(t)$. It is observed that the gPC approximation overestimates the variance compared to MC and the linear covariance propagation underestimates the variance. The $P_{\gpc}=1$ projection corresponds to a Gaussian approximation that includes the cross correlation between the state and uncertainty.}
    \label{fig:gpc_propgation}
\end{figure}
The full nonlinear discrete time ODE with the stacked vector $\gpcX$ is given as follows:
\begin{align}
    \gpcX[k+1] =& \gpcX[k] + \Bar{f}(\gpcX[k],\Bar{u}[k],\Delta t) \nonumber\\
    &+\Bar{g}(\gpcX[k],\Bar{u}[k],\sqrt{\Delta t}).\label{eq:odefull_form}
\end{align}
\Cref{fig:gpc_propgation} shows an example of propagation using~\cref{eq:odefull_form}. While not discussed in this paper, the projection is also applicable to a higher-order discretization methods~\cite{platen2010numerical}. The sequential convex programming method used for trajectory optimization involves successive linearizations~\cite{morgan2014model} of the dynamics about a given trajectory and discretization for time integration. In~\cref{prop:exist_uniq}, we present the conditions for existence and uniqueness of the solution to the projected system. The existence and uniqueness of solution to the ODE surrogate ensure convergence of any Picard iteration scheme used for integration.
\begin{prop} \label{prop:exist_uniq}
The ODE system~\cref{eq:ode_form} obtained using gPC approximation of the SDE has a solution and the solution is unique, for a given initial condition, assuming that the SDE satisfies the existence and uniqueness conditions in~\cref{eq:lipshitz_stoc},~\cref{eq:growth_stoc} and the expectation
\begin{equation}
    \begin{aligned}
    K_{g_{ij}} & = \frac{K}{k_{j}} \Exp(L_{g_{j}} (\xi)), K_{f_{j}} & = \frac{K}{k_{j}}\Exp(L_{f_{j}}(\xi))
    \end{aligned} \label{eq:bound_lipz}
\end{equation}
in~\cref{eq:bound_lipz} are bounded for each $j = 0,1,\cdots,\ell$, where:$k_{j} = \langle \phi_{j},\phi_{j} \rangle$, $L_{f_{j}}(\xi) = |\phi_{j}(\xi)|\|\big[\begin{smallmatrix}\Bar{\Phi} & 0 \\ 0 & \mathbb{I} \end{smallmatrix}\big]\|_{2}$, $L_{g_{j}} (\xi)= L_{f_{j}}(\xi) |\phi_{1}(\xi)|$. The constants $K_{g_{ij}}$ and $K_{f_{j}}$ are the Lipschitz coefficients of the projected functions $\bar{g}_{ij}$ and $\bar{f}_{j}$ respectively.
\end{prop}
\begin{proof}See Proposition 2 in~\cite{nakka2019nsoc} for the proof.\end{proof} 
While the projection operation preserves the existence and uniqueness properties of the SDE, it might not retain the controllability of the moments of the system. Following examples discuss on how the $\epsilon_c$ controllability in~\cref{defn:stoc_controllability} of the SDE effects the controllablility of the projected ODE system. 
\begin{example} \label{example:linear_example_1}
Consider the linear SDE $dx = xdt + \bar{u}dt + \sqrt{dt}\xi$, where $x \in \real^{1}$, $\bar{u} \in \real^{1}$ and $\xi \sim \setN(0,1)$. Using the random variable $\xi$ as the variable, we can construct the first order gPC expansion $x = x_{0} + x_{1}\xi$ of the state with $x_{0},x_{1} \in \real^{1}$. The projected dynamics using the expansion is given as follows:
\begin{equation}
    \begin{bmatrix}
    dx_{0} \\ dx_{1}
    \end{bmatrix} = \begin{bmatrix}
    1 & 0 \\ 0 & 1 \end{bmatrix} \begin{bmatrix} x_{0} \\ x_{1} \end{bmatrix} dt + \begin{bmatrix}
    1 \\ 0 \end{bmatrix} \bar{u} dt +  \begin{bmatrix} 0 \\ \sqrt{dt} \end{bmatrix}.
    \label{eq:linear_sde_example}
\end{equation}
The dynamics of $x_{1}$ is decoupled from $x_{0}$ and the propagation is not influenced by the control $\bar{u}$. Notice that, even though the original SDE ($dx=xdt+\bar{u}dt$) is controllable, the projected system~\cref{eq:linear_sde_example} is not fully controllable. The projection operation converts a SDE to an ODE in higher dimensions. Though this operation enables for fast uncertainty propagation, the linear projected system is underactuated and not fully controllable. 
\end{example}
\begin{rmrk} \label{remark:feedback_covariance_control}
Using a stochastic state feedback of the form $u = -kx$ in~\cref{example:linear_example_1}, we get the closed-loop SDE $dx=(1-k)xdt + \sqrt{dt}\xi$. The gPC projection of the system is as follows:
\begin{equation}
    \begin{bmatrix}
    dx_{0} \\ dx_{1}
    \end{bmatrix} = \begin{bmatrix}
    1-k & 0 \\ 0 & 1-k \end{bmatrix} \begin{bmatrix} x_{0} \\ x_{1} \end{bmatrix} dt +  \begin{bmatrix} 0 \\ \sqrt{dt} \end{bmatrix}.
    \label{eq:linear_sde_randomfeedback_example}
\end{equation}
Using a stochastic feedback, the state $x_{1}$ that corresponds to the variance of the SDE can be controlled. 
\end{rmrk}

\begin{example}\label{example:nonlinear_example_2} The gPC projection of the nonlinear SDE $dx = x^2 dt + \sqrt{dt}\xi$ using the expansion $x = x_{0} + x_{1}\xi$ is given as follows:
\begin{equation}
    \begin{bmatrix}
    dx_{0} \\ dx_{1}
    \end{bmatrix} = \begin{bmatrix}
    x^2_{0} + x^2_{1}+ \bar{u}\\ 2x_{0}x_{1} \end{bmatrix}dt +  \begin{bmatrix} 0 \\ \sqrt{dt} \end{bmatrix}.
    \label{eq:nonlinear_sde_example}
\end{equation}
The projected system~\cref{eq:nonlinear_sde_example} is underactuated. In the case of nonlinear systems, the coupling between the dynamics of $x_0$ and $x_1$ allows for indirectly controlling the state $x_1$.
\end{example}

\begin{rmrk}\label{remark:control} The gPC projected ODE system in~\cref{eq:ode_form} might not be fully controllable as discussed in~\cref{example:linear_example_1}. We introduce a slack variable terminal constraint on the variance of the state variable to ensure the feasibility of~\cref{prob:cc_stoc_opt_cntrl} in accordance with~\cref{defn:stoc_controllability}. Based on the value of the slack variable, this would increase (or) decrease the probability of reaching the terminal set.
\end{rmrk}
With~\cref{remark:control} on the controllability of the projected system, we proceed to construct a finite-dimensional approximation of the cost functional and chance-constraints to formulate the convex-constrained nonlinear deterministic optimal control problem.
\subsection{Cost Function}
\label{sec:cost_functioj}
Using the notation in~\cref{eq:state_kronck_notation}, the expectation of the cost functional in~\cref{eq:cost_stopt} is expressed in the gPC coefficients as follows:
\begin{align}
\begin{aligned}
  J_{\gpc}(\gpcX(t),\Bar{u}(t))&= \gpcX(t)^\top Q_{\gpc} \gpcX(t)  + \|\bar{u}\|_{p},\\
    J_{\gpc_{f}}(\gpcX(t_{f}))&=\gpcX(t_f)^\top Q_{\gpc_{f}} \gpcX(t_f),
\end{aligned}\label{eq:cost_odeopt}
\end{align}
where $Q_{\gpc} = \Exp(\Bar{\Phi}^\top Q \Bar{\Phi})$ and $Q_{\gpc_{f}} = \Exp(\Bar{\Phi}^\top Q_{f} \Bar{\Phi})$. Since the gPC projection is a canonical transformation, we can prove that the projected matrix $Q_{\gpc}$ is positive definite.
\begin{prop}\label{prop:positive_definite_cost}
The expectation matrix $\Exp(\Bar{\Phi}^\top \Bar{\Phi})$ is a positive definite matrix.
\end{prop}
\begin{proof}
 We can prove the following equality by expanding the matrix multiplication. 
 \begin{equation}
     \Exp(\Bar{\Phi}^\top \Bar{\Phi}) = \mathbb{I} \otimes \Exp(\Phi \Phi^\top) \label{eq: }
 \end{equation}
 The block matrix $\Exp(\Phi \Phi^\top)$ is positive definite as the functions $\phi_i$ used to construct the column vector $\Phi$ are orthogonal with respect to the density function $\rho$. Therefore, $\Exp(\bar{\Phi}^\top \bar{\Phi})$ is positive definite, since $\Exp(\Phi \Phi^\top)$ is positive definite.
\end{proof}
\begin{lemma}\label{lemma:cost_gpc}
If $Q$ is a positive definite matrix, then the expectation $Q_{\gpc} = \Exp(\Bar{\Phi}^\top Q \Bar{\Phi})$ is a positive definite matrix.
\end{lemma} 
\begin{proof}
Since $Q$ is a positive definite matrix, we have $Q \succcurlyeq \lambda_{\min} (Q)\mathbb{I}$ where $\lambda_{\min} (Q) >0$. The expectation $\Exp(\bar{\Phi}^\top Q \Bar{\Phi})$ can be lower bounded as follows:
$
\Exp(\bar{\Phi}^\top Q \Bar{\Phi}) \succcurlyeq \Exp(\bar{\Phi}^\top \lambda_{\min} (Q) \mathbb{I} \Bar{\Phi})
 \succcurlyeq \lambda_{\min} (Q). \Exp(\bar{\Phi}^\top\Bar{\Phi})$. 
Using~\cref{prop:positive_definite_cost}, we conclude that $\Exp(\bar{\Phi}^\top Q \Bar{\Phi})$ is a positive definite matrix. 
\end{proof}
\changed{\begin{corollary}
\label{coroll:cost_gpc}
If the polynomials $\phi_{j}$ used for gPC projection are Hermite polynomials, then $\Exp(\bar{\Phi}^\top Q \Bar{\Phi}) \succcurlyeq \lambda_{\min}(Q) \mathbb{I}$.
\end{corollary}
\begin{proof} 
From~\cref{lemma:cost_gpc}, we have $\Exp(\bar{\Phi}^\top Q\bar{\Phi}) \succcurlyeq \lambda_{\min}(Q) \Exp(\bar{\Phi}^\top \bar{\Phi}).$ Using the gPC expansion in~\cref{eq:state_kronck_notation}, we have
\begin{equation}
\begin{aligned}
     \lambda_{\min}(Q)& \Exp(\bar{\Phi}^\top  \bar{\Phi}) \\ &= \lambda_{\min}(Q) \Exp(\left(\mathbb{I} \otimes \Phi(\xi) \right)\left(\mathbb{I} \otimes \Phi(\xi)\right)^\top) \\
    & \succcurlyeq \lambda_{\min}(Q) \mathbb{I} \Big ( \min \{\Exp(\phi^2_0),\Exp(\phi^2_1),\dots,\Exp(\phi^2_{d_x})\} \Big)
\end{aligned} \nonumber
\end{equation}
For the Hermite polynomials~\cite{xiu2009fast}, $ \min \{\Exp(\phi^2_0),$ $\Exp(\phi^2_1),\dots,\Exp(\phi^2_{d_x})\} = \Exp(\phi^2_0) = 1 $. Hence, we have the inequality $\Exp(\bar{\Phi}^\top Q \Bar{\Phi}) \succcurlyeq \lambda_{\min}(Q) \mathbb{I}$.
\end{proof}
\cref{lemma:cost_gpc} and \cref{coroll:cost_gpc} prove that the projected cost functional in the gPC space is positive-definite. In the following example, we illustrate the projection of a quadratic cost function using the gPC method.}
\changed{\begin{example}\label{example:quadratic_cost}
Consider the quadratic cost, $J = \Exp(x q_0 x)$, where $x,q_0 \in \real^1$ and $q_0$ and $x$ is a random variable. Let us represent the the random variable $x$ using the Hermite polynomials with single uncertainty $\xi \sim \mathcal{N}(0,1)$ upto order 2 as follows:
\begin{equation}
    x = x_0 + x_1\xi + +x_2 (\xi^2 - 1).\label{eq:example_x_expansion}
\end{equation}
The projected cost $J_\gpc$ in terms of the gPC state $X = [x_0, x_1, x_2]^\top$ is given as follows:
\begin{equation}
\begin{aligned}
    \Exp(x q_0 x) & = X^\top  \Exp \Big([1,\xi,\xi^2-1]^\top q_0 [1,\xi,\xi^2-1] \Big) X \\
    & = X^\top \begin{bmatrix}
    q_0 \Exp(1) & \Exp(\xi) & \Exp(\xi^2-1) \\ \Exp(\xi) & \Exp(\xi^2) & \Exp(\xi(\xi^2-1)) \\ \Exp(\xi^2-1) & \Exp(\xi(\xi^2-1)) & \Exp((\xi^2-1)^2)
    \end{bmatrix} X
\end{aligned} \nonumber
\end{equation} 
The expectation operation is with respect to a normal probability distribution. Additionally, due to the orthogonality of the Hermite polynomials we have only diagonal terms. The projected cost is of the following quadratic form. 
\begin{equation}
    \Exp(x q_0 x)  = X^\top \Big[\begin{smallmatrix}
    q_0 & 0 & 0 \\ 0 & q_0 & 0 \\ 0 & 0 & 2q_0
    \end{smallmatrix} \Big] X
\end{equation}
Note that the projected cost is also quadratic and positive definite as proved in~\cref{lemma:cost_gpc} and~\cref{prop:positive_definite_cost}.
\end{example}}
\subsection{Convex Approximation of the Chance Constraint}
The deterministic approximations of the chance constraints discussed in~\cref{subsec:stoc_opt_prob} are expressed in terms of the gPC coefficients that define a feasible set for the deterministic optimal control problem with gPC coefficients as decision variables.
\begin{lemma}\label{lemma:lin_socc_gpc}
The second-order cone constraint given below
\begin{align}
  (a^\top \otimes M) \gpcX + b + \sqrt{\frac{1-\epsilon}{\epsilon}}\sqrt{\gpcX^\top U N N^\top U^\top \gpcX} \leq 0
    \label{eq:socc_gpc}
\end{align}
is equivalent to the deterministic approximation of the DRLCC in~\cref{eq:drcc_lin} as $\ell \to \infty.$, where the matrices $M,U,N$ are given by 
\begin{equation}
\begin{aligned}
    M & = \begin{bmatrix}1 & 0 & \cdots & 0 \end{bmatrix}_{1 \times (\ell+1)} \\
     U & = \begin{bmatrix}
    a_{1} &0 & 0\\
    0& \ddots & 0\\
    0&0 & a_{d_{x}}
  \end{bmatrix} \otimes \mathbb{I}_{(\ell+1)\times(\ell+1)} \\ 
   N & = \mathds{1}_{d_{x}\times d_{x}} \otimes \mathds{H} ; \ \mathds{H}  = \begin{bmatrix}
    0 & \mathbb{O} \\
    \mathbb{O} & \sqrt{\Exp(HH^\top)} 
  \end{bmatrix} \\ 
  \text{where} \ & H = \begin{bmatrix} \phi_{1}(\xi) & \cdots & \phi_{\ell}(\xi)\end{bmatrix}^\top
\end{aligned} \label{eq:mat_lin_con_gpc}
\end{equation} and $\mathds{1}$ is a matrix with entries as $1$.
\end{lemma}
\begin{proof} It is sufficient to prove that $(a^\top \otimes M) \approx a^\top \mu_x$ and $\gpcX^\top U NN^\top U^\top \gpcX \approx a^\top \Sigma_{x} a $ as $\ell \to \infty$. Invoking Lemma~\ref{lemma:conv_gpc} and Remark~\ref{rmrk:exp_var_gpc}, the polynomials of gPC coefficients can be replaced by mean and variable of the variable $x$.  
\begin{align}
    \begin{aligned}
    (a^\top \otimes M) \gpcX& =  \begin{bmatrix} a_{1}M & a_{2}M&\cdots &a_{d_{x}}M \end{bmatrix}\gpcX \\
    & = a_{1}x_{10} + a_{2}x_{20} +\cdots +a_{d_{x}}x_{dx0} \\
    & \approx a^\top \mu_x
    \end{aligned}\label{eq:mean_eq}
\end{align}
\noindent Equation~(\ref{eq:mean_eq}) shows the steps involved to prove $(a^\top \otimes M) \approx a^\top \mu_x$. Let us define a vector $p_{i} = \begin{bmatrix}x_{i0} & \Bar{p}^\top_{i} \end{bmatrix}^\top$ where $\Bar{p}_{i} = \begin{bmatrix} x_{i1} & \cdots & x_{i\ell} \end{bmatrix}^\top$. \begin{equation}
    U^\top \gpcX = \begin{bmatrix} a_{1}p^\top_{1} & a_{2}p^\top_{2} & \cdots & a_{d_{x}}p^\top_{d_{x}} \end{bmatrix}^{\top} \label{eq:expnd1}
\end{equation}
\begin{align}
    NN^\top U^\top \gpcX =& \begin{bmatrix} \mathds{H} a_{1}p_{1}& \mathds{H} a_{2} p_{2}& \cdots & \mathds{H} a_{d_{x}}p_{d_{x}}\end{bmatrix}\label{eq:expnd2}\\
   \hspace{-30pt}\gpcX^\top U NN^\top U^\top X & = \sum_{i = 1}^{d_{x}}\sum_{j = 1}^{d_{x}} a_{i}a_{j}p^\top_{i}\mathds{H}p_{j} \label{eq:expnd3}\\ &= \sum_{i = 1}^{d_{x}}\sum_{j = 1}^{d_{x}} a_{i}a_{j}\Bar{p}^\top_{i}\Exp (HH^\top)\Bar{p}_{j} \approx a^\top \Sigma_{x} a \nonumber
\end{align}
Using this notation, the matrices in~(\ref{eq:socc_gpc}) are expanded as shown in~(\ref{eq:expnd1}),~(\ref{eq:expnd2}), and~(\ref{eq:expnd3}). Therefore, the equivalence is proved by Lemma~\ref{lemma:conv_gpc} as $\ell \to \infty$. \end{proof}
In the following example, we show the projection of a one-dimensional linear constraint.
\changed{\begin{example}\label{example:}
Consider the linear chance constraint,$\Prob(ax+b\leq0) \geq 1-\epsilon$,
where $a,b \in \real^1$ and $x\in \real^1$ is a random variable. Let us express $x$ using Hermite polynomials upto order 2, $x= [1,\xi,\xi^2-1]X$, where $\xi \sim \mathcal{N}(0,1)$ and $X = [x_0, x_1, x_2]^\top$. We can express the above linear chance constraint as an inequality constraint using~\cref{lemma:lin_consev_approx,lemma:lin_socc_gpc}, and the matrices $M, N, U$ defined in~\cref{eq:socc_gpc} are computed as:
\begin{equation}
\begin{aligned}
  & a^\top \otimes M = a \begin{bmatrix}1 & 0 &0\end{bmatrix},\ \ U = a \otimes \mathbb{I}_{3\times 3},\\& N = 1 \otimes \begin{bmatrix}
  0 & 0 & 0\\ 0 & \sqrt{\Exp(\xi^2)} & 0 \\ 0 & 0 & \sqrt{\Exp((\xi^2-1)^2)}
  \end{bmatrix} = \begin{bmatrix}
  0 & 0 & 0\\ 0 & 1 & 0 \\ 0 & 0 & \sqrt{2}
  \end{bmatrix}.
\end{aligned}\nonumber
\end{equation}
Then, the deterministic constraint \cref{eq:socc_gpc} becomes
\begin{equation}
\begin{aligned}
    & [a,0,0] X + b + \sqrt{\tfrac{1-\epsilon}{\epsilon}} \sqrt{X^\top \Big[\begin{smallmatrix}
  0 & 0 & 0\\ 0 & 1 & 0 \\ 0 & 0 & 2\end{smallmatrix}\Big] X } \leq 0.
\end{aligned}
\end{equation}
The second-order cone constraint enables the use of convex programming, where the deterministic constraint $a\mu_x + b + \sqrt{ a\sigma_x a} \leq 0$ is non-convex in terms of the mean $\mu_x$ and variance $\sigma_x$ of the random variable $x$. 
\end{example}}
\begin{lemma} \label{lemma:sdp_con_gpc} The quadratic inequality
\begin{equation}
    \sum_{i = 1}^{d_{x}} \sum_{k = 1}^{\ell} a_{i}\langle \phi_{k},\phi_{k}\rangle x_{ik}^2 \leq \epsilon c,
    \label{eq:sdp_con_gpc}
\end{equation}
expressed in terms of the gPC coefficients is equivalent to the constraint in~\cref{eq:sdp_constraint} as $\ell \to \infty$, where $A$ is a diagonal matrix with $i^\mathrm{th}$ diagonal element as $a_{i}$ and $\langle\phi_k,\phi_k\rangle=\int_{\mathbb{D}}\boldsymbol{\rho}(\xi)\phi_k\phi_k d\xi$. 
\end{lemma}
\begin{proof}The deterministic approximation, $\Trace(A\Sigma_{x}) \leq c \epsilon$, of the QCC in~\cref{eq:quad_chnc_cons} can be expanded as follows.
\begin{equation}
    \begin{aligned}
    \Trace(A\Sigma_{x}) \leq c \epsilon & \equiv \sum_{i = 1}^{d_{x}} a_{i} \Exp((x_{i}-\mu_{x_i})^\top (x_{i}-\mu_{x_i})) \leq c \epsilon\\
    & \equiv \sum_{i = 1}^{d_{x}} \sum_{j = 1}^{\ell}  a_{i} \langle\phi_{j},\phi_{j} \rangle x^2_{ij}  \leq c \epsilon
    \end{aligned}\label{eq:proof_sdp}
\end{equation}The equivalence is proved by directly expanding the trace and using Remark~\ref{rmrk:exp_var_gpc} as shown in~(\ref{eq:proof_sdp}).\end{proof}
\begin{figure}
    \centering
    \includegraphics[width=1.\columnwidth,height=2.5in]{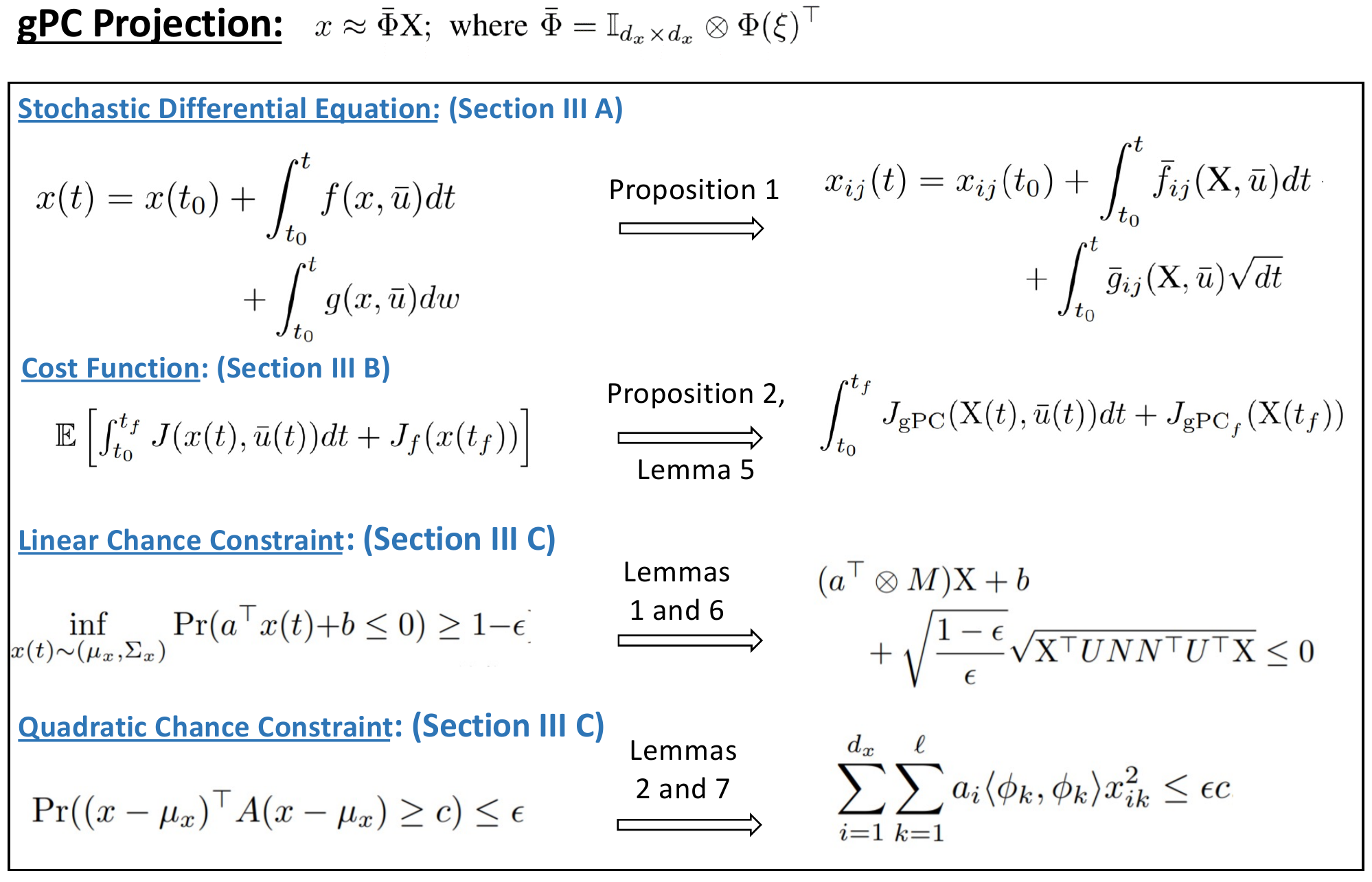}
    \caption{Illustration of the gPC projection method. We use the gPC projection method to derive a deterministic surrogate of the chance-constrained optimal control problem. We use Lemmas 1 to 7 and Propositions 1 and 2 to prove Theorem 1.}
    \label{fig:projection_lemmas}
\end{figure}
\changed{
\begin{example}
Consider the one-dimensional quadratic constraint $\Prob(a(x-\mu_x)^2\leq c) \geq1-\epsilon$. where $x \in \real^1$ is a random variable with mean $\mu_x$, $a, c \in \real^1$. Let us express $x$ using Hermite polynomials up to order 2, $x= [1,\xi,\xi^2-1]X$, where $\xi \sim \mathcal{N}(0,1)$ and $X = [x_0, x_1, x_2]^\top$. The quadratic chance constraint in terms of $\sigma_x$ can be expressed in $X$ as follows using~\cref{lemma:quad_const,lemma:sdp_con_gpc}:
\begin{equation}
    \begin{aligned}
      \Trace(a\sigma_x) &= a \Exp( 1x^2_0 + \xi^2 x^2_1 + (\xi^2 -1)^2 x^2_2 )\\
      & = a x^2_0 + a x^2_1 + 2a x^2_2 \leq c\epsilon .
    \end{aligned}
\end{equation}
The semidefinite constraint $\Trace(a \sigma_x)$ is transformed into a quadratic constraint in the gPC state $X$.
\end{example}}
Using the projected dynamics~\cref{eq:continuous_ode_form}, the cost functional in~\cref{eq:cost_odeopt}, and the linear and quadratic chance constraints~\cref{eq:socc_gpc},~\cref{eq:sdp_con_gpc} in the gPC coefficients (as shown in~\cref{fig:projection_lemmas}) we can formulate the following distributionally robust deterministic nonlinear optimal control problem with the gPC states $\gpcX$ and the control $\bar{u}$ as decision variables.
\begin{problem}
\label{prob:dr_det_optimal_cntrl}
Distributionally Robust Deterministic Nonlinear Optimal Control Problem.
\begin{equation*}
\begin{aligned}
J_{\gpc}^{*} =  & \underset{\gpc\gpcX(t),\Bar{u}(t)}{\min}
& &  \int_{t_{0}}^{t_{f}}J_{\gpc}(\gpcX(t),\Bar{u}(t))dt + J_{\gpc_{f}}(\gpcX(t_{f})) \\
& \text{s.t.}
& & \cref{eq:odefull_form},~\cref{eq:socc_gpc},~\cref{eq:sdp_con_gpc}\\
&  & & \Bar{u}(t) \in \setU \quad \forall t \in [t_{0},t_{f}]\\
&  & & \gpcX(t_{0}) = \gpcX_{0} \quad \gpcX(t_{f}) \in \setX_{\gpcX_{f}}
\end{aligned}
\end{equation*}
\end{problem}
\noindent where the projection of the initial condition $x_{0}$ in the gPC space is $\gpcX_{0}$. The terminal set $\setX_{\gpcX_{f}}$ is constructed using a distributionally robust polytope (or) a conservative ellipsoid approximation of the set $x_{f}$ in~\cref{prob:cc_stoc_opt_cntrl} with probabilistic guarantees using \cref{lemma:lin_socc_gpc,lemma:sdp_con_gpc}. We make the following observations about the distributional robustness and gPC projection discussed above to transform~\cref{prob:cc_stoc_opt_cntrl} to~\cref{prob:dr_det_optimal_cntrl}.
\begin{itemize}
    \item The infinite-dimensional optimal control problem in state space and time, as described in~\cref{prob:cc_stoc_opt_cntrl}, is projected to~\cref{prob:dr_det_optimal_cntrl} that is finite-dimensional in space and infinite-dimensional in time. 
    \item The ODE approximation of the SDE using the gPC projection diverges over a long-horizon problem (or) when the uncertainty affecting the SDE has large variance (or) when the uncertainty model has large gradients with respect to state and control. A multi-element gPC method can be used to overcome this divergence due to the finite-dimensional approximation. The structure of the proposed constraint reformulation is invariant to the multi-element gPC method.
    \item The choice of the terminal set used in~\cref{prob:cc_stoc_opt_cntrl} is restricted due to the $\epsilon_{c}-$controllability of the SDE. We use slack variables on the terminal state variance to ensure the feasibility of both~\cref{prob:cc_stoc_opt_cntrl,prob:dr_det_optimal_cntrl}.
    \item The projected cost functional preserves the positive definite property of the quadratic cost used in~\cref{prob:cc_stoc_opt_cntrl}.
    \item \changed{For a given risk measure, the linear and quadratic chance constraints are second-order cone and semidefinite constraints in the gPC coefficients, respectively.}  
\end{itemize}
\Cref{prob:dr_det_optimal_cntrl} (DNOC) enables the use of techniques such as the pseudospectral method and sequential convex programming for solving \cref{prob:dr_det_optimal_cntrl} (SNOC). We use sequential convex programming to solve~\cref{prob:dr_det_optimal_cntrl} and apply this technique to compute safe and optimal motion plans under uncertainty. We extend the gPC-SCP to formulate a predictor-corrector formulation for fast real-time planning under uncertainty.
\subsection{gPC-SCP: Generalized Polynomial Chaos-Based Sequential Convex Programming}\label{subsec:gPC-SCP-scp}
We formulate the gPC-SCP problem by constructing a sequential convex programming (SCP) approximation of~\cref{prob:dr_det_optimal_cntrl} with gPC state ($\gpcX$) and control ($\bar{u}$) as decision variables. The convex program is then solved iteratively using an interior point method until a convergence criterion is satisfied and projected back to the probability space from the gPC space to compute a solution of~\cref{prob:dr_stoc_opt_cntrl}.  

The SCP problem formulation involves two steps: 1) Discretizing the continuous-time optimal control problem into a discrete-time optimal control problem, and 2) Convexifing the non-convex constraints and cost function about a nominal initial state and control trajectory. Following this approach, the projected integral cost functional~\cref{eq:cost_odeopt}, the nonlinear dynamics~\cref{eq:odefull_form}, and the second-order cone constraint~\cref{eq:socc_gpc} and the semi-definite constraint~\cref{eq:sdp_con_gpc} are discretized using a first-order hold approach for $T$ time steps between the time horizon $[t_0,t_f]$ with gPC state and control as decision variables. 

At iteration $i$, the cost functional, constraints~\cref{eq:socc_gpc} and~\cref{eq:sdp_con_gpc}, and feasible control set $\setU$ are convex. The discretized gPC dynamics in~\cref{eq:odefull_form} is a nonlinear equality constraint at each time step. We convexify the nonlinear dynamics~\cref{eq:odefull_form} by linearizing it about the state and the control trajectory $S^{(i-1)} = \{\gpcX^{(i-1)},\bar{u}^{(i-1)}\}$ computed at $(i-1)^{th}$ iteration. The linearized equations form a set of linear constraints on the state and control action as follows.
\begin{equation}
\begin{aligned}
    \gpcX^{(i)}[k+1] & = \gpcX^{(i)}[k] + A^{(i)}[k] \gpcX^{(i)}[k] + B^{(i)}[k] \Bar{u}^{(i)}[k]\\ & + Z^{(i)}[k], \quad \mathrm{where:} \ k \in \{1,\dots,T-1\}.
\end{aligned}\label{eq:linearize_discrete_dyn}
\end{equation}
\begin{align}
    A^{(i)}& = \frac{\partial(\Bar{f} + \Bar{g})}{\partial \gpcX} \Big|_{\scalebox{0.85}{$S^{(i-1)}$}}; \quad B^{(i)} = \frac{\partial(\Bar{f} + \Bar{g})}{\partial \Bar{u}} \Big|_{\scalebox{0.85}{$S^{(i-1)}$}} \nonumber
\end{align}
\begin{align}
    Z^{(i)} & = \Bar{f}(S^{(i-1)},\Delta t)+\Bar{g}(S^{(i-1)},\sqrt{\Delta t}) \nonumber\\ & -A^{(i)} \gpcX^{(i-1)}- B^{(i)}\Bar{u}^{(i-1)}\label{eq:linear_mat}
\end{align}
The gPC-SCP problem at iteration $i$, after discretization and convexification, is given below in~\cref{prob:dr_trajopt}.
\begin{problem}
\label{prob:dr_trajopt}
gPC-SCP: Generalized Polynomial Chaos-based Sequential Convex Programming.
\begin{align}
& \underset{\gpcX^{(i)},\Bar{u}^{(i)}}{\min}
& &\hspace{-10pt}\sum_{k = 1}^{T-1} J_{\gpc}(\gpcX^{(i)}[k],\Bar{u}^{(i)}[k])\Delta t + J_{\gpc_{f}}(\gpcX^{(i)}[T])  \nonumber \\
& \textit{s.t.}& &  \mathrm{Projected \ Dynamics: } \cref{eq:linearize_discrete_dyn} \nonumber\\
& & & \mathrm{Constraints:} ~\{\cref{eq:socc_gpc},~\cref{eq:sdp_con_gpc}\} \nonumber\\
&  & &  \Bar{u}^{(i)}[k] \in \setU \quad \forall \ k \in \{1,\dots,T-1\} \nonumber\\
&  & &  \gpcX^{(i)}[1] = \gpcX_{0} \quad \gpcX^{(i)}[T] \in \setX_{\gpcX_{f}} \nonumber\\ 
&  & & \hspace{-30pt}\|\gpcX^{(i)}[k]-\gpcX^{(i-1)}[k]\|^2 _{2} \leq \alpha_{x}\beta \ \forall \ k \in \{1,\dots,T\} \label{eq:state_trust}\\
&  & &  \hspace{-30pt}\|\baru^{(i)}[k]-\baru^{(i-1)}[k]\|^2 _{2} \leq \alpha_{u}\beta \ \forall \ k \in \{1,\dots,T-1\}\label{eq:control_trust}
\end{align}
\end{problem}
\cref{prob:dr_trajopt} shows the SCP formulation at $i^{\mathrm{th}}$, given a nominal trajectory $S^{(i-1)} = \{\gpcX^{(i-1)},\bar{u}^{(i-1)}\}$ computed at $(i-1)^\mathrm{th}$ iteration with the constraint set at each time step $k$ and at iteration $i$, where $\setX_{\gpcX_{f}}$ is the projected terminal constraint. The nominal trajectory $S^{0}$ at $i=1$, used to initialize gPC-SCP, is computed using a deterministic trajectory optimization for the nominal dynamics $\dot{x} = f(x,\bar{u})$, which ignores the uncertainty that affects the system. For the motion planning problem, the nominal trajectory $S^{0}$ is computed using kinodynamic motion planning algorithms such as asymptotically optimal and rapid exploration of random trees~\cite{hauser2016}.

An additional trust region constraint on the gPC state~\cref{eq:state_trust} and control~\cref{eq:control_trust} is used to ensure the convergence and feasibility of the SCP as $i \to \infty$, where $\alpha_x >0$, $\alpha_x >0$, and $\beta \in ( 0, 1)$. The choice of $\beta$  ensures the convergence of the trust region as the number of iterations increases. This acts as a convergence criterion, while ensuring that the search space is small.  The trust region in the gPC state in~\cref{eq:state_trust} can be equivalently understood as a probabilistic constraint of the form $\Prob(\|x^{i}-x^{i-1}\| \leq \alpha_{xp}) \geq 1-\epsilon_{t}$, where $\alpha_{x}$ is a function of $\alpha_{xp}$ using the quadratic projection discussed in~\cref{lemma:sdp_con_gpc}. The SCP algorithm is known to converge to the KKT point of the DNOC problem under mild conditions. For a detailed analysis of convergence, see~\cite{morgan2016swarm,morgan2014model,morgan2012spacecraft}. 
We ensure the feasibility of gPC-SCP: 1) by using stochastic reachable terminal sets, as discussed in~\cite{hewing2018stochastic}, that are constructed using the linearized approximation of the dynamics, and 2) by increasing the trust region in loop with $\beta>1$ when infeasibility occurs. 
\subsection{Sub-Optimality and Convergence}
\label{sec:finite_dimensional_approximation}
In this subsection, we study the optimality of~\cref{prob:dr_det_optimal_cntrl} and show that~\cref{prob:dr_det_optimal_cntrl} computes a sub-optimal solution to~\cref{prob:cc_stoc_opt_cntrl}. We make a two-step approximation of~\cref{prob:cc_stoc_opt_cntrl} by using distributional robustness to formulate~\cref{prob:dr_stoc_opt_cntrl} with known mean and variance of the state and then use gPC propagation to construct the deterministic optimal control~\cref{prob:dr_det_optimal_cntrl} that is solved using SCP. In~\cref{lemma:sub_optimal_dr_snoc}, we prove the sub-optimality of the optimal cost $J^{*}_{\mathrm{SNOC}}$ of~\cref{prob:cc_stoc_opt_cntrl} compared to the optimal cost $J^{*}_{\mathrm{DR-SNOC}}$ of \cref{prob:dr_stoc_opt_cntrl} that is distributionally robust.
\begin{lemma}\label{lemma:sub_optimal_dr_snoc}
The optimal solution of~\cref{prob:dr_stoc_opt_cntrl} is a sub-optimal solution of~\cref{prob:cc_stoc_opt_cntrl}, i.e., $J^{*}_{\mathrm{SNOC}} \leq J^{*}_{\mathrm{DR-SNOC}}$. 
\end{lemma}
\begin{proof}
The constraint set $\setX_{\mathrm{DRLCC}}$ and $\setX_{\mathrm{CQCC}}$ in~\cref{prob:dr_stoc_opt_cntrl} are a subset of the constraint set $\setX_{\mathrm{LCC}}$ and $\setX_{\mathrm{QCC}}$ of~\cref{prob:cc_stoc_opt_cntrl} respectively. Therefore, $J^{*}_{\mathrm{SNOC}} \leq J^{*}_{\mathrm{DR-SNOC}}$ as the feasible space of~\cref{prob:cc_stoc_opt_cntrl} is larger than the feasible space of~\cref{prob:dr_stoc_opt_cntrl}.
\end{proof}
\Cref{prob:dr_det_optimal_cntrl} (DNOC) computed by the gPC projection converges asymptotically to \Cref{prob:dr_stoc_opt_cntrl} (SNOC). The following theorems discuss the conditions for convergence.
\begin{theorem} \label{theorem:convergence_to_true_optimal}
The surrogate deterministic nonlinear optimal control~\cref{prob:dr_det_optimal_cntrl} with convex constraints
is a sub-optimal surrogate for the stochastic nonlinear optimal control~\cref{prob:cc_stoc_opt_cntrl} with the following being true:\\
\textbf{(a)}  In the case without chance constraints, the cost $|J_{\gpc}^{*} - J^*| \to 0$ as $\ell \to \infty$\\
\textbf{(b)} In the case with linear and quadratic chance constraints, any feasible solution of~\cref{prob:dr_det_optimal_cntrl} is a feasible solution of~\cref{prob:cc_stoc_opt_cntrl} as $\ell \to \infty$ and $J^{*}_{\mathrm{SNOC}} \leq J^{*}_{\gpc}$, assuming that a feasible solution exists.
\end{theorem}
\begin{proof}\textbf{Case (a)}: It is sufficient to prove that the cost function and the dynamics are exact as $\ell \to \infty$. Using the Kronecker product notation, due to \cref{lemma:conv_gpc}, we have the following 
\begin{equation}
    \| x - \bar{\Phi} \gpcX\|_{\mathcal{L}_2} \to 0 \ \text{as} \ \ell \to \infty
    \label{eq:conv_kon_not}
\end{equation}
\begin{equation}
    (\ref{eq:conv_kon_not}) \implies\ \| \dot{x} \to \bar{\Phi}\dot{\gpcX} \|_{\mathcal{L}_2}\to 0 \ \text{as} \ \ell \to \infty
     \label{eq:dyn_equi}
\end{equation}
\begin{equation}
    \begin{aligned}
    (\ref{eq:conv_kon_not}) \implies \ & |J_{\gpc} - J | \to 0 \ \text{as} \ \ell \to \infty \\
    & |J_{\gpc_{f}} - J_{f}  | \to 0 \ \text{as} \ \ell \to \infty
    \end{aligned} \label{eq:cost_equi}
\end{equation}
From~\cref{eq:dyn_equi}, and~\cref{eq:cost_equi} we conclude that the optimal value $|J_{\gpc}^* - J^*| \to 0$ as $\ell \to \infty$, since the cost function, the dynamics, and the initial and terminal conditions converge to the original stochastic formulation (\cref{prob:cc_stoc_opt_cntrl}) as $\ell \to \infty$.\\
\noindent \textbf{Case (b):} Consider the sets $\setX_\mathrm{LgPC}$, and $\setX_\mathrm{QgPC}$ defined below. 
\begin{equation}
    \setX_\mathrm{LgPC} = \{ x \in \setX: x \approx \bar{\Phi}\gpcX \ \text{where} \ \gpcX \ \mathrm{satisfies}~(\ref{eq:socc_gpc}) \}
    \label{eq:set_lin_gpc}
\end{equation}
\begin{equation}
    \setX_\mathrm{QgPC} = \{ x \in \setX: x \approx \bar{\Phi}\gpcX \ \text{where} \ \gpcX \ \mathrm{satisfies}~(\ref{eq:sdp_con_gpc}) \}
    \label{eq:set_sdp_gpc}
\end{equation}
Using \cref{lemma:lin_socc_gpc,lemma:sdp_con_gpc}, we see that the approximate convex constraints converge to the deterministic equivalent of the distributionally robust chance constraint as $\ell \to \infty$.
\begin{equation}
    \begin{aligned}
    \cref{lemma:lin_socc_gpc} & \implies \setX_\mathrm{LgPC} \to \setX_\mathrm{DRLCC} \ \text{as} \ \ell \to \infty  \\
    \cref{lemma:sdp_con_gpc} & \implies \setX_\mathrm{QgPC} \to \setX_\mathrm{CQCC} \ \text{as} \ \ell \to \infty
    \end{aligned} \label{eq:pp1}
\end{equation}
Using \cref{lemma:lin_consev_approx,lemma:quad_const}, we have the following:
\begin{equation}
    \begin{aligned}
    \cref{lemma:lin_consev_approx} & \implies \setX_\mathrm{DRLCC} \subseteq \setX_\mathrm{LCC}\\ 
    \cref{lemma:quad_const} & \implies \setX_\mathrm{CQCC} \subseteq \setX_\mathrm{QCC}\\
    \end{aligned}\label{eq:pp2}
\end{equation}
\begin{equation}
    \begin{aligned}
    \{(\ref{eq:pp1}),~(\ref{eq:pp2})\} \implies 
\begin{cases}
\setX_\mathrm{LgPC} \subseteq \setX_\mathrm{LCC} \ \text{as} \ \ell \to \infty  \\
\setX_\mathrm{QgPC} \subseteq \setX_\mathrm{QCC} \ \text{as} \ \ell \to \infty
\end{cases}
    \end{aligned} \label{eq:proof_caseb_thm1}
\end{equation}
Combining~(\ref{eq:pp1}) and~(\ref{eq:pp2}), we can conclude that~(\ref{eq:proof_caseb_thm1}) holds as $\ell \to \infty$. In \cref{fig:projection_lemmas}, we illustrate Lemmas 1, 2, 6 and 7 used in proving this theorem. This proves that if there is a feasible solution exists for~\cref{prob:dr_det_optimal_cntrl} then it is a feasible solution of~\cref{prob:cc_stoc_opt_cntrl} as $\ell \to \infty$. Using~\cref{lemma:sub_optimal_dr_snoc}, as $\ell \to \infty$ we have $J^{*}_{\gpc} \to J^{*}_{\mathrm{DR-SNOC}}$. This implies that $J^{*}_{\mathrm{SNOC}} \leq J^{*}_{\mathrm{\gpc}}.$ 
\end{proof}
Theorem~\ref{theorem:convergence_to_true_optimal} proves the consistency of the gPC projection method as $\ell \to \infty$. The asymptotic convergence of the cost and chance constraints is achieved with a large number of polynomials $\phi$. This leads to a deterministic optimal control problem of size $\ell d_{x}$. The choice of $\ell$ depends on the number of uncertainties in the system and the nature of the state distribution. A computationally efficient approach is to use $P_{\gpc} = 2$ to generate the functions $\phi$ used in the projection. This replicates the computational efficiency of linear covariance propagation techniques, while ensuring the convexity of the chance constraints in the gPC space. We study the chance constraint formulation for collision checking under uncertainty in dynamics and obstacle locations in the following section using distributional robustness and gPC projection.

\section{Motion Planning Under Uncertainty}\label{sec:motion_planning_gpc_scp}
The gPC-SCP method is applied to plan a safe trajectory under uncertain obstacles and stochastic nonlinear dynamics. We formulate the motion planning problem to incorporate uncertainty in the dynamics and then show that deterministic projections of chance constraints in the gPC space enable convex formulations of the collision constraint for obstacles with both deterministic and stochastic state models. 

The motion planning problem is to compute an optimal and safe trajectory ($x \in \setX_{\setF}$) for the SDE in~\cref{eq:nl_stochastic_dynamics} from an initial state $x_{0}\in\setX$ to the terminal set $\setX_{f}\subseteq\setX$ on a given map with static obstacles. In the following, we derive a chance constraint formulation of the collision constraint and the terminal-state constraint. The chance constraints are used to formulate an SNOC problem as described in~\cref{prob:cc_stoc_opt_cntrl}. The SNOC problem is then projected into the gPC space for solving via the SCP method. At each SCP iteration, the collision constraints are approximated as a linear chance constraint around the nominal trajectory and form a second-order cone constraint in the gPC state $\gpcX$ as discussed in~\cref{lemma:lin_socc_gpc}. The terminal set is defined as a chance constraint on an ellipsoidal set and forms a semidefinite constraint in the gPC states as discussed in~\cref{lemma:sdp_con_gpc}. 

In the following, we first discuss the linear chance constraint formulation for collision checking with a deterministic obstacle, and then extend it to include the uncertainty in obstacle locations for SCP. We prove that the approximation is a subset of the original nonlinear chance constraint. We then discuss the chance constraint formulation of the terminal set constraint. The chance constraint formulations for collision checking and terminal set are used to design the motion planning algorithm that integrates an asymptotically optimal sampling-based planner~\cite{hauser2016} with the gPC-SCP~\cref{prob:dr_trajopt} for computing a safe trajectory under uncertainty. 
\subsection{Collision Checking with Deterministic Obstacles} \label{subsec:det_obstacle}
We derive a second-order cone constraint approximation of the circular obstacle in the gPC coordinates under uncertainty in dynamics at any point in time $t \in [t_0,t_f]$. The approximation involves two steps. We first derive a conservative linear chance constraint approximation of the nonlinear collision chance constraint. In the second step, we project the linear chance constraint into a second-order cone constraint in the gPC coordinates. Let the state of the obstacle be $\barp_{\obs}$ at time $t$ and the radius of the obstacle be $r_{\obs}$. The collision chance constraint at any time $t$ for a robot with the state distribution $x$ and radius $r_{\rob}$ is given as follows:
\begin{equation}
\Prob\left( \|\matC(x - \barp_{\obs})\|_2 \geq r_{\rob} + r_{\obs}\right) \geq 1 - \epsilon_{\col},
    \label{eq:chance_constraint_det_obst}
\end{equation}
where the matrix $\matC$ is used to compute the position of the obstacle and the robot given the states $\bar{p}_{\obs}$ and $x$, respectively. The probability of collision is tuned using the risk measure $\epsilon_{\col} \in [0.001,0.1]$.
\begin{figure}[t]
    \centering
    \includegraphics[width=\columnwidth]{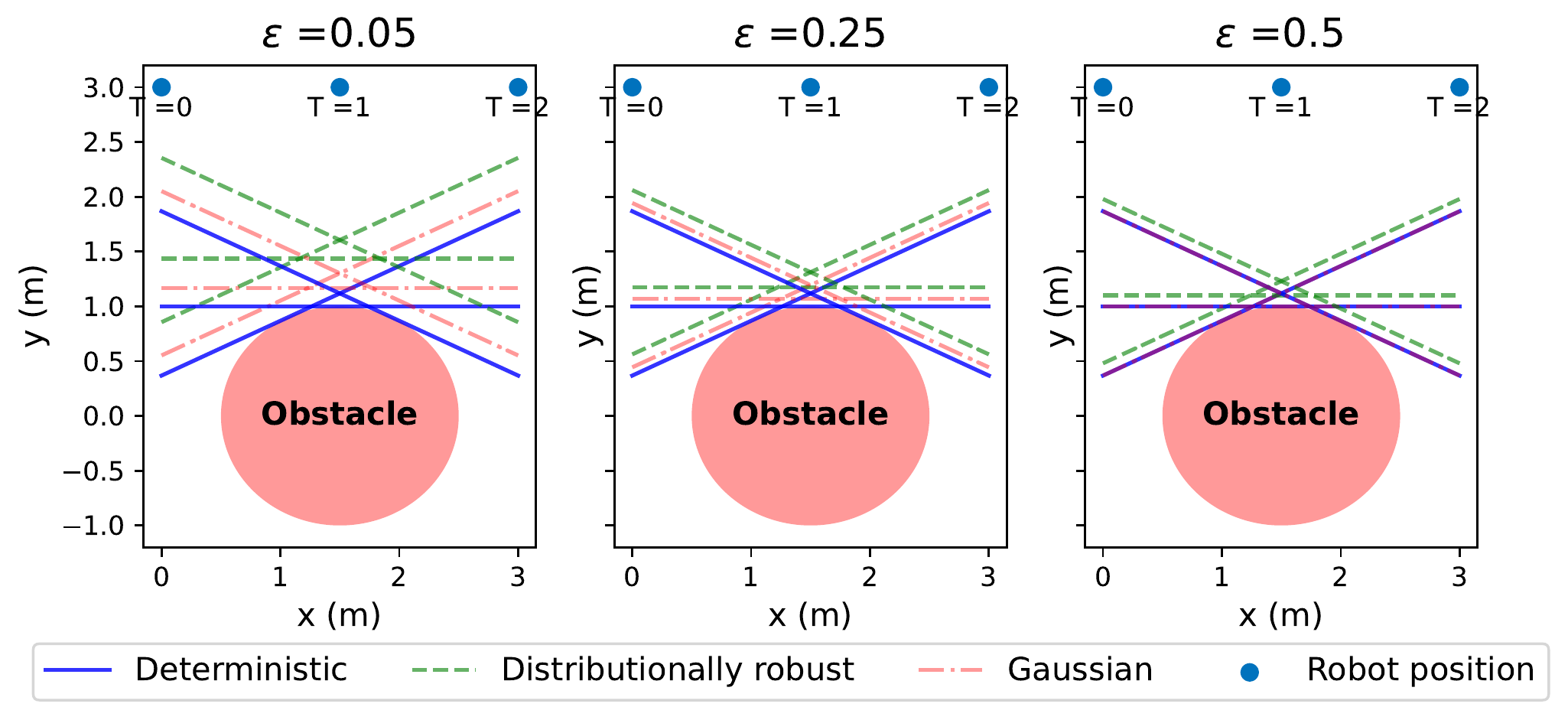}
    \caption{\changed{A comparison of hyperplane approximations for deterministic and stochastic obstacle models using distributional robust~\cref{eq:lin_socc} and Gaussian approximations~\cref{eq:gaussian_linear_constraint} at three different robot locations. The obstacle position follows a multivariate Gaussian distribution $p_{\obs} \sim \setN \left([1.5,0],\mathrm{diag\left([0.01,0.01]\right)} \right)$.}}
    \label{fig:hyperplane_comparison}
\end{figure}
In the following theorem, we prove that, given a nominal state distribution trajectory $x_{\nom}$, we can compute a conservative linear chance constraint approximation of the collision constraint~\eqref{eq:chance_constraint_det_obst}. The nominal trajectory $x_{\nom}$ is computed by using a deterministic planner without considering the uncertainty in dynamics.
\begin{lemma}\label{theorem:deterministic_obstacles}
The linear chance constraint,
\begin{align}
    \Prob\Big(&(\bar{x}_{\nom}- \bar{p}_{\obs})^\top \matC^\top \matC (x - \bar{p}_{\obs}) \nonumber \\ &\geq r_{\safe} \|\matC(\bar{x}_{\nom}- \bar{p}_{\obs})\|_{2} \Big) \geq 1 - \epsilon_{\col},
    \label{eq:lin_chance_constraint_det_obst}
\end{align}
in the robot state distribution, $x$ is a conservative approximation of the nonlinear collision chance constraint in~\eqref{eq:chance_constraint_det_obst} at any time $t \in [t_0, t_f]$, where $\bar{x}_{\nom}$ is a realization of the nominal state distribution $x_{\nom}$ of the robot that satisfies~\eqref{eq:chance_constraint_det_obst}, $\bar{p}_{\obs}$ is the state of the obstacle, and $r_{\safe} = r_{\rob} + r_{\obs}$.
\end{lemma}
\begin{proof}
Consider the set $\setX_{\free}$, defined as $\setX_{\free} = \{x:  \|\matC(x - \barp_{\obs})\|_2 \geq r_{\safe}\}$. Given a nominal trajectory $x_{\nom}$, the set $\setX_{s}$ defined as $\setX_{s} = \{x:(\bar{x}_{\nom}- \bar{p}_{\obs})^\top \matC^\top \matC (x - \bar{p}_{\obs}) \geq r_{\safe} \|\matC(\bar{x}_{\nom}- \bar{p}_{\obs})\|_{2}\}$ is such that $\setX_{s} \subseteq \setX_{\free}$. For a proof of $\setX_{s} \subseteq \setX_{\free}$ see~\cite{morgan2014model}. \Cref{fig:hyperplane_comparison} shows an example of the set $\setX_{s}$ and $\setX_{\free}$, where the hyperplane used to linearize the circular constraint leads to a reduced feasible space. We can construct an indicator function $\setI_{\free}(x)$ such that $\setI_{\free}(x) = 1$ if $x \in \setX_{\free}$ and $\setI_{\free}(x) = 0$ otherwise. Similarly, the indicator $\setI_s(x)$ is such that $\setI_{s}(x) = 1$ if $x \in \setX_{s}$ and $\setI_{s}(x) = 0$ otherwise. 
\begin{align}
& \ \textrm{Since} \ \setX_s \subseteq \setX_{\free}, \ x \in \setX_{s} \implies x \in \setX_{\free},  \nonumber\\
& \ \ \textrm{and} \ \setI_{s}(x) = 1 \implies \setI_{\free}(x) =1. \label{eq:theorem_det_obstacle_eq1}
\end{align}
Therefore, if $\Exp(\setI_{s}) \geq 1 - \epsilon_{\col}$, then $\Exp(\setI_{\free}) \geq 1 - \epsilon_{\col}$ with at least $1- \epsilon_{\col}$ probability. Note that $\Prob(x \in \setX_{\free}) = \Exp(\setI_{\free}(x))$ and $\Prob(x \in \setX_{s}) = \Exp(\setI_{s}(x))$. This implies that if the chance constraint in~\eqref{eq:lin_chance_constraint_det_obst} is satisfied with the probability $1- \epsilon_{\col}$, then the constraint in~\eqref{eq:chance_constraint_det_obst} is satisfied with at least a probability of $1 - \epsilon_{\col}$. The distributional robustness approach can be visualized as a robust ball around the robot's state for collision checking using the convex feasible subset $\setX_s$ of the non-convex feasible space $\setX_{\free}$ that is consistent with the specified risk of constraint violation.\end{proof}

\begin{rmrk} \label{rmrk:socp_lin_collision_constraint} 
The constraint~\eqref{eq:lin_chance_constraint_det_obst} is of the form $\Prob(a^\top x+b \leq 0)\geq 1- \epsilon_{\col}$, where $a = - (\bar{x}_{\nom} - \barp_{\obs})^\top \matC^\top \matC$, and $b = (\bar{x}_{\nom} - \barp_{\obs})^\top \matC^\top \matC \barp_{\obs} + r_{\safe} \|\matC(\bar{x}_{\nom} - \barp_{\obs})\|_2$.
Using the Lemma~2, we formulate a second-order cone constraint that is used in the SCP problem for collision checking. 
\end{rmrk}
\subsection{Collision Checking with Stochastic Obstacle} \label{subsec:stoc_obstacle}
We extend the deterministic formulation of the linear chance constraint in~\cref{eq:lin_chance_constraint_det_obst} to include uncertainty in obstacle state. Let the distribution of the obstacle state be $p_{\obs} \sim \setN (\mu_p,\Sigma_p)$, where $\mu_p$ is the mean, $\Sigma_p$ is the variance matrix and the radius of the obstacle is $r_{\obs}$.
\begin{asmp}
\label{asmp:decorrelated_obstacle}
The obstacle state distribution $p$ is uncorrelated to the state distribution $x$ of the robot.
\end{asmp}
The collision chance constraint at any time $t$ for a state distribution $x$ and radius $r_{\rob}$ is given as follows:
\begin{equation}
\Prob\left( \|\matC(x - p_{\obs})\|_2 \geq r_{\rob} + r_{\obs}\right) \geq 1 - \epsilon_{\col},
    \label{eq:chance_constraint_stoc_obst}
\end{equation}
where both $x$ and $p_{\obs}$ are random variables, unlike~\eqref{eq:chance_constraint_det_obst}.

\begin{lemma}\label{theorem:stochastic_obstacles}
The linear chance constraint,
\begin{align}
    \Prob\Big(&(\bar{x}_{\nom}- \barp_{\obs})^\top \matC^\top \matC (x - p_{\obs}) \nonumber\\ &\geq r_{\safe} \|\matC(\bar{x}_{\nom}- \barp_{\obs})\|_{2} \Big) \geq 1 - \epsilon_{\col},
    \label{eq:lin_chance_constraint_stoc_obst}
\end{align}
 in the robot state distribution $x$ and the obstacle state distribution $p_{\obs}$ is a conservative approximation of the nonlinear collision chance constraint in~\eqref{eq:chance_constraint_stoc_obst} at any time $t \in [t_0, t_f]$, where $\bar{x}_{\nom}$ is a realization of the nominal state distribution $x_{\nom}$ of the robot, $\barp_{\obs}$ is a realization of the obstacle state distribution $p_{\obs}$, and $r_{\safe} = r_{\rob} + r_{\obs}$.
\end{lemma}
\begin{proof}
  Consider the sets $\setX_{\free}$ and $\setX_{s}$, defined as $\setX_{\free} = \{x:  \|\matC(x - p_{\obs})\|_2 \geq r_{\safe}$  and $\setX_s = \{x:(\bar{x}_{\nom}- \bar{p}_{\obs})^\top \matC^\top \matC (x - p_{\obs}) \geq r_{\safe} \|\matC(\bar{x}_{\nom}- \bar{p}_{\obs})\|_{2}\}$ respectively, where $\bar{p}_{\obs}$ is a sample of the obstacle state distribution $p_{\obs}$. In constraint $\setX_s$, $(x - p_{\obs})$ is the decision variable. Note that for any realization of the states $\bar{x}$ and $\barp_{\obs}$ we have $\setX_s \subseteq \setX_{\free}$ (see~\cite{morgan2014model} for proof). Using the arguments in \cref{theorem:deterministic_obstacles}, constraint \cref{eq:lin_chance_constraint_stoc_obst} is a conservative approximation of constraint \cref{eq:chance_constraint_stoc_obst}.
\end{proof}
\begin{rmrk} \label{remark:socp_stoc_obstacles}
The constraint~\cref{eq:lin_chance_constraint_stoc_obst} is a linear chance constraint of the form $\Prob(a^\top  (x - p_{obs}) + b \leq 0) \geq 1 - \epsilon_{\col}$, where $ a = - (\bar{x}_{\nom}- \barp_{\obs})^\top \matC^\top \matC$, $b = r_{\safe} \|\matC(\bar{x}_{\nom}- \barp_{\obs})\|_{2}$, and $p_{\obs} \sim \setN(\mu_p, \Sigma_p)$. In this case, the distributionally robust deterministic surrogate is computed for the stacked state $x_{c} = [x^\top p_{\obs}^\top]^\top$, which includes both the robot state and the obstacle state. The surrogate constraint is given as follows:
\begin{equation}
    a^\top \mu_x - a^\top \mu_p + b + \sqrt{\tfrac{1-\epsilon_{\col}}{\epsilon_{\col}}} \sqrt{a^\top \Sigma_{x} a + a^\top \Sigma_p a} \leq 0.
    \label{eq:det_surrogate_stoc_obstacles}
\end{equation}
Using~\cref{lemma:lin_socc_gpc}, the inequality constraint in the moment space is transformed into a second-order cone constraint in terms of the gPC states $\gpcX$ of the robot dynamics.
\end{rmrk}
\begin{rmrk} \label{remark:socp_correlated_obstacles}
For a correlated obstacle state $p$ and a robot state $x$, with the cross-correlation matrix $\Sigma_{xp}$, the deterministic surrogate of~\cref{eq:chance_constraint_stoc_obst} is given as follows:
\begin{align}
        & a^\top \mu_x - a^\top \mu_p + b + \nonumber\\ & \sqrt{\tfrac{1-\epsilon_{\col}}{\epsilon_{\col}}} \sqrt{a^\top \Sigma_{x} a + 2 a^\top \Sigma_{xp} a + a^\top \Sigma_p a} \leq 0.
        \label{eq:det_surrogate_stoc_obstacles_decorrelated}
\end{align}
The derivation uses the stacked state $x_c$, as shown in~\cref{remark:socp_stoc_obstacles}.
\end{rmrk}
\begin{rmrk} \cref{theorem:stochastic_obstacles} can be applied for safe multi-agent reconfiguration under uncertainty by replacing the obstacle state $p_{obs}$ with the neighbouring robots state. The robots communicate the moments used in~\cref{eq:det_surrogate_stoc_obstacles_decorrelated}  for collision checking with the neighbouring agents.
\end{rmrk}
\subsection{Terminal Constraint}\label{subsec:term_constraint} The terminal constraint is defined as an ellipsoidal set $(x-\bar{x}_f)^\top Q_{\setX_f}(x-\bar{x}_{f}) \leq c_{f}$ around a terminal point $\bar{x}_{f}$, where $Q_{\setX_f}$ is a positive definite matrix. The chance constraint formulation of the terminal set involves two steps: 1) constrain the mean of the terminal point as $\mu_{f} = \bar{x}_{f}$ and 2) formulate the quadratic chance constraint $\Prob((x-\bar{x}_f)^\top Q_{\setX_f}(x-\bar{x}_{f}) \leq c_{f})\geq 1- \epsilon_{f}$ around the mean $\mu_f$ with the risk measure $\epsilon_f$ of not reaching the terminal set. We use the conservative deterministic constraint discussed in~\cref{lemma:quad_const}, that bounds the variance of the state. The terminal constraints are summarized as follows.
\begin{align}
    \mu_{f} = \bar{x}_{f},\quad \frac{1}{c_f}\text{tr}(Q_{\setX_f}\Sigma_{x_{f}}) \leq \epsilon_{f} \label{eq:terminal_set_mp}
\end{align}
where $\mu_f$ is the mean and $\Sigma_{x_{f}}$ is the variance of the terminal state. The conservative approximations we presented in this section are a trade-off between the knowledge of the moments available and the computational speed achieved by convex constraints. For a linear SDE with obstacles whose uncertainty is described by the Gaussian distribution, a tighter equivalent deterministic surrogate constraint can be derived using the covariance propagation technique for uncertainty propagation and the inverse cumulative distribution function for the Gaussian distribution. 

\subsection{Motion Planning Algorithm (gPC-SCP)}
For motion planning, we integrate the deterministic approximations discussed in~\cref{subsec:det_obstacle,subsec:stoc_obstacle,subsec:term_constraint} with the asymptotically optimal rapidly exploring random trees~\cite{hauser2016} (AO-RRT). Following Algorithm~\ref{Algo:MP-gPC-SCP}, outlines the motion planning method using gPC-SCP for a dynamical system under uncertainty.
\begin{algorithm}
\DontPrintSemicolon
\caption{gPC-SCP}
\label{Algo:MP-gPC-SCP}
\KwInput{Map, obstacle location, $x_{0}$, $\setX_{f}$, $\Delta t$, $\ell$,}
\KwInput{Uncertainty model of $g(x,\baru)$ in SDE~\cref{eq:nl_stochastic_dynamics}.}
\KwOutput{Optimal and safe state  distribution $\setX_{\sol} = \{x_{0},x_{1},...,x_{T}\}$ and control input $\setU_{\sol} = \{\bar{u}_{0},\bar{u}_{1},...,\bar{u}_{T-1}\}$.}
\Comment{\emph{Stage 1: gPC Projection.}}
\cref{prob:dr_det_optimal_cntrl} $\leftarrow$ \blockgpcprojection { Formulate the collision constraint using~\cref{theorem:deterministic_obstacles,theorem:stochastic_obstacles},\;
Formulate the terminal set $\setX_f$ using~\cref{eq:terminal_set_mp},\;
Project the SDE using~\cref{eq:continuous_ode_form},\; 
Project the collision constraint using~\cref{lemma:lin_socc_gpc},\; Project the terminal set using~\cref{lemma:sdp_con_gpc},\;
Setup and project cost function using~\cref{eq:cost_odeopt},\;
\textbf{return:}~\cref{prob:dr_det_optimal_cntrl} in gPC space.\;}
\cref{prob:dr_trajopt} $\leftarrow$ \Kwlinearize
(\KwDiscretize(\cref{prob:dr_det_optimal_cntrl}))\\
Save \cref{prob:dr_trajopt}.\\
\Comment{\emph{Stage 2: Compute a nominal trajectory using AO-RRT.}}
$\{\setX^{0}_{\sol},\setU^{0}_{\sol},T\}$ $\leftarrow$ \Kwaorrt($x_0,\setX_f,\Delta t,\dot{x} =f(x,\bar{u})$) 
\tcc*[r]{For detailed implementation of AO-RRT see~\cite{hauser2016}.}
\Comment{\emph{Stage 3: gPC-SCP.}}
$\{\setX_{\sol},\setU_{\sol}\}$ $\leftarrow$\KwSolve(\cref{prob:dr_trajopt},$\{\setX^{0}_{\sol},\setU^{0}_{\sol},T\}$)\tcc*[r]{The sequential convex programming (SCP) approach is described in~\cref{subsec:gPC-SCP-scp}.}
\end{algorithm}

Algorithm~\ref{Algo:MP-gPC-SCP} has three stages. In Stage 1, we formulate the linear chance constraint for collision checking and the quadratic chance constraint for the terminal constraint, respectively. Using the chance constraints, we setup \cref{prob:dr_stoc_opt_cntrl} (SNOC) and project it to \cref{prob:dr_det_optimal_cntrl} (DNOC). We formulate gPC-SCP in~\cref{prob:dr_trajopt} by discretizing~\cref{prob:dr_det_optimal_cntrl}. In Stage 2, we use AO-RRT to compute an initial feasible trajectory $\{\setX^0_{\sol},\setU^0_{\sol}\}$ for the nominal dynamics $\dot{x} = f(x,\baru)$. In Stage 3, the feasible trajectory is then used to initialize the SCP iterations in gPC-SCP, which optimizes for the uncertainty in dynamics. The output of stage 3 is a safe and optimal state trajectory in the gPC space. Using the gPC polynomials in~\cref{eq:state_kronck_notation}, the gPC space trajectory is projected back to the state space distribution to output $\{\setX_{\sol},\setU_{\sol}\}$ in line 12 of Algorithm~\ref{Algo:MP-gPC-SCP}. Note that RRT in AO-RRT can be replaced with sparse tree~\cite{li2016asymptotically} algorithm for improved speed and with RRT$^{*}$~\cite{karaman2011sampling} for optimality. We discuss the application of Algorithm~\ref{Algo:MP-gPC-SCP} in~\cref{sec:simulations_and_experiments}.
\color{black}
\subsection{Real-Time Planning using Predictor-Corrector gPC-SCP}
In the gPC-SCP method, we compute the full trajectory distribution of the state $x$ in the gPC state $X$. However, in many applications, we might need only the mean of the state trajectory $\mu_x$. Therefore, we present an iterative algorithm, called predictor-corrector gPC-SCP (gPC-SCP$^\mathrm{PC}$), to compute the mean trajectory $\mu_x$ safe under stochastic uncertainty in dynamics and obstacles. In the gPC-SCP$^\mathrm{PC}$ algorithm, at any SCP iteration, we first predict a covariance correction term by propagating the gPC dynamics~\cref{eq:continuous_ode_form} using the nominal trajectory $\{\setX^0_{\sol},\setU^0_{\sol}\}$ and then solve a deterministic optimal control problem to compute the mean of the trajectory that accommodates for the uncertainty through the correction term. We describe the two main steps involved in the gPC-SCP$^\mathrm{PC}$ method and the algorithm for computing a mean trajectory.

\subsubsection{Prediction Step}\label{subsec:prediction} Given a nominal open-loop control input $\setU_{\sol} = \{\baru_0, \baru_1,\dots,\baru_{T-1}\}$ and the initial condition $x_0$, we propagate the gPC dynamics~\cref{eq:continuous_ode_form} using the control $\setU_{\sol}$ and compute the corresponding trajectory of the gPC state $X(t)$.
\begin{equation}
     b_{\mathrm{c}}(t) = X(t)^\top U N N^\top U^\top X(t) \label{eq:gpc_to_state_covariance_correction}
\end{equation}
 Using \cref{eq:gpc_to_state_covariance_correction} and the matrices defined in~\cref{eq:mat_lin_con_gpc}, we compute the covariance correction term $b_{\mathrm{c}}$. The deterministic linear collison chance constraint~\cref{eq:det_surrogate_stoc_obstacles} in terms of $\mu_x$ and the correction term $b_{\mathrm{c}}$ for a given risk of collison $\epsilon_{\col}$ is given as follows:
 \begin{equation}
         a^\top \mu_x - a^\top \mu_p + b + \sqrt{\tfrac{1-\epsilon_{\col}}{\epsilon_{\col}}} \sqrt{b_{\mathrm{c}} + a^\top \Sigma_p a} \leq 0.
         \label{eq:mean_lin_col_checking}
 \end{equation}
where $\Sigma_p$ is the known (or estimated) covariance matrix of the obstacle state as defined in~\cref{remark:socp_stoc_obstacles}. Note that~\cref{eq:mean_lin_col_checking} is a linear constraint in $\mu_x$ with known $b_{\mathrm{c}}$ and risk value $\epsilon_{\col}$, where as~\cref{eq:det_surrogate_stoc_obstacles} is a second-order cone constraint. This reduces the complexity of the SCP problem and improves the computation speed. Using~\cref{eq:mean_lin_col_checking}, the safe set defined using the distributionally robust linear joint chance constraint $\inf_{x(t) \sim (\mu_x,\Sigma_{x})} \Prob(\land_{i = 1}^{m} a_{i}^\top x + b_{i} \leq 0) \geq 1- \epsilon$ is transformed into a set of linear constraints shown below.
 \begin{align}
   & a_{i}^\top \mu_x - a_{i}^\top \mu_p + b_{i} \nonumber \\ &+ \sqrt{\tfrac{1-\epsilon_{\col}}{\epsilon_{\col}}} \sqrt{b_{\mathrm{c}} + a_{i}^\top \Sigma_p a_{i}} \leq 0 \  \forall i \in \{1,\dots,m\} \label{eq:mean_safe_set}
 \end{align}
 The terminal constraint is defined using the mean of the terminal state~\cref{eq:terminal_set_mp}. The prediction step involves the propagation of the SDE using gPC approximated nonlinear dynamics. Although this is an additional step in the overall planning approach, it reduces the complexity of the optimal control problem. 
 
\subsubsection{Correction Step}\label{subsec:correction_step}
The prediction step is followed by a correction optimization to optimize the nominal trajectory for uncertainty. We describe the optimal control problem in the following with mean $\mu_x$ and $\bar{u}$ as decision variables. 
\begin{problem}\label{prob:correction_optimal_control}
Distributionally Robust DNOC with Mean State Variable.
\begin{align}
\underset{\mu_{x(t)},\Bar{u}(t)}{\min} &
\scalebox{0.9}{$\left[\int_{t_{0}}^{t_{f}}\left(\mu^\top_x Q \mu_x + \|\baru\|_{p} \right)dt +\mu^\top_x(t_f) Q_{f} \mu_x(t_f)\right]$} \nonumber \\
\text{s.t.} \quad
& \scalebox{0.9}{$d\mu_{x} = f(\mu_{x}(t),\Bar{u}(t))dt$} \label{eq:mean_dynamics}\\
& \mathrm{Safe\ Set:}\ \cref{eq:mean_safe_set} \\
&\scalebox{0.9}{$\Bar{u}(t) \in \setU \quad \forall t \in [t_{0},t_{f}]$} \label{eq:control_constraints}\\
&\scalebox{0.9}{$\mu_{x}(t_{0}) = \Exp(x_0) \quad \mu_{x}(t_{f}) = \bar{x}_f$} \label{eq:mean_init_term_constraints}
\end{align}
\end{problem}
\noindent where $Q$ and $Q_f$ are positive definite matrices, as in~\cref{eq:cost_stopt}. The dynamics of the mean state is given in~\cref{eq:mean_dynamics}. We compute the mean dynamics using only $\phi_0$ as the polynomial in the gPC expansion method. The initial and terminal constraints in~\cref{eq:mean_init_term_constraints} are the expected values of the initial and terminal distributions. The above optimal control problem is linearized and discretized to formulate a convex optimization problem as described in~\cref{subsec:gPC-SCP-scp}. We use the prediction and correction steps to formulate the gPC-SCP$^\mathrm{PC}$ algorithm by modifying the gPC-SCP algorithm. The gPC-SCP$^\mathrm{PC}$ algorithm is given in the following Algorithm~\ref{Algo:PC-gPC-SCP}. We initialize gPC-SCP$^\mathrm{PC}$ using AO-RRT similar to Algorithm~\ref{Algo:MP-gPC-SCP}.

\textbf{Convergence and Feasibility.} Similar to the gPC-SCP method, we use a trust region-based iteration scheme to ensure feasibility of~\cref{prob:correction_optimal_control}. Additionally, in~\cref{prob:correction_optimal_control}, the risk of collision $\sqrt{\frac{1-\epsilon_{\col}}{\epsilon_\col}}$ can be used as a decision variable to compute a safe trajectory with optimal risk. This also ensures feasibility by computing a high-risk trajectory when a specified risk violation trajectory is not available. The convergence of gPC-SCP$^\mathrm{PC}$ depends on the size of the uncertainty $g$ and the SCP iteration scheme similar to gPC-SCP. We refer to the work in~\cite{foust2020optimal} for the proof of convergence when the prediction step is integrated with SCP for correction.

\begin{table}[h]
    \centering
    \begin{tabular}{|c|c|c|}
    \hline
     & gPC-SCP& gPC-SCP$^\mathrm{PC}$\\
     \hline
    Trajectory & Full distribution ($x$) & Mean ($\mu_x$)\\
    \hline
     Convex Problem Type & SOCP & QP \\
     \hline
    Solution & $(x,\bar{u})$  & $(\mu_x,\bar{u})$\\
    \hline
    Covariance Minimization & Yes  & No\\
    \hline
    Computation Time & Medium (\SI{10}{s})  & Small (\SI{0.9}{s})\\
    \hline
    \end{tabular} \caption{A comparison of gPC-SCP and gPC-SCP$^\mathrm{PC}$.}
    \label{tab:gPC_vs_PC_SCP}
\end{table}

\textbf{gPC-SCP vs. gPC-SCP$^\mathrm{PC}$.}
In \cref{tab:gPC_vs_PC_SCP}, we make a comparison of the gPC-SCP and gPC-SCP$^\mathrm{PC}$ methods. In the gPC-SCP method, we compute the full state trajectory distribution ($x$), while in the gPC-SCP$^\mathrm{PC}$ method, we compute the mean state trajectory ($\mu_x$). The convexified collision constraint in gPC-SCP is a second-order cone, whereas, in gPC-SCP$^\mathrm{PC}$, it is a quadratic constraint. In the gPC-SCP method, we can compute a trajectory with minimum variance and design a trajectory with specified terminal state covariance.  
\begin{algorithm}
\DontPrintSemicolon
\caption{gPC-SCP$^\mathrm{PC}$}
\label{Algo:PC-gPC-SCP}
\KwInput{Map, obstacle location, $\mu_{x}(t_0)$, $\mu_x(t_f)$, $\Delta t$, $\ell$,}
\KwInput{Uncertainty model of $g(x,\baru)$ in SDE.}
\KwOutput{Optimal and safe mean trajectory $\mathbf{\mu}_{\sol} = \{\mu_{x_0},\mu_{x_1},...,\mu_{x_T}\}$ and control input $\setU_{\sol} = \{\bar{u}_{0},\bar{u}_{1},...,\bar{u}_{T-1}\}$.}
\Comment{\emph{Stage 1: gPC Projection}}
\cref{eq:continuous_ode_form} $\leftarrow$ $\textbf{gPC\ Projection}$~\cref{eq:nl_stochastic_dynamics}\\
\Comment{\emph{Stage 2: Formulate the Mean Variable Optimal Control Problem.}}
\textbf{Correction-SCP} $\leftarrow$ \Kwlinearize
(\KwDiscretize(\cref{prob:correction_optimal_control}))\\
\Comment{\emph{Compute Nominal Trajectory.}}
$\{\setX^{0}_{\sol},\setU^{0}_{\sol},T\}$ $\leftarrow$ \Kwaorrt($x_0,\setX_f,\Delta t,\dot{x} =f(x,\bar{u})$) 
\tcc*[r]{For detailed implementation of AO-RRT see~\cite{hauser2016}.}
\Comment{\emph{Stage 3: gPC-SCP$^\mathrm{PC}$ Algorithm.}}
$\setX_{\sol},\setU_{\sol}$ $\leftarrow$ $\setX^{0}_{\sol},\setU^{0}_{\sol}$\\
\blockpcscp{ 
  \While{Not Converged}{
    \Comment{Step 1: Prediction}
    $X(t)$ $\leftarrow$ Propagate~\cref{eq:continuous_ode_form} using $\setU_{\sol}$ as discussed in~\cref{subsec:prediction}\;
    $b_{\mathrm{c}}(t)$ $\leftarrow$ $X(t)^\top U N N^\top U^\top X(t)$\;
    \Comment{Step 2: Correction}
    $\setX_{\sol},\setU_{\sol}$ $\leftarrow$ $\textbf{Correction-SCP}$ as discussed in~\cref{subsec:correction_step}\; $\left(\{\setX_{\sol},\setU_{\sol},T\},b_{\mathrm{c}}(t) \right)$ \;
    }
}
\end{algorithm}

\changed{In Section~\ref{sec:simulations_and_experiments}, we compare the motion plans generated by the gPC-SCP$^\mathrm{PC}$ Algorithm~\ref{Algo:PC-gPC-SCP} with the plans computed using the gPC-SCP Algorithm~\ref{Algo:MP-gPC-SCP}.}

\color{black}
\section{Simulations and Experiments}
\label{sec:simulations_and_experiments}
We apply the gPC-SCP method (Algorithm~\ref{Algo:MP-gPC-SCP}) and the gPC-SCP$^\mathrm{PC}$ method (Algorithm~\ref{Algo:PC-gPC-SCP}) to design the safe and optimal motion plan distribution and a safe mean trajectory, respectively, for the dynamics of the modular robotic spacecraft simulator with three degrees of freedom configuration~\cite{nakka2018six} and six degrees of freedom configuration~\cite{nakka2018six}. For the dynamics of the spacecraft simulator, we conducted an empirical study using simulation to demonstrate the safety provided by Algorithms~\ref{Algo:MP-gPC-SCP} and~\ref{Algo:PC-gPC-SCP} compared to the Gaussian collision constraint~\cite{zhu2019chance,du2011probabilistic,blackmore2010probabilistic,blackmore2011chance,janson2018monte} given in~\cref{eq:gaussian_linear_constraint}. We ran the simulations on a Mac machine with the configuration: $7^{th}$ generation, Intel Core i7 process, and 16 GB RAM.  We solve the SCP problem using CVXpy~\cite{diamond2016cvxpy} and the ECOS~\cite{domahidi2013ecos} solver. We then validate the experimental results on the spacecraft simulator hardware platform, where the motion plans are computed on an NVIDIA Jetson TX2 computer
\subsection{Robotic Spacecraft Dynamics Simulator~\cite{nakka2018six}}\label{subsec:problem_setup}
\begin{figure}
    \centering
    \includegraphics[width=0.8\columnwidth]{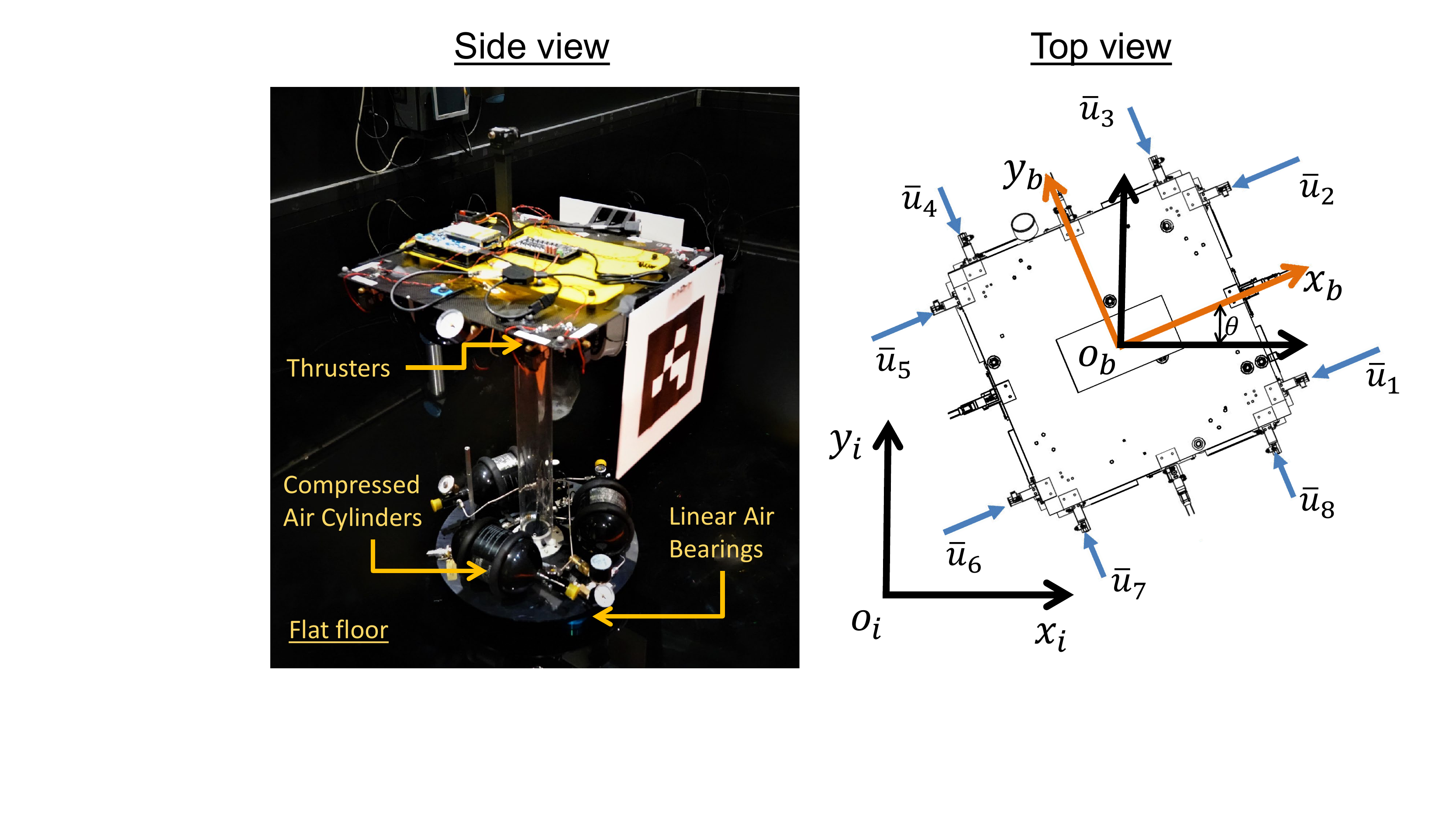}
    \caption{The top and side view of the Caltech's robotic spacecraft dynamics simulator.}
    \label{fig:mstar_setup}
\end{figure}

The Caltech's 3-DOF M-STAR (Multi-Spacecraft Testbed for Autonomy Research) is shown in~\cref{fig:mstar_setup}. The testbed floats on an ultraprecise epoxy floor using linear air bearings to achieve 3DOF frictionless motion. The M-STAR is equipped with eight thrusters for position $(x,y)$ and yaw angle ($\theta$) control. The dynamics of the robot is given as follows:
\begin{align}
    d\vecx = f(\vecx,\bar{u})dt + \sigma g(\vecx,\bar{u})dw,
\end{align}
where $\vecx = [x,y,\theta,\dot{x},\dot{y},\dot{\theta}]^\top$, $\bar{u} \in \real^{8}$, $\sigma \in \real$. The functions $f(\vecx,\bar{u})$ and $g(\vecx,\bar{u})$ are given below:
\begin{align}
    f(\vecx,\bar{u}) &= \begin{bmatrix} \mathbb{I}_{3\times3}&0\\0&0\end{bmatrix}\vecx + \begin{bmatrix}0\\B(m,\mathrm{I},l,\theta)\baru\end{bmatrix},\\
    g(\vecx,\bar{u}) &=\mathrm{blkdiag} \{0,B(m,I,l,\theta)\baru \}. \label{eq:spacecraft_dynamics}
\end{align}
The control effort $\bar{u}$ is constrained to be $0 \leq \baru\leq$\SI{0.45}{N}, and $B(m,\mathrm{I},l,\theta) \in \real^{3 \times 8}$ is the control allocation matrix (see~\cite{nakka2018six}), where $m=$ \SI{10}{kg} and $\mathrm{I}=$ \SI{1.62}{kg.m.s^{-2}} are the mass and the inertia matrix, and $l=$ \SI{0.4}{m} is the moment arm. Uncertainty $\sigma g(\vecx,\bar{u})$ is due to viscous friction between the robot and the flat floor, drift due to gravity gradient, and uncertainty in thruster actuation. We choose $\sigma = 0.1$, this value encompasses all the above forms of uncertainty. With this model, we study the convergence and collision checking discussed in~\cref{theorem:convergence_to_true_optimal,theorem:deterministic_obstacles,theorem:stochastic_obstacles}.
\subsubsection{Simulation}\label{subsec:simulations}
\begin{figure}[htpb]
    \centering
    \includegraphics[width=\columnwidth]{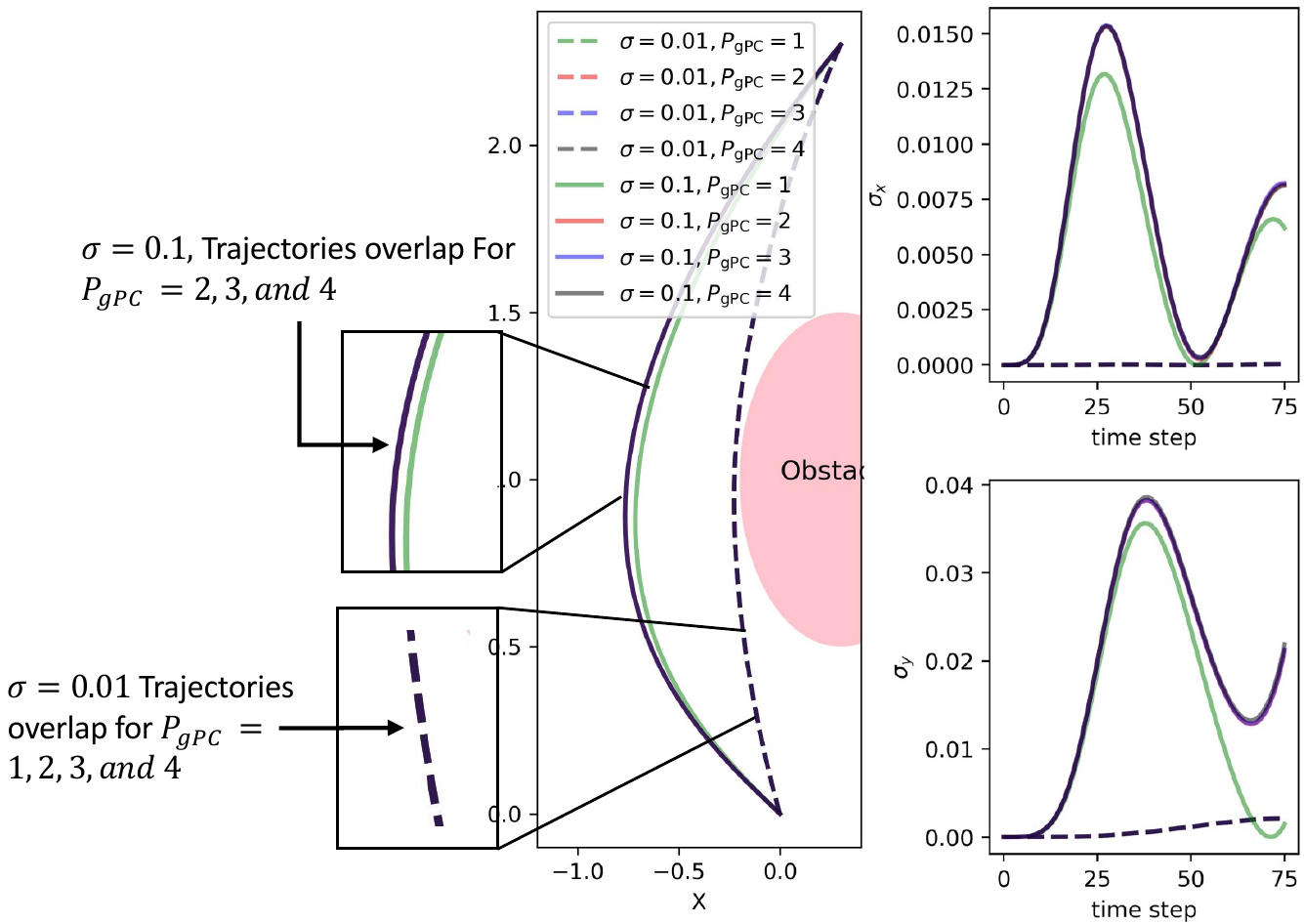}
    \caption{The figure demonstrates convergence of the mean (shown in the $(x,y)$ plane) and the variance ($\sigma_{x},\sigma_{y}$) (shown as a function of time) of the states $(x,y)$ with increasing $P_{\gpc}$ for $\sigma=\{0.01,0.1\}$.}
    \label{fig:motion_planning_variation_pgpc}
\end{figure}

Consider the map shown in~\cref{fig:motion_planning_variation_pgpc}. We design a safe and optimal trajectory, $J = \|\bar{u}\|_{2}$, $J_f = \Exp(\vecx_{f}^\top \vecx_{f})$, from the initial state $\Exp(x_{0}) = 0$ to the terminal state $\Exp(x_{f}) = [0.3,2.3,0,0,0,0]^\top$, while avoiding the obstacle located at $p_{\obs} = [0.3,1,0,0,0,0]^\top$ with radius $r_{\safe}=$ \SI{0.5}{m}. We formulate the collision constraint using~\cref{theorem:deterministic_obstacles,theorem:stochastic_obstacles} and bound the terminal variance using a slack variable to ensure feasibility.
In~\cref{fig:motion_planning_variation_pgpc}, we show the mean and variance of the position of M-STAR, $(x,y)$, computed using Algorithm~\ref{Algo:MP-gPC-SCP}. We compare the mean and variance computed for the gPC polynomial degree $P_{\gpc}=\{1,2,3,4\}$ and the variance $\sigma = \{0.01,0.1\}$ with $g(\vecx,\bar{u}) = [0,B \bar{u}]^\top$ in~\cref{fig:motion_planning_variation_pgpc}. The convergence of mean and variance with increasing $P_{\gpc}$, implies convergence with respect to $\ell$, validating~\cref{theorem:convergence_to_true_optimal}. Since there are no known methods to compute global optimal solution for \cref{prob:cc_stoc_opt_cntrl}, we cannot comment on the sub-optimality of the solution. For the case with low variance $\sigma=0.01$, we observe that the degree gPC polynomials $P_{\gpc} =1$ are sufficient to accurately calculate the mean and variance. However, for a large variance $\sigma = 0.1$, we need gPC polynomials of degree $P_{\gpc}=2$. We can use $P_{\gpc}=1$ with large variance in dynamics for the following two cases: 1) short-horizon planning and 2) iterative planning with closed-loop state information updates. We use gPC polynomials with degree $P_{\gpc} =2$ in the following motion planning analysis.
\begin{figure}[htpb]
    \centering
    \includegraphics[width=0.8\columnwidth]{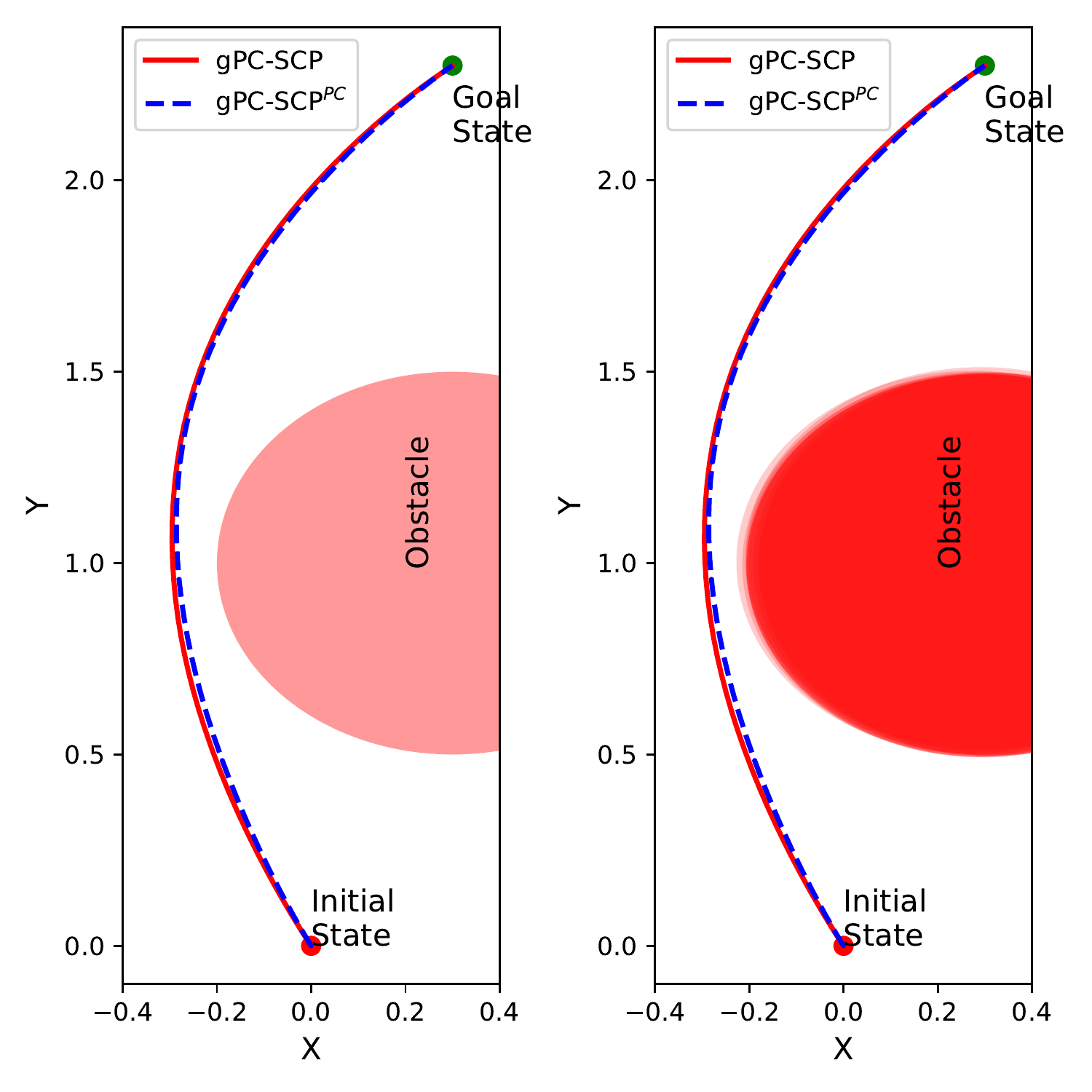}\caption{\changed{We compare the state trajectories generated using gPC-SCP and gPC-SCP$^\mathrm{PC}$ for deterministic(left) and stochastic obstacle (right).}}
    \label{fig:gpc_vs_pc_gpc}
\end{figure}
\begin{figure}[htpb]
    \centering
    \includegraphics[width=0.9\columnwidth,height=1.3in]{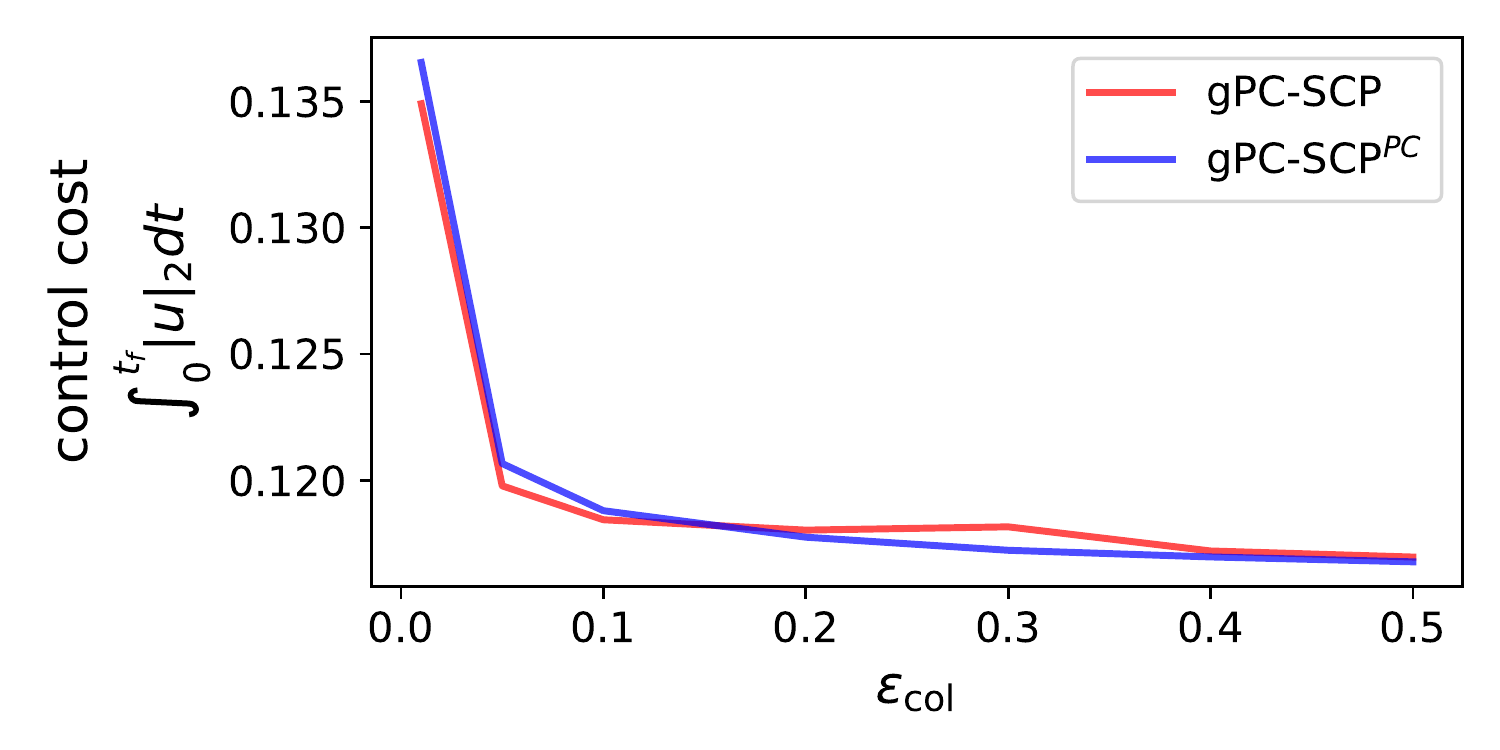}
    \caption{\changed{Integral cost of control for the open-loop trajectories computed using gPC-SCP and gPC-SCP$^\mathrm{PC}$ for various collison risk measures $\epsilon_{\col}=\{0.01,0.05,0.1,0.2,0.3,0.4,0.5\}$.}}
    \label{fig:cost_vs_risk}
\end{figure}

\changed{\subsubsection{gPC-SCP vs. gPC-SCP$^\mathrm{PC}$}\label{subsec:results_sim_motionplan}
In~\cref{fig:gpc_vs_pc_gpc}, we show the mean of the state trajectory distribution $x$ computed using gPC-SCP and the mean trajectory $\mu_x$ computed using gPC-SCP$^\mathrm{PC}$ for deterministic (light red) and stochastic obstacle (dark red). The uncertainty in the obstacle position is assumed to be a multivariate Gaussian with covariance $\Sigma_p=\mathrm{diag}([1e-4,1e-4])$. We use the distributionally robust linear chance constraint~\cref{eq:lin_socc} with risk measure $\epsilon_{\col}=0.05$ to convexify the free space, as discussed in~\cref{theorem:deterministic_obstacles,theorem:stochastic_obstacles}. For an equivalent comparison of gPC-SCP and gPC-SCP$^\mathrm{PC}$, we only use the control cost $\|u\|_2$ and constrain the expected initial and terminal states $\Exp{x}_0$ and $\Exp{x}_f$, respectively.}

\changed{We observe in~\cref{fig:gpc_vs_pc_gpc} that the mean of the trajectory distribution $x$ computed using gPC-SCP and the mean trajectory $\mu_x$ computed using gPC-SCP$^\mathrm{PC}$ are equivalent for the risk measure $\epsilon_{\col} = 0.05$. The computation time for gPC-SCP was \SI{10.86}{s} and  for gPC-SCP$^\mathrm{PC}$ was \SI{0.9}{s}. Note that using gPC-SCP, we can calculate the full trajectory distribution $x$, which is useful for applications such as safe exploration~\cite{nakka2020chance} and covariance minimization, while using gPC-SCP$^\mathrm{PC}$, we can calculate a probabilistic safe trajectory under tight computational time constraints for closed-loop feedback control. Furthermore, in~\cref{fig:cost_vs_risk}, we observe that the total optimal control cost for various collision risk measures $\epsilon_{\col} = \{0.01,0.05,0.1,0.2,0.3,0.4,0.5\}$ is similar for gPC-SCP and gPC-SCP$^\mathrm{PC}$ as we use the deterministic control assumption in the gPC-SCP method. For this reason, we use the mean of the state trajectory computed using the gPC-SCP method as the nominal state and control trajectory for hardware validation. Note that the optimal control cost decreases with increasing risk measures. As the risk measure increases, the robustness provided by the DRLCC~\cref{eq:lin_socc} decreases, leading to an increase in the feasible space. The larger feasible space leads to a more optimal and less safe trajectory.} 
\begin{figure}[htpb]
    \centering
    \includegraphics[width=\columnwidth]{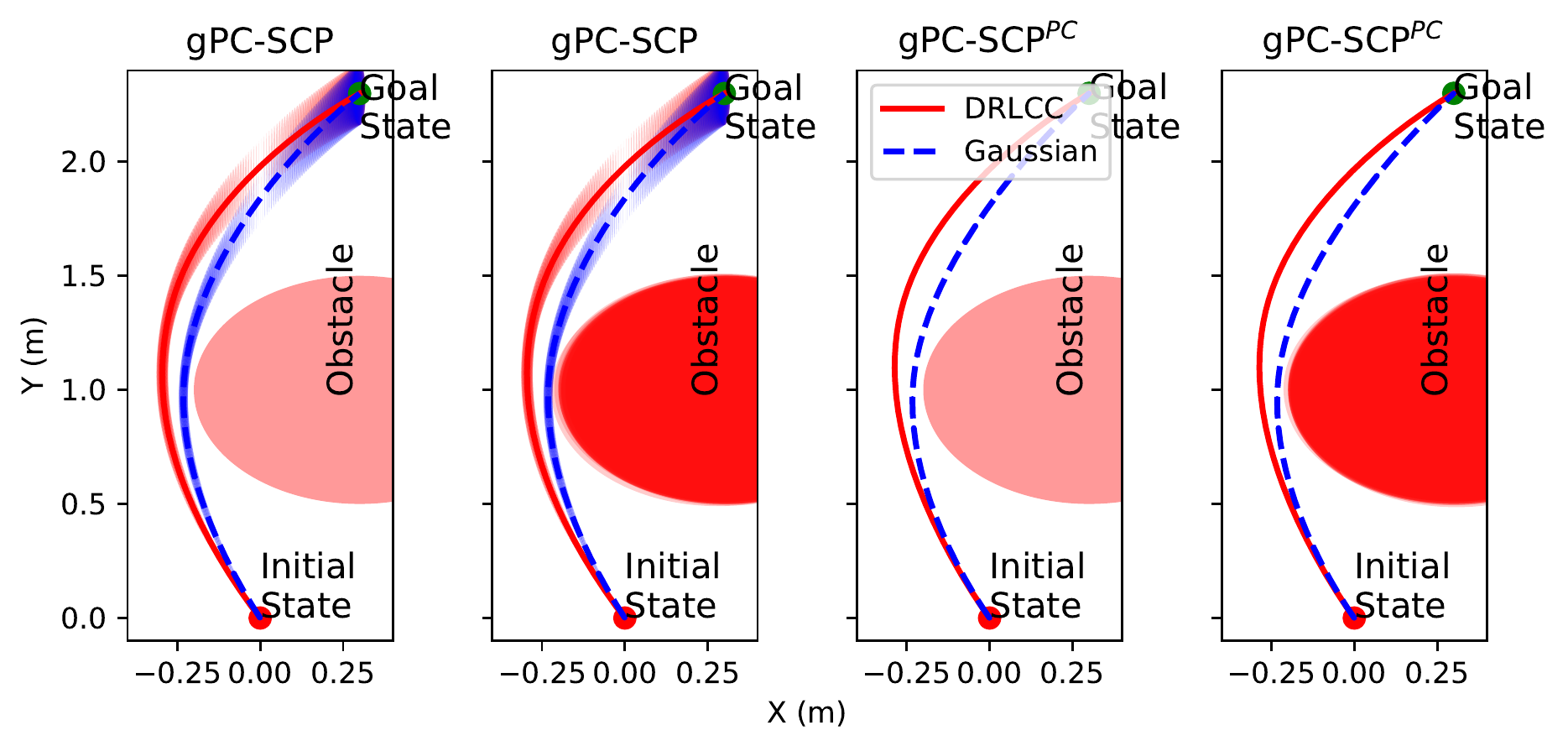}\caption{\changed{We compare the trajectories computed using gPC-SCP and gPC-SCP$^\mathrm{PC}$ with distributionally-robust linear chance constraint (DRLCC)~\cref{eq:lin_socc} and Gaussian chance constraint~\cref{eq:gaussian_linear_constraint} for collision checking for deterministic (light red) and stochastic obstacles (dark red).}}
    \label{fig:drlcc_vs_gaussian}
\end{figure}

\begin{figure}[htpb]
    \centering
    \includegraphics[width=0.85\columnwidth]{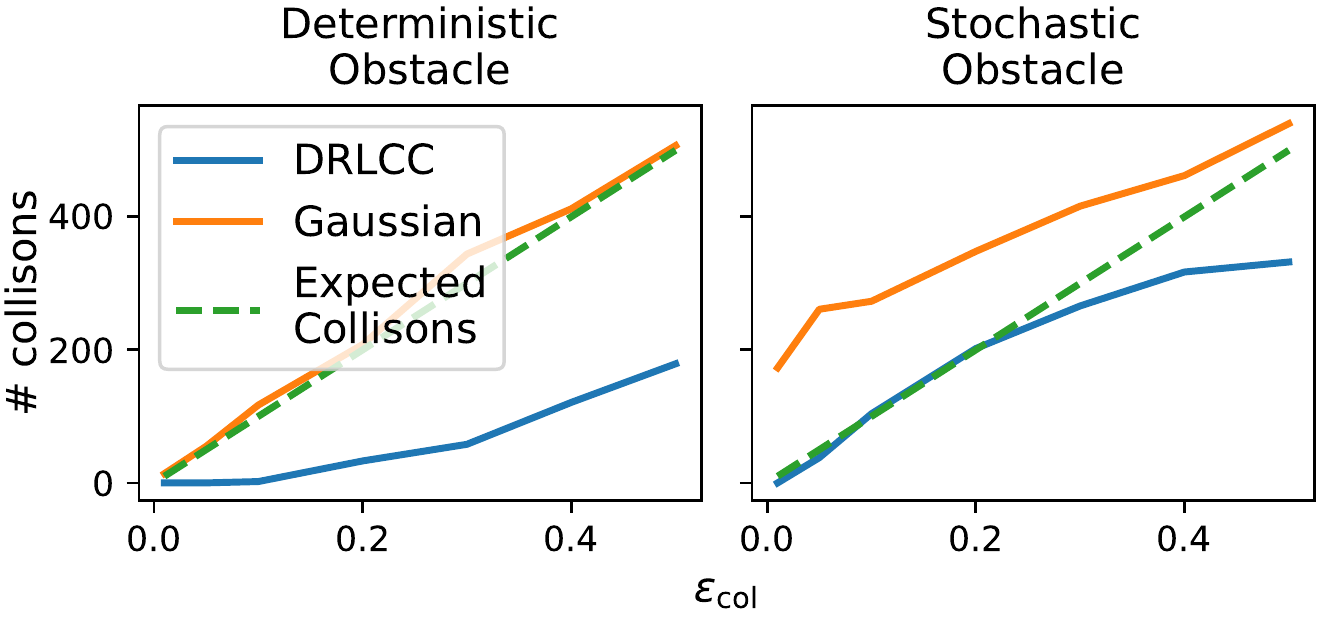}\caption{\changed{We show the results of 1000 Monte Carlo feedback control executions of the trajectories computed using gPC-SCP with distributionally-robust linear chance constraint (DRLCC)~\cref{eq:lin_socc} and Gaussian chance constraintx~\cref{eq:gaussian_linear_constraint} for collision checking for various risk measures $\epsilon_{\col}=\{0.01,0.05,0.1,0.2,0.3,0.4,0.5\}$.}}
    \label{fig:closed_loop_linear_constraint}
\end{figure}
\begin{figure}[htpb]
    \centering
    \includegraphics[width=0.9\columnwidth]{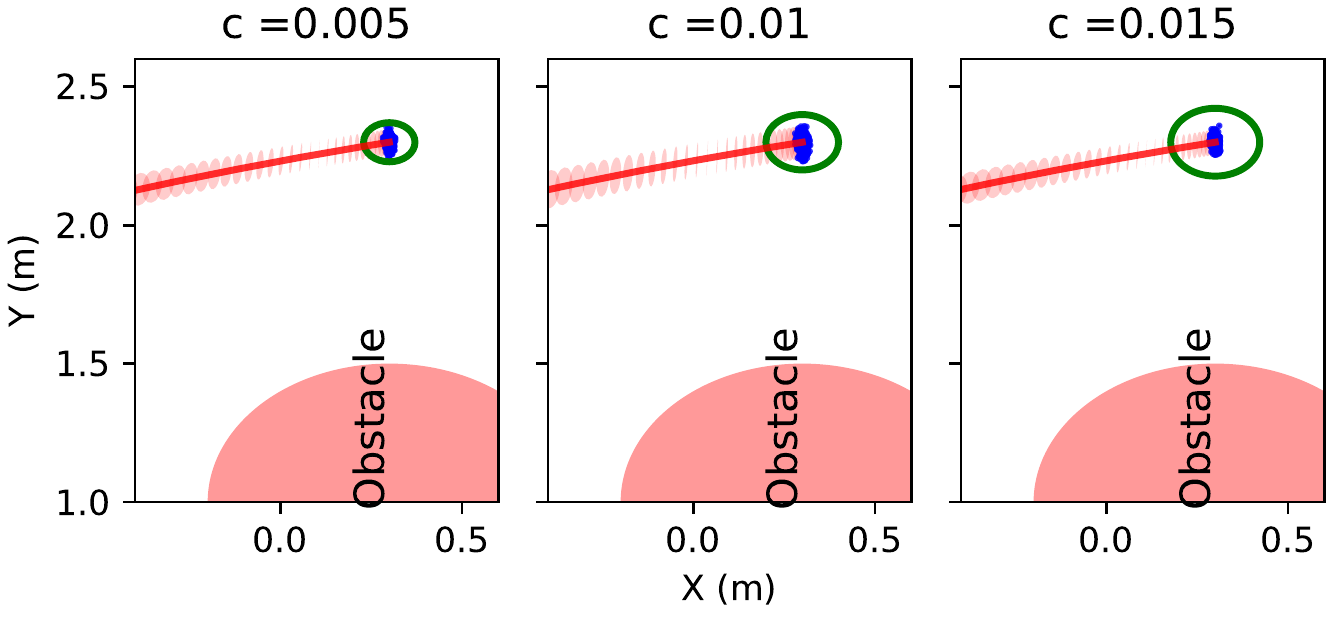}
    \caption{\changed{We compare the trajectories generated for different values $c=\{0.005,0.01,0.015\}$ of \cref{eq:quad_chnc_cons} (shown as the green circle) of the terminal set using the robust quadratic constraint~\cref{eq:sdp_constraint} with risk measure $\epsilon = 0.05$ for the terminal set. Finally, we show the terminal state of the robot (blue) when a nominal trajectory (sampled from the probabilistic trajectory) is executed using an exponentially-stabilizing controller.}}
    \label{fig:dr_terminal_set}
\end{figure}

\begin{figure}
    \centering
    \includegraphics[width=0.9\columnwidth]{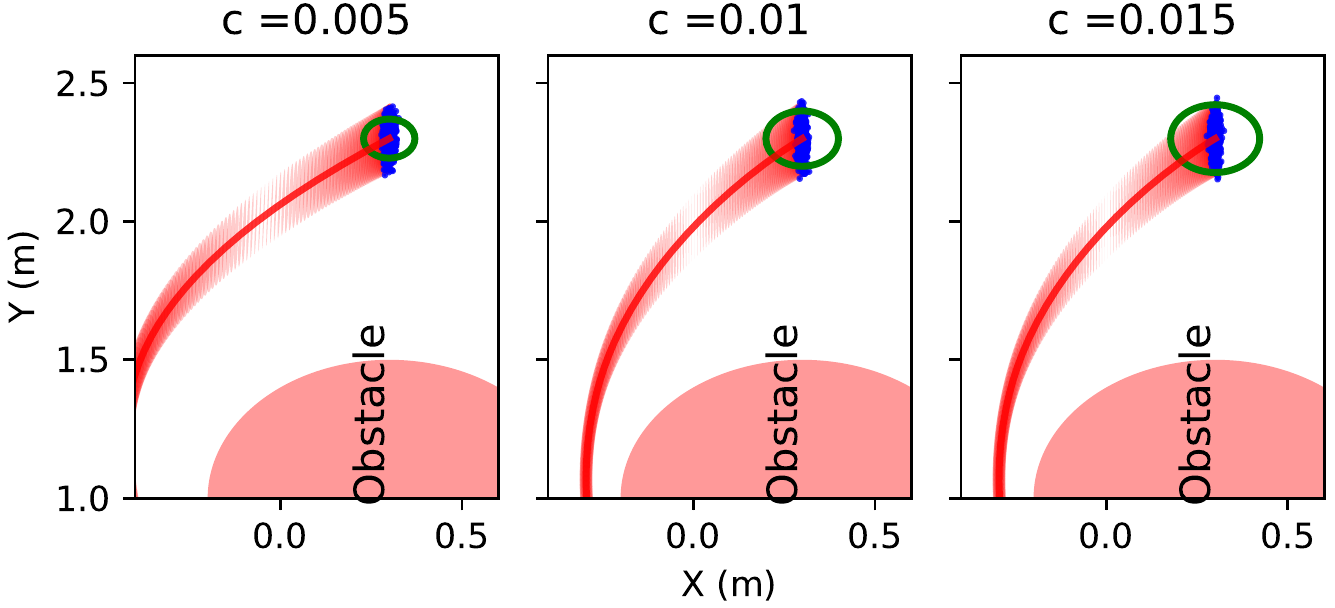}
    \caption{\changed{We compare the trajectories generated for different values $c=\{0.005,0.01,0.015\}$ of \cref{eq:quad_chnc_cons} (shown as the green circle) of the terminal set using $3\sigma$ confidence-based terminal set $3\Trace(\Sigma_x) \leq c$ with risk measure $\epsilon = 0.05$.}}
    \label{fig:gaussian_terminal_set}
\end{figure}
\changed{\subsubsection{Chance Constraints}
In~\cref{fig:drlcc_vs_gaussian}, we compare the motion plans computed using the DRLCC~\cref{eq:lin_socc} and the Gaussian linear chance constraint ~\cref{eq:gaussian_linear_constraint} for collision checking in both the gPC-SCP and the gPC-SCP$^\mathrm{PC}$ algorithms. We validate the safety of the motion plans computed using gPC-SCP by tracking a sampled trajectory with the exponentially-stabilizing controller designed in~\cite{nakka2018six} for the nominal dynamics. We sample a trajectory $\vecx$ using the projection equation $\vecx = \Phi^{\top}(\xi) \gpcX$. The gPC-SCP algorithm computes the gPC coordinates $\gpcX$. We compute $\vecx$ by randomly sampling the multivariate normal distribution $\xi$. Using the motion plans shown in~\cref{fig:drlcc_vs_gaussian} as input to the controller, we get the collision profile shown in~\cref{fig:closed_loop_linear_constraint} for various risk measures $\epsilon_{\col}=\{0.01,0.05,0.1,0.2,0.3,0.4,0.5\}$ over 1000 trials.} 

 \changed{ We observe in~\cref{fig:closed_loop_linear_constraint} that the DRLCC performs better than Gaussian collision checking for all the risk measures. Although the number of collisions when using Gaussian approximation is equivalent to the expected collisions for the deterministic obstacle case, it performs poorly in the stochastic obstacle case. The increased number of collisions under stochastic obstacles when using Gaussian approximation could be because the robot dynamics constraint in the SCP formulation is weakly nonlinear under deterministic constraints and strongly nonlinear under stochastic constraints. We observed that the performance of gPC-SCP and gPC-SCP$^\mathrm{PC}$ is equivalent as the trajectory computed by gPC-SCP$^\mathrm{PC}$ is a subset of the state distribution computed by gPC-SCP.}

\changed{The goal reaching of the robot for various terminal set sizes $c = \{0.005,0.01,0.015\}$m with $A = \identity$ is shown in~\cref{fig:dr_terminal_set} for the robust quadratic chance constraint~\cref{eq:sdp_constraint} and in~\cref{fig:gaussian_terminal_set} for the $3\sigma$-confidence-bound $3\Trace(\Sigma_x) \leq c$. We use a risk measure of $0.05$ for terminal constraint violation. The robust quadratic constraint is formulated as described in~\cref{subsec:term_constraint}. We observe in \cref{fig:dr_terminal_set} that for 1000 trials of feedback execution there are no violations of the terminal constraint using~\cref{eq:sdp_constraint}, while in \cref{fig:gaussian_terminal_set} for the $3\sigma$-confidence bound on variance the constraint violations for $c = \{0.005,0.01,0.015\}$ are $\{104, 31, 7\}$, respectively. Through this simulation, we conclude that the proposed constraint formulations provide higher degree of safety in comparison with the state-of-the-art Gaussian approximation.}

\subsubsection{Experiments}\label{subsec:experiments}

\begin{figure}
    \centering 
    \includegraphics[width=0.8\columnwidth]{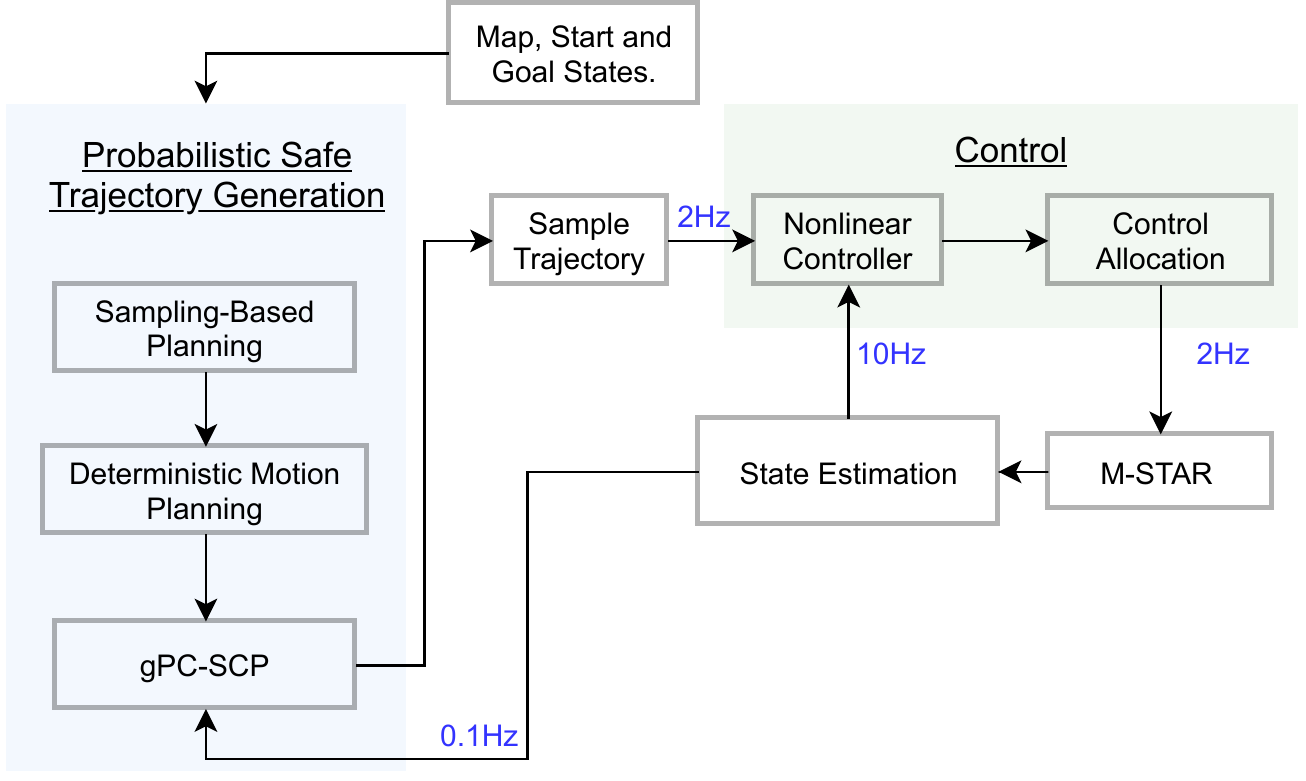}
    \caption{The guidance, navigation and control loop used for planning a distributionally-robust safe trajectory using gPC-SCP and controlling the 3 DOF spacecraft simulators.}
    \label{fig:autonomy_loop_gnc}
\end{figure}

\begin{figure*}
    \centering
    \includegraphics[width=0.3\textwidth]{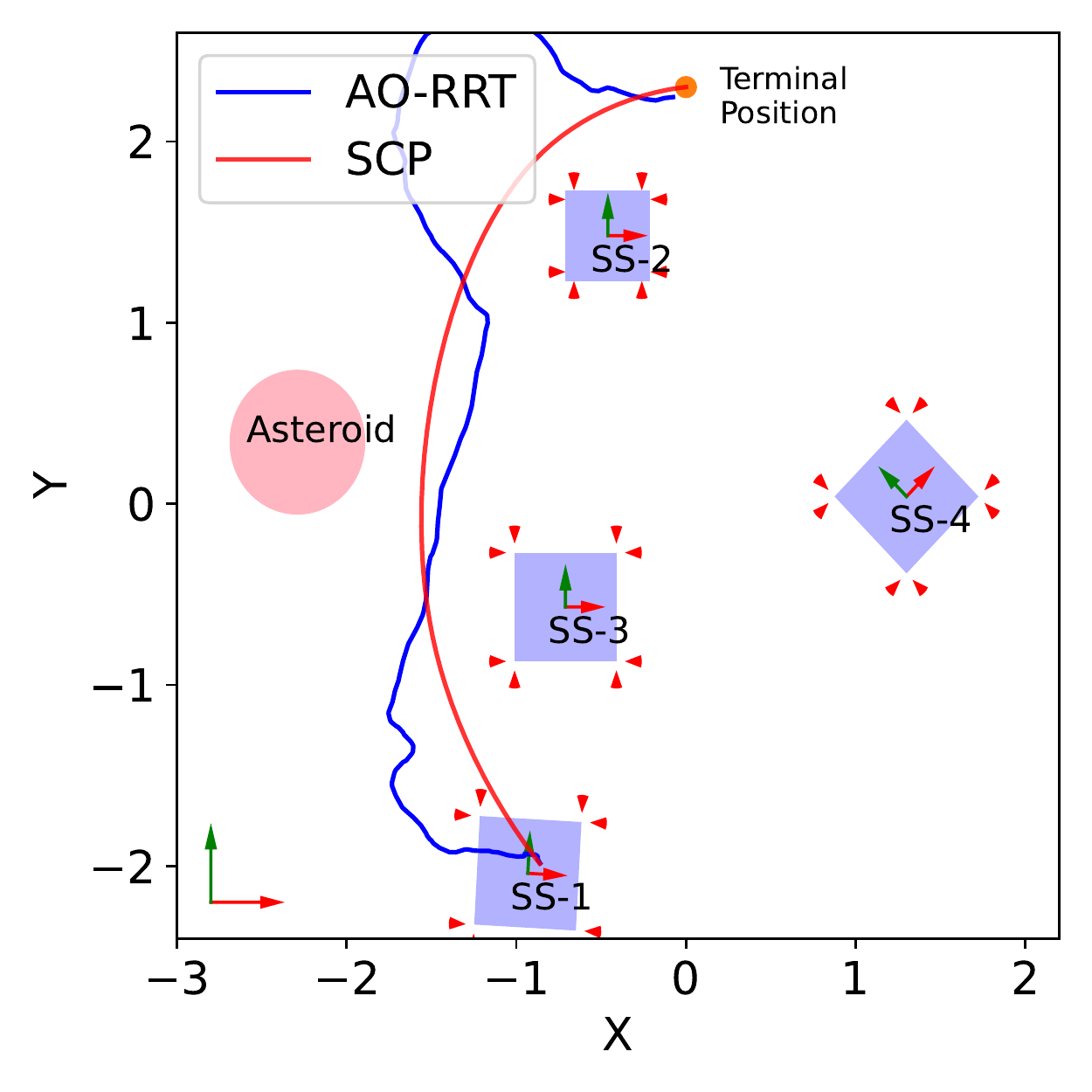}\includegraphics[width=0.3\textwidth]{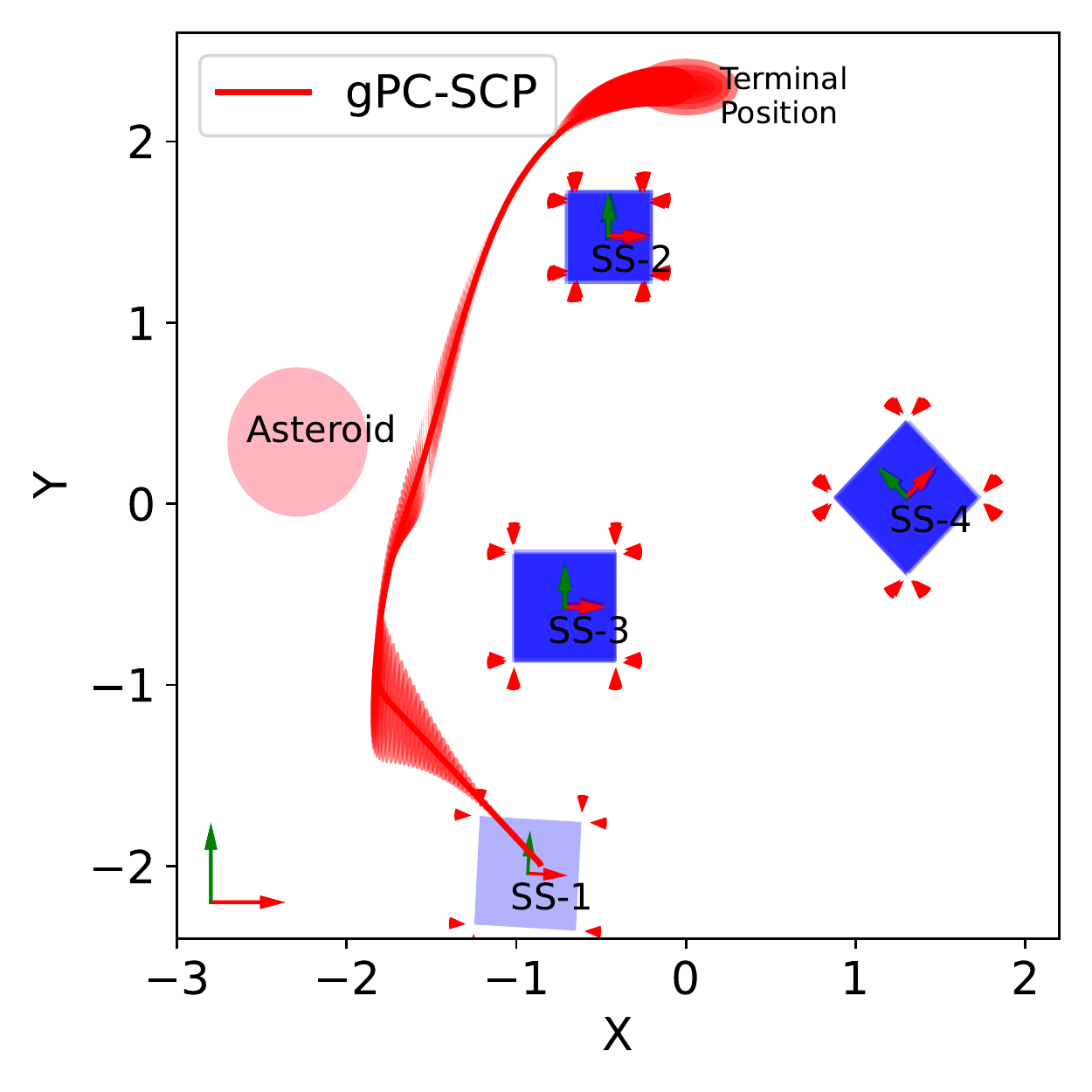}\includegraphics[width=0.3\textwidth]{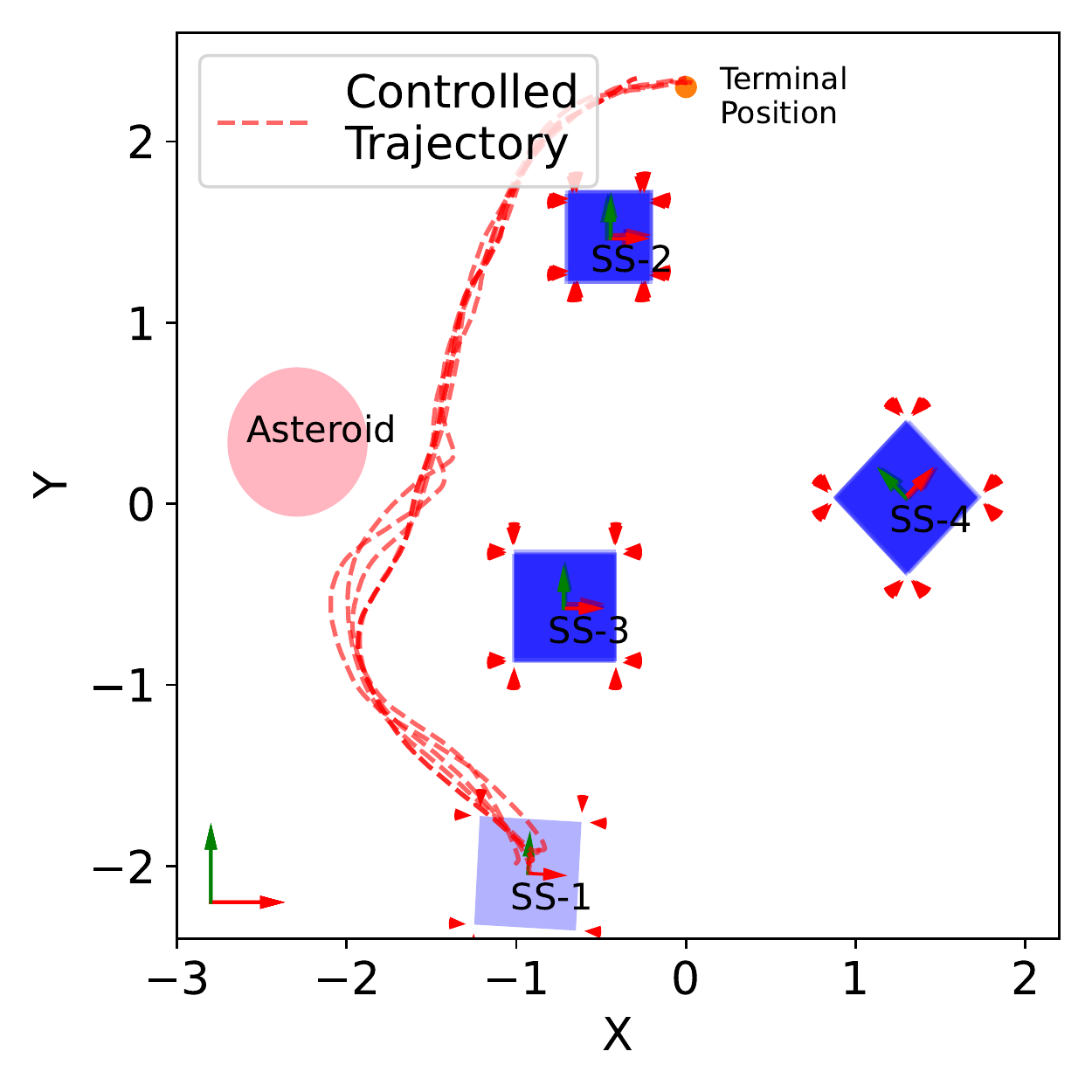}
    \includegraphics[width=0.3\textwidth]{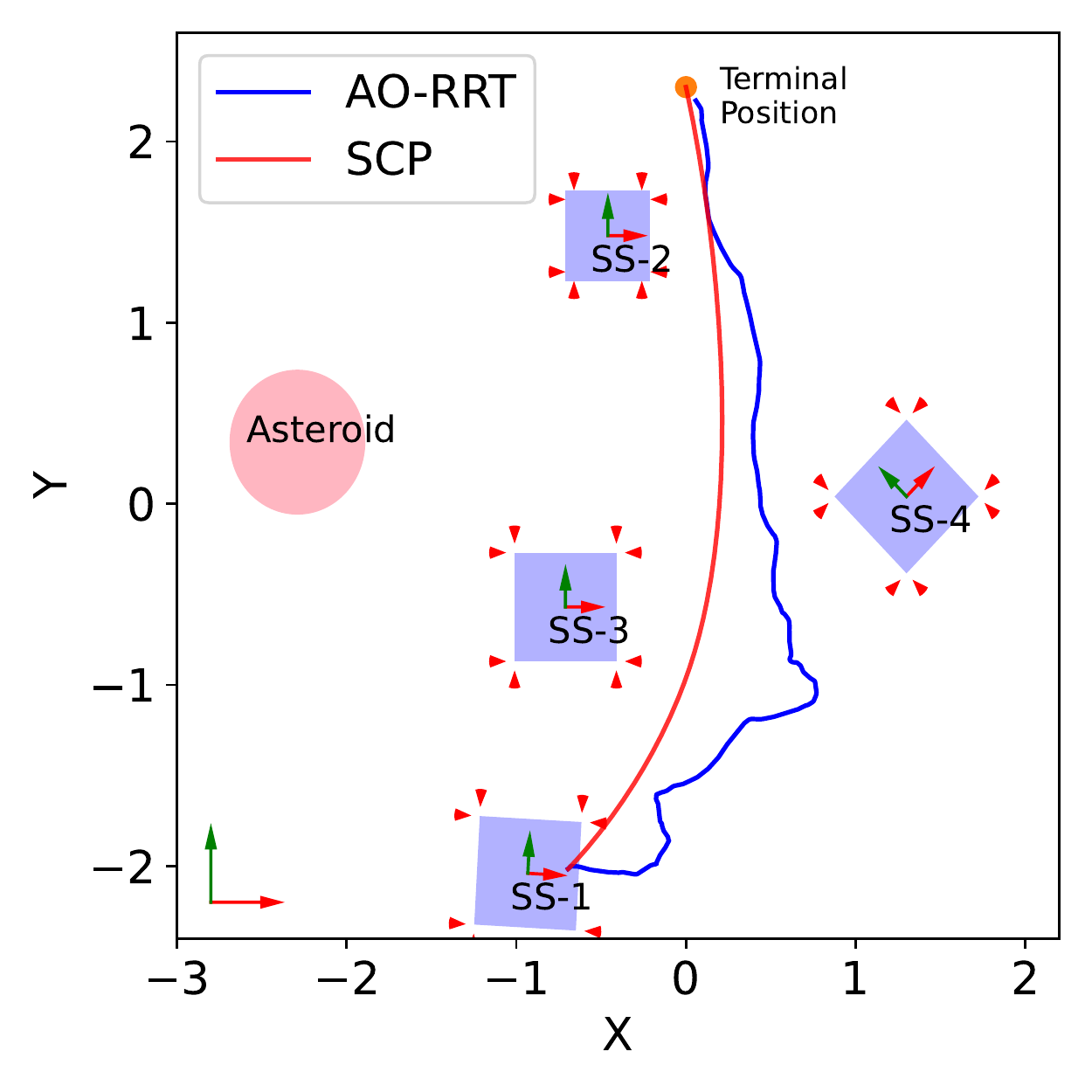}\includegraphics[width=0.3\textwidth]{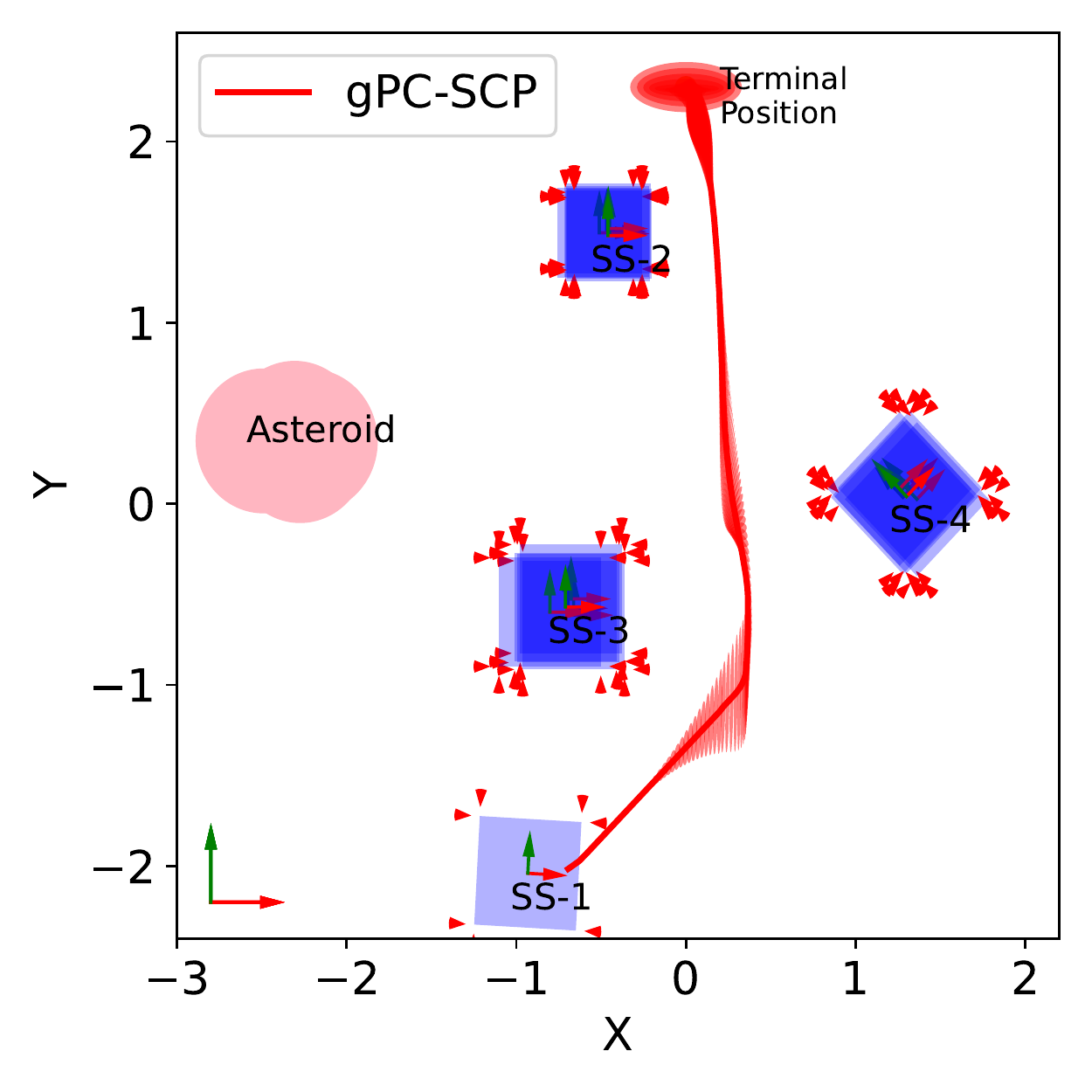}\includegraphics[width=0.3\textwidth]{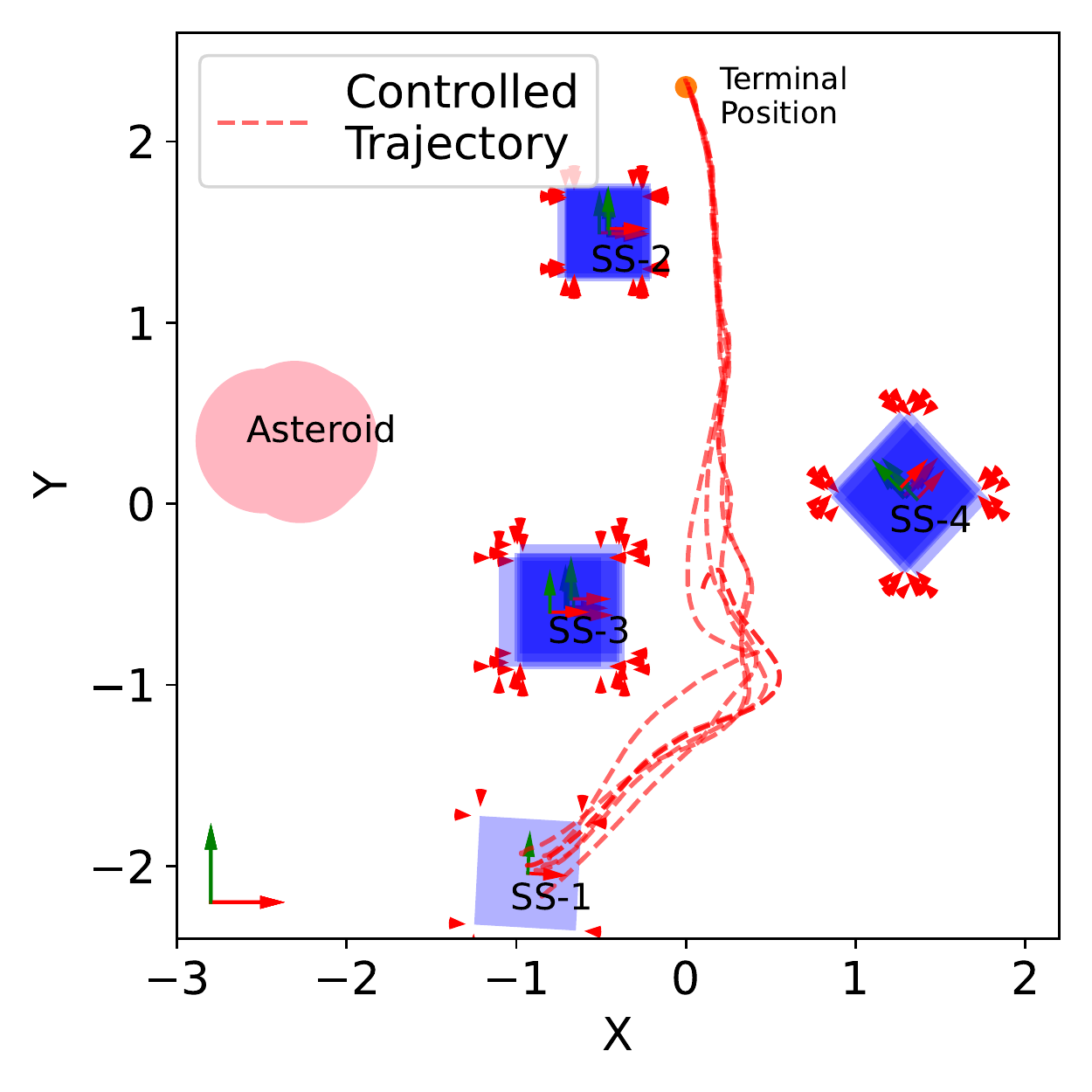}
    \caption{We show the output of the gPC-SCP method applied to the scenario shown in~\cref{fig:space_lab_experiment} at each stage of Algorithm~\ref{Algo:MP-gPC-SCP} and 10 trials of closed-loop trajectory tracking by using an exponentially-stabilizing controller designed in~\cite{nakka2018six}. Left: We show the output of AO-RRT for 5000 nodes and the SCP for the nominal dynamics. Middle: We show the probabilistic safe trajectory (2$\sigma$-confidence level) generated using the gPC-SCP method with a risk measure of $\epsilon = 0.05$ for collision checking. Right: We observe one failure in the 10 trials of the closed-loop trajectory execution.}
    \label{fig:spacecraft_simulator_experiments}
\end{figure*}

We apply Algorithm~\ref{Algo:MP-gPC-SCP} to the scenario shown in~\cref{fig:space_lab_experiment} using the closed-loop described in~\cref{fig:autonomy_loop_gnc} to design and execute safe plans for SS-1 in \cref{fig:space_lab_experiment} under uncertainty in dynamics and obstacle location. This scenario is relevant to the low-earth orbit, on-orbit, servicing application discussed in~\cite{nakka2021information}. For details on the sensing module to estimate the full state, control, and control allocation algorithm, refer to~\cite{nakka2018six}. We use the location of obstacles SS-2$=[-0.46,1.48]$, SS-3$=[-0.71,-0.57]$, SS-4$=[1.3,0.04]$, and Asteroid$=[-2.29,0.34]$ with radius \SI{0.4}{m} as shown in~\cref{fig:spacecraft_simulator_experiments}, and the uncertainty in position of obstacles $\Sigma_{p}=\mathrm{diag}([1e-4,1e-4])$ as input to Algorithms~\ref{Algo:MP-gPC-SCP},~\ref{Algo:PC-gPC-SCP}. The initial and terminal states of SS-1 are $\Exp(x_{0})=[-0.9,-2.3,0,0,0,0]^\top$ and $\Exp(x_{f})=[0,2.3,0,0,0,0]^\top$, respectively. We minimize the total control cost $\|u\|_2$ and  terminal variance of the trajectory using gPC-SCP method. The uncertainty description in~\cref{subsec:simulations} is valid for the following experimental setup.

\changed{In \cref{fig:spacecraft_simulator_experiments}, we present the results for 10 trials of the closed-loop tracking experiment. We compute an initial anytime trajectory using AO-RRT and optimize it for nominal dynamics. We use the optimized solution as an input for the gPC-SCP method. We observe that the method is biased towards the initial trajectory. As shown, the gPC-SCP method outputs a safe trajectory. We use the mean of the output trajectory as a reference trajectory for the controller. As shown in~\cref{fig:spacecraft_simulator_experiments}, the uncertainty in the model leads to drift in the system. We use a risk measure of $0.05$ for safe collision checking. The gPC-SCP method provides a safe trajectory for control by accommodating the uncertainty in dynamics by using the distributionally robust linear chance constraints. We observe one failure out of the 10 trials of closed-loop tracking. The failure was due to a large disturbance torque from the damaged floor on SS-1. Out of the 10 trials, 7 closed-loop tracking trials reached the expected terminal set. This validates the safety provided by the gPC-SCP method under uncertainty.}
\begin{figure}
    \centering
    \includegraphics[width=0.4\columnwidth]{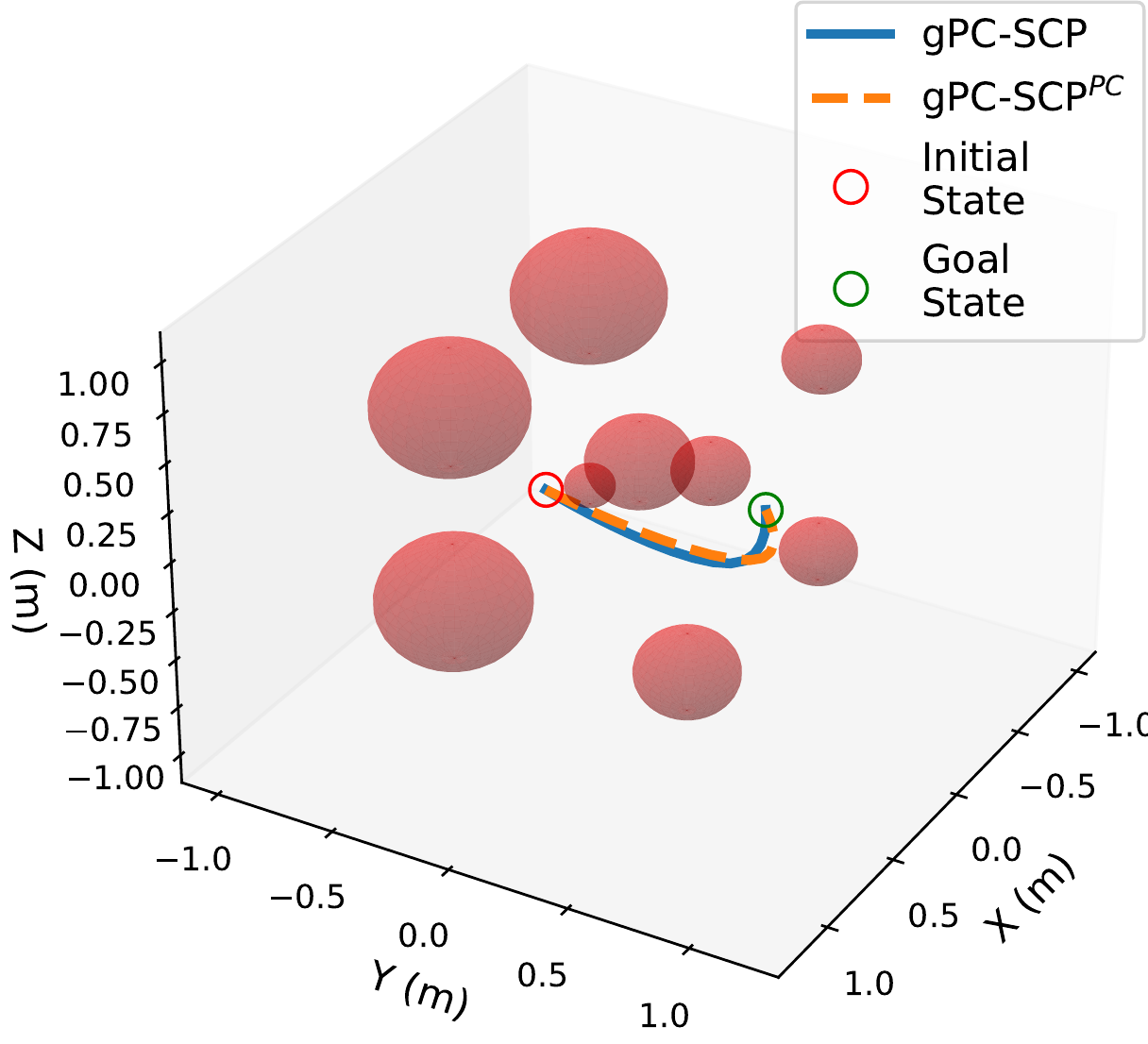}\includegraphics[width=0.42\columnwidth]{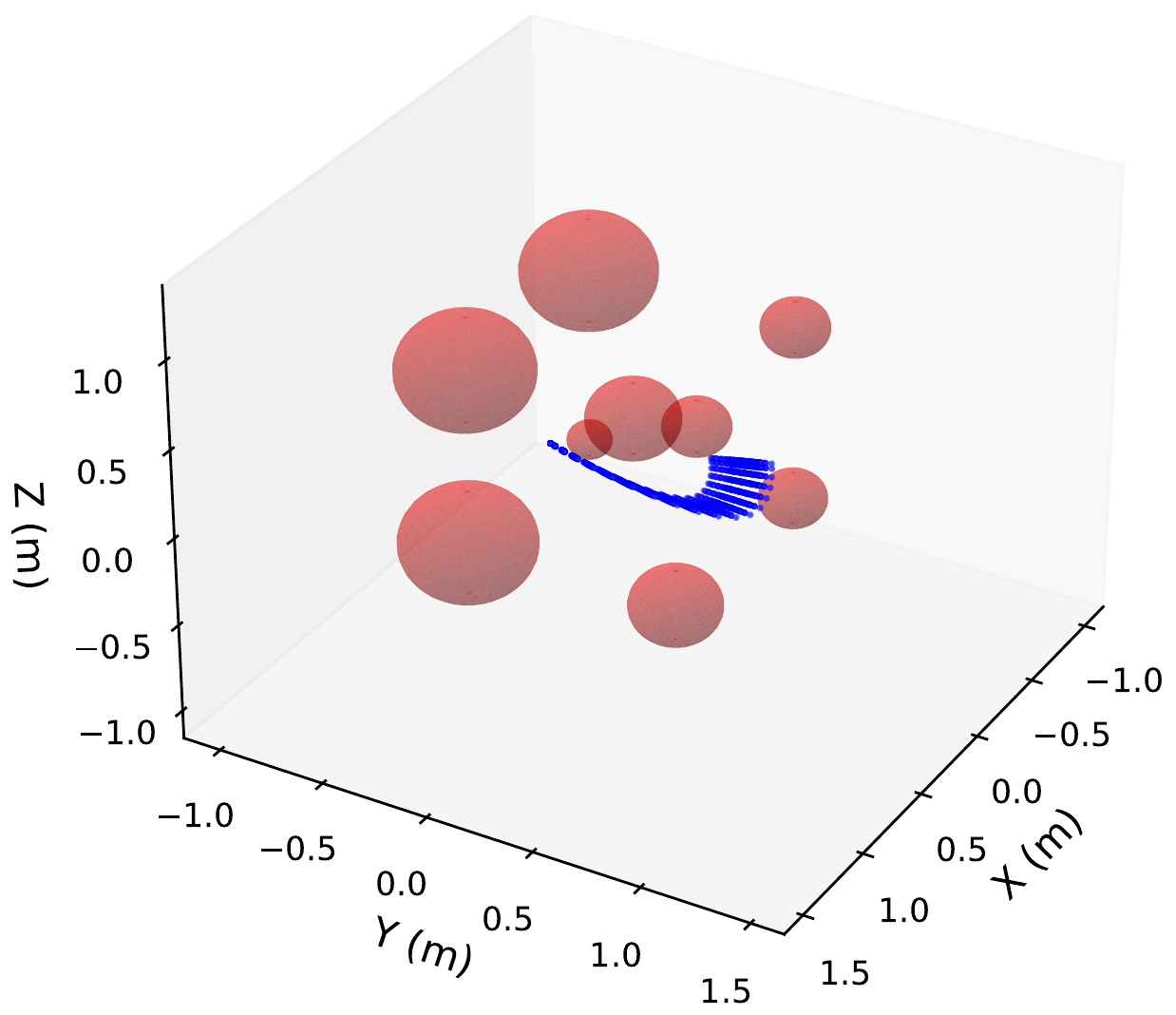}
    \caption{\changed{Left: We show the safe and feasible trajectories computed using gPC-SCP and gPC-SCP$^\mathrm{PC}$ for the Six-DOF spacecraft under uncertainty in dynamics. Right: Monte Carlo simulations of the closed-loop tracking of the trajectory computed using gPC-SCP.}}
    \label{fig:6DOF_free_flyer_simulation}
\end{figure}
\vspace{-0.5pt}
\changed{\subsection{Six-DOF Spacecraft Simulator}
We apply the gPC-SCP method and the gPC-SCP$^\mathrm{PC}$ method to a Six-DOF spacecraft simulator model~\cite{nakka2018six} and demonstrate the scalability of the methods to higher dimensions for planning a safe trajectory. The dynamics of the spacecraft simulator in states $\vecx = [\vecp,\vecv,\vecq,\vecw] \in \real^{13}$ is given as $\dot{\vecp}=\vecv$, $m\dot{\vecv} = \bar{\vecu} + \sigma_0 \xi_0 \bar{\vecu}$, $\dot{\vecq} = \frac{1}{2}\boldsymbol{\Omega}(\vecw)$ and $J \dot{\vecw} = -S(\vecw)J\vecw + \bar{\vect} + \Xi (-S(\vecw)\vecw + \bar{\vect}) $, 
where $m$ is the mass, $J =\mathrm{diag}([J_1,J_2,J_3])$ is the moment of inertial, $\bar{\vecu} \in \real^3$ and $\bar{\vect} \in \real^3$ are force and torque input. We refer the readers to~\cite{nakka2018six} for the definitions of $S(.)$ and $\boldsymbol{\Omega}(.)$, and $\Xi = \textrm{diag}([\sigma_1\xi_1,\sigma_2\xi_2,\sigma_3\xi_3])$. The Gaussian variables $\xi_0$ and $\Xi$ are due to uncertainty in $m$ and $J$, respectively. The uncertainty is a function of both state $\vecw$ and control $[\bar{\vecu},\bar{\vecw}]$. The mass $m=$\SI{10}{kg}, moment of inertial $J_1=J_2=$\SI{0.07}{kg m^2}, $J_3=$\SI{0.1}{kg m^2}. We have $\sigma_0=0.03$, $\sigma_1=0.002$, $\sigma_1=0.002$, and $\sigma_3=0.002$. The control limits are $[\bar{\vecu}_{\max},\bar{\vect}_{\max}] = [1,1,1,0.3,0.3,0.3])$, $[\bar{\vecu}_{\min},\bar{\vect}_{\min}]=-[\bar{\vecu}_{\max},\bar{\vect}_{\max}]$.} 

\changed{We design safe motion plan for the Six-DOF spacecraft in an obstacle rich environment with initial and terminal states as $\Exp(\vecx_0)= [-1,-1,-1,0,0,0,0,0,0,1,0,0,0]$ and $\Exp(\vecx_f)=[1.2,1.2,1.0,0,0,0,-0.5,0.5,-0.5,0.5,0,0,0]$, respectively. We minimize the control cost $\|[\bar{\vecu},\bar{\vect}]\|_2$ and the terminal variance of the trajectory $Q_f = \identity$. The risk measure for collision avoidance is $\epsilon_{\col} =0.05$. We show the safe motion plans in~\Cref{fig:6DOF_free_flyer_simulation}. Note that the gPC-SCP and gPC-SCP$^\mathrm{PC}$ compute equivalent mean trajectories. The computation time for gPC-SCP and gPC-SCP$^\mathrm{PC}$ was \SI{40}{s} and \SI{5}{s}, respectively. We execute the designed motion plans using an exponentially-stabilizing feedback controller~\cite{nakka2018six}. Over 1000 Monte Carlo trials of feedback execution, we had 31 constraint violations for our method, while for Gaussian confidence-based collision constraint~\cref{eq:gaussian_linear_constraint}, it was 57.}

\section{Conclusion}\label{sec:conclusion}
We present a generalized polynomial chaos-based sequential convex programming method for safe and optimal motion planning and control under uncertainty in dynamics and constraints. The method uses generalized polynomial chaos projection and distributional robustness to compute a convex subset of the multimodel state-dependent chance constraints and a high-fidelity deterministic surrogate of the stochastic dynamics and the cost functional. \changed{We prove the asymptotic convergence of the surrogate problem to the stochastic optimal control problem. The asymptotic convergence property of the deterministic surrogate and distributional robustness allows for achieving a greater degree of safety. Furthermore, we derive a predictor-corrector formulation to improve the computation speed by an order of magnitude without compromising the safety and optimality of the motion trajectories.}

\changed{We validate our approach in simulations and on the robotic spacecraft simulator hardware and demonstrate a higher success rate in ensuring the safety of motion trajectories compared to a Gaussian approximation of the chance constraints. Our approach outperforms Gaussian approximation in computing safe motion plans under stochastic uncertainty in dynamics and obstacles by at least 20 percent over 1000 trials of feedback execution.} 
\section*{Acknowledgement}
The authors thank Amir Rahmani, Fred Y. Hadaegh, Joel Burdick, Richard Murray, and Yisong Yue for stimulating discussions and technical help.

\bibliographystyle{IEEEtran}
\bibliography{gpc_scp.bib}

\begin{thebibliography}{10}
\providecommand{\url}[1]{#1}
\csname url@samestyle\endcsname
\providecommand{\newblock}{\relax}
\providecommand{\bibinfo}[2]{#2}
\providecommand{\BIBentrySTDinterwordspacing}{\spaceskip=0pt\relax}
\providecommand{\BIBentryALTinterwordstretchfactor}{4}
\providecommand{\BIBentryALTinterwordspacing}{\spaceskip=\fontdimen2\font plus
\BIBentryALTinterwordstretchfactor\fontdimen3\font minus
  \fontdimen4\font\relax}
\providecommand{\BIBforeignlanguage}[2]{{%
\expandafter\ifx\csname l@#1\endcsname\relax
\typeout{** WARNING: IEEEtran.bst: No hyphenation pattern has been}%
\typeout{** loaded for the language `#1'. Using the pattern for}%
\typeout{** the default language instead.}%
\else
\language=\csname l@#1\endcsname
\fi
#2}}
\providecommand{\BIBdecl}{\relax}
\BIBdecl

\bibitem{nakka2019nsoc}
Y.~K. {Nakka} and S.~{Chung}, ``Trajectory optimization for chance-constrained
  nonlinear stochastic systems,'' in \emph{IEEE Conf. on Decis. and Control},
  2019, pp. 3811--3818.

\bibitem{blackmore2010probabilistic}
L.~Blackmore, M.~Ono, A.~Bektassov, and B.~C. Williams, ``A probabilistic
  particle-control approximation of chance-constrained stochastic predictive
  control,'' \emph{IEEE Trans. Robot.}, vol.~26, no.~3, pp. 502--517, 2010.

\bibitem{blackmore2011chance}
L.~Blackmore, M.~Ono, and B.~C. Williams, ``Chance-constrained optimal path
  planning with obstacles,'' \emph{IEEE Trans. Robot.}, vol.~27, no.~6, pp.
  1080--1094, 2011.

\bibitem{toit2012robot}
N.~E. {Du Toit} and J.~W. {Burdick}, ``Robot motion planning in dynamic,
  uncertain environments,'' \emph{IEEE Trans. Robot.}, vol.~28, no.~1, pp.
  101--115, 2012.

\bibitem{tsukamoto2020}
H.~{Tsukamoto} and S.~J. {Chung}, ``Robust controller design for stochastic
  nonlinear systems via convex optimization,'' \emph{IEEE Trans. Autom.
  Control}, pp. 1--1, 2020.

\bibitem{zhu2019chance}
H.~Zhu and J.~Alonso-Mora, ``Chance-constrained collision avoidance for mavs in
  dynamic environments,'' \emph{IEEE Trans. Robot. Autom. Lett.}, vol.~4,
  no.~2, pp. 776--783, 2019.

\bibitem{kaelbling2013integrated}
L.~P. Kaelbling and T.~Lozano-P{\'e}rez, ``Integrated task and motion planning
  in belief space,'' \emph{Int.J. Robot. Res.}, vol.~32, no. 9-10, pp.
  1194--1227, 2013.

\bibitem{nakka2021information}
Y.~K. Nakka, W.~H{\"o}nig, C.~Choi, A.~Harvard, A.~Rahmani, and S.-J. Chung,
  ``Information-based guidance and control architecture for multi-spacecraft
  on-orbit inspection,'' in \emph{AIAA GNC Conf.}, 2021, p. 1103.

\bibitem{nakka2020chance}
Y.~K. {Nakka}, A.~{Liu}, G.~{Shi}, A.~{Anandkumar}, Y.~{Yue}, and S.~J.
  {Chung}, ``Chance-constrained trajectory optimization for safe exploration
  and learning of nonlinear systems,'' \emph{IEEE Trans. Robot. Autom. Lett.},
  vol.~6, no.~2, pp. 389--396, 2021.

\bibitem{wabersich2021}
K.~P. {Wabersich}, L.~{Hewing}, A.~{Carron}, and M.~N. {Zeilinger},
  ``Probabilistic model predictive safety certification for learning-based
  control,'' \emph{IEEE Trans. Autom. Control}, pp. 1--1, 2021.

\bibitem{Nakka2021SpacecraftLearning}
\BIBentryALTinterwordspacing
Y.~K. Nakka, ``{Spacecraft Motion Planning and Control under Probabilistic
  Uncertainty for Coordinated Inspection and Safe Learning},'' Ph.D.
  dissertation, California Institute of Technology, Pasadena, 5 2021. [Online].
  Available:
  \url{https://resolver.caltech.edu/CaltechTHESIS:05142021-163257155}
\BIBentrySTDinterwordspacing

\bibitem{nakka2018six}
Y.~K. Nakka, R.~C. Foust, E.~S. Lupu, D.~B. Elliott, I.~S. Crowell, S.-J.
  Chung, and F.~Y. Hadaegh, ``Six degree-of-freedom spacecraft dynamics
  simulator for formation control research,'' in \emph{AAS/AIAA Astrodynamics
  Specialist Conference}, 2018.

\bibitem{ridderhof2018uncertainty}
J.~Ridderhof and P.~Tsiotras, ``Uncertainty quantication and control during
  mars powered descent and landing using covariance steering,'' in \emph{AIAA
  GNC Conf.}, 2018, p. 0611.

\bibitem{shi2018neural}
G.~Shi, X.~Shi, M.~O'Connell, R.~Yu, K.~Azizzadenesheli, A.~Anandkumar, Y.~Yue,
  and S.-J. Chung, ``Neural lander: Stable drone landing control using learned
  dynamics,'' in \emph{Proc. IEEE Int. Conf. Robot. Autom.}, 2019.

\bibitem{shi2020neural}
G.~Shi, W.~H{\"o}nig, X.~Shi, Y.~Yue, and S.-J. Chung, ``Neural-swarm2:
  Planning and control of heterogeneous multirotor swarms using learned
  interactions,'' \emph{IEEE Trans. Robot.}, 2021.

\bibitem{xiu2002wiener}
D.~Xiu and G.~E. Karniadakis, ``The wiener--askey polynomial chaos for
  stochastic differential equations,'' \emph{SIAM J. Sci. Comp.}, vol.~24,
  no.~2, pp. 619--644, 2002.

\bibitem{xiu2009fast}
D.~Xiu, ``Fast numerical methods for stochastic computations: a review,''
  \emph{Com. in Comp. physics}, vol.~5, no. 2-4, pp. 242--272, 2009.

\bibitem{ghanem2003stochastic}
R.~G. Ghanem and P.~D. Spanos, \emph{Stochastic finite elements: a spectral
  approach}.\hskip 1em plus 0.5em minus 0.4em\relax Courier Corporation, 2003.

\bibitem{nemirovski2007convex}
A.~Nemirovski and A.~Shapiro, ``Convex approximations of chance constrained
  programs,'' \emph{SIAM Journal on Optimization}, vol.~17, no.~4, pp.
  969--996, 2007.

\bibitem{calafiore2006distributionally}
G.~C. Calafiore and L.~El~Ghaoui, ``On distributionally robust
  chance-constrained linear programs,'' \emph{Journal of Optimization Theory
  and Applications}, vol. 130, no.~1, pp. 1--22, 2006.

\bibitem{zymler2013distributionally}
S.~Zymler, D.~Kuhn, and B.~Rustem, ``Distributionally robust joint chance
  constraints with second-order moment information,'' \emph{Mathematical
  Programming}, vol. 137, no. 1-2, pp. 167--198, 2013.

\bibitem{morgan2014model}
D.~Morgan, S.-J. Chung, and F.~Y. Hadaegh, ``Model predictive control of swarms
  of spacecraft using sequential convex programming,'' \emph{Journal of
  Guidance, Control, and Dynamics}, vol.~37, no.~6, pp. 1725--1740, 2014.

\bibitem{morgan2012spacecraft}
D.~Morgan, S.-J. Chung, and F.~Hadaegh, ``Spacecraft swarm guidance using a
  sequence of decentralized convex optimizations,'' in \emph{AIAA/AAS Astro.
  Spec. Conf.}, 2012, p. 4583.

\bibitem{morgan2016swarm}
D.~Morgan, G.~P. Subramanian, S.-J. Chung, and F.~Y. Hadaegh, ``Swarm
  assignment and trajectory optimization using variable-swarm, distributed
  auction assignment and sequential convex programming,'' \emph{Int. J. Robot.
  Research}, vol.~35, no.~10, pp. 1261--1285, 2016.

\bibitem{kushner1967stochastic}
H.~J. Kushner, ``Stochastic stability and control,'' Brown Univ Providence RI,
  Tech. Rep., 1967.

\bibitem{khasminskii2011stochastic}
R.~Khasminskii, \emph{Stochastic stability of differential equations}.\hskip
  1em plus 0.5em minus 0.4em\relax Springer Science \& Business Media, 2011,
  vol.~66.

\bibitem{hauser2016}
K.~{Hauser} and Y.~{Zhou}, ``Asymptotically optimal planning by feasible
  kinodynamic planning in a state–cost space,'' \emph{IEEE Trans. Robot.},
  vol.~32, no.~6, pp. 1431--1443, 2016.

\bibitem{lavalle_2006}
S.~M. LaValle, \emph{Planning Algorithms}.\hskip 1em plus 0.5em minus
  0.4em\relax Cambridge University Press, 2006.

\bibitem{du2011probabilistic}
N.~E. Du~Toit and J.~W. Burdick, ``Probabilistic collision checking with chance
  constraints,'' \emph{IEEE Trans. Robot.}, vol.~27, no.~4, pp. 809--815, 2011.

\bibitem{todorov2005generalized}
E.~Todorov and W.~Li, ``A generalized iterative lqg method for locally-optimal
  feedback control of constrained nonlinear stochastic systems,'' in
  \emph{Proc. of American Control Conference}, 2005, pp. 300--306.

\bibitem{van2012motion}
J.~Van Den~Berg, S.~Patil, and R.~Alterovitz, ``Motion planning under
  uncertainty using iterative local optimization in belief space,''
  \emph{Int.J. Robot. Res.}, vol.~31, no.~11, pp. 1263--1278, 2012.

\bibitem{ridderhof2019nonlinear}
J.~Ridderhof, K.~Okamoto, and P.~Tsiotras, ``Nonlinear uncertainty control with
  iterative covariance steering,'' in \emph{IEEE 58th Conference on Decision
  and Control}, 2019, pp. 3484--3490.

\bibitem{calafiore2013stochastic}
G.~C. Calafiore and L.~Fagiano, ``Stochastic model predictive control of lpv
  systems via scenario optimization,'' \emph{Automatica}, vol.~49, no.~6, pp.
  1861--1866, 2013.

\bibitem{janson2018monte}
L.~Janson, E.~Schmerling, and M.~Pavone, ``Monte carlo motion planning for
  robot trajectory optimization under uncertainty,'' in \emph{Robotics
  Research}.\hskip 1em plus 0.5em minus 0.4em\relax Springer, 2018, pp.
  343--361.

\bibitem{calafiore2013}
G.~C. {Calafiore} and L.~{Fagiano}, ``Robust model predictive control via
  scenario optimization,'' \emph{IEEE Trans. Autom. Control}, vol.~58, no.~1,
  pp. 219--224, 2013.

\bibitem{arnold1974stochastic}
L.~Arnold, \emph{Stochastic differential equations}.\hskip 1em plus 0.5em minus
  0.4em\relax John Wiley \& Sons, 1974.

\bibitem{castillo2020real}
M.~Castillo-Lopez, P.~Ludivig, S.~A. Sajadi-Alamdari, J.~L. Sanchez-Lopez,
  M.~A. Olivares-Mendez, and H.~Voos, ``A real-time approach for
  chance-constrained motion planning with dynamic obstacles,'' \emph{IEEE
  Trans. Robot. Autom. Lett.}, vol.~5, no.~2, pp. 3620--3625, 2020.

\bibitem{vandenberghe2007generalized}
L.~Vandenberghe, S.~Boyd, and K.~Comanor, ``Generalized chebyshev bounds via
  semidefinite programming,'' \emph{SIAM review}, vol.~49, no.~1, pp. 52--64,
  2007.

\bibitem{tlew2020}
T.~{Lew}, R.~{Bonalli}, and M.~{Pavone}, ``Chance-constrained sequential convex
  programming for robust trajectory optimization,'' in \emph{Proc. Eu. Control
  Conf.}, 2020, pp. 1871--1878.

\bibitem{mesbah2016stochastic}
A.~Mesbah, ``Stochastic model predictive control: An overview and perspectives
  for future research,'' \emph{IEEE Cont. Sys. Mag.}, vol.~36, no.~6, pp.
  30--44, 2016.

\bibitem{mesbah2014stochastic}
A.~Mesbah, S.~Streif, R.~Findeisen, and R.~D. Braatz, ``Stochastic nonlinear
  model predictive control with probabilistic constraints,'' in \emph{IEEE
  American control conference}, 2014, pp. 2413--2419.

\bibitem{bavdekar2016stochastic}
V.~A. Bavdekar and A.~Mesbah, ``Stochastic nonlinear model predictive control
  with joint chance constraints,'' \emph{IFAC-PapersOnLine}, vol.~49, no.~18,
  pp. 270--275, 2016.

\bibitem{hover2006application}
F.~S. Hover and M.~S. Triantafyllou, ``Application of polynomial chaos in
  stability and control,'' \emph{Automatica}, vol.~42, no.~5, pp. 789--795,
  2006.

\bibitem{fisher2008stability}
J.~Fisher and R.~Bhattacharya, ``Stability analysis of stochastic systems using
  polynomial chaos,'' in \emph{Proc. American Control Conference}, 2008, pp.
  4250--4255.

\bibitem{kim2013wiener}
K.-K. Kim, D.~E. Shen, Z.~K. Nagy, and R.~D. Braatz, ``Wiener's polynomial
  chaos for the analysis and control of nonlinear dynamical systems with
  probabilistic uncertainties [historical perspectives],'' \emph{IEEE Control
  Systems Magazine}, vol.~33, no.~5, pp. 58--67, 2013.

\bibitem{boutselis2017stochastic}
G.~I. Boutselis, Y.~Pan, G.~De~La~Tore, and E.~A. Theodorou, ``Stochastic
  trajectory optimization for mechanical systems with parametric
  uncertainties,'' \emph{arXiv preprint arXiv:1705.05506}, 2017.

\bibitem{fisher2011optimal}
J.~Fisher and R.~Bhattacharya, ``Optimal trajectory generation with
  probabilistic system uncertainty using polynomial chaos,'' \emph{Journal of
  Dynamic Systems, Measurement, and Control}, vol. 133, no.~1, p. 014501, 2011.

\bibitem{buehler2017efficient}
E.~Buehler, ``Efficient uncertainty propagation for stochastic model predictive
  control,'' Ph.D. dissertation, UC Berkeley, 2017.

\bibitem{foust2020optimal}
R.~Foust, S.-J. Chung, and F.~Y. Hadaegh, ``Optimal guidance and control with
  nonlinear dynamics using sequential convex programming,'' \emph{Journal of
  Guidance, Control, and Dynamics}, vol.~43, no.~4, pp. 633--644, 2020.

\bibitem{cheng2021limits}
R.~Cheng, R.~M. Murray, and J.~W. Burdick, ``Limits of probabilistic safety
  guarantees when considering human uncertainty,'' \emph{arXiv preprint
  arXiv:2103.03388}, 2021.

\bibitem{chen2007new}
X.~Chen, ``A new generalization of chebyshev inequality for random vectors,''
  \emph{arXiv preprint arXiv:0707.0805}, 2007.

\bibitem{hiro2008}
M.~{Ono} and B.~C. {Williams}, ``Iterative risk allocation: A new approach to
  robust model predictive control with a joint chance constraint,'' in
  \emph{2008 47th IEEE Conf. on Decis. and Control}, Dec 2008, pp. 3427--3432.

\bibitem{cameron1947orthogonal}
R.~H. Cameron and W.~T. Martin, ``The orthogonal development of non-linear
  functionals in series of fourier-hermite functionals,'' \emph{Annals of
  Math.}, pp. 385--392, 1947.

\bibitem{muhlpfordt2017comments}
T.~M{\"u}hlpfordt, R.~Findeisen, V.~Hagenmeyer, and T.~Faulwasser, ``Comments
  on truncation errors for polynomial chaos expansions,'' \emph{IEEE Cont. Sys.
  Let.}, vol.~2, no.~1, pp. 169--174, 2017.

\bibitem{blatman2011adaptive}
G.~Blatman and B.~Sudret, ``Adaptive sparse polynomial chaos expansion based on
  least angle regression,'' \emph{J. of Comp. Physics}, vol. 230, no.~6, pp.
  2345--2367, 2011.

\bibitem{oladyshkin2012data}
S.~Oladyshkin and W.~Nowak, ``Data-driven uncertainty quantification using the
  arbitrary polynomial chaos expansion,'' \emph{Reliability Engineering \&
  System Safety}, vol. 106, pp. 179--190, 2012.

\bibitem{xu2018novel}
Y.~Xu, L.~Mili, and J.~Zhao, ``A novel polynomial-chaos-based kalman filter,''
  \emph{IEEE Signal Processing Letters}, vol.~26, no.~1, pp. 9--13, 2018.

\bibitem{platen2010numerical}
E.~Platen and N.~Bruti-Liberati, \emph{Numerical solution of stochastic
  differential equations with jumps in finance}.\hskip 1em plus 0.5em minus
  0.4em\relax Springer Science \& Business Media, 2010, vol.~64.

\bibitem{hewing2018stochastic}
L.~Hewing and M.~N. Zeilinger, ``Stochastic model predictive control for linear
  systems using probabilistic reachable sets,'' in \emph{IEEE Conf. on Dec. and
  Control}, 2018, pp. 5182--5188.

\bibitem{li2016asymptotically}
Y.~Li, Z.~Littlefield, and K.~E. Bekris, ``Asymptotically optimal
  sampling-based kinodynamic planning,'' \emph{Int.J. Robot. Res.}, vol.~35,
  no.~5, pp. 528--564, 2016.

\bibitem{karaman2011sampling}
S.~Karaman and E.~Frazzoli, ``Sampling-based algorithms for optimal motion
  planning,'' \emph{Int.J. Robot. Res.}, vol.~30, no.~7, pp. 846--894, 2011.

\bibitem{diamond2016cvxpy}
S.~Diamond and S.~Boyd, ``Cvxpy: A python-embedded modeling language for convex
  optimization,'' \emph{J. of Mach. Learn. Res.}, vol.~17, no.~1, pp.
  2909--2913, 2016.

\bibitem{domahidi2013ecos}
A.~Domahidi, E.~Chu, and S.~Boyd, ``Ecos: An socp solver for embedded
  systems,'' in \emph{IEEE Euro. Control Conf.}, 2013, pp. 3071--3076.

\end{thebibliography}
\begin{IEEEbiography}[{\includegraphics[width=1in,height=1.25in,clip,keepaspectratio]{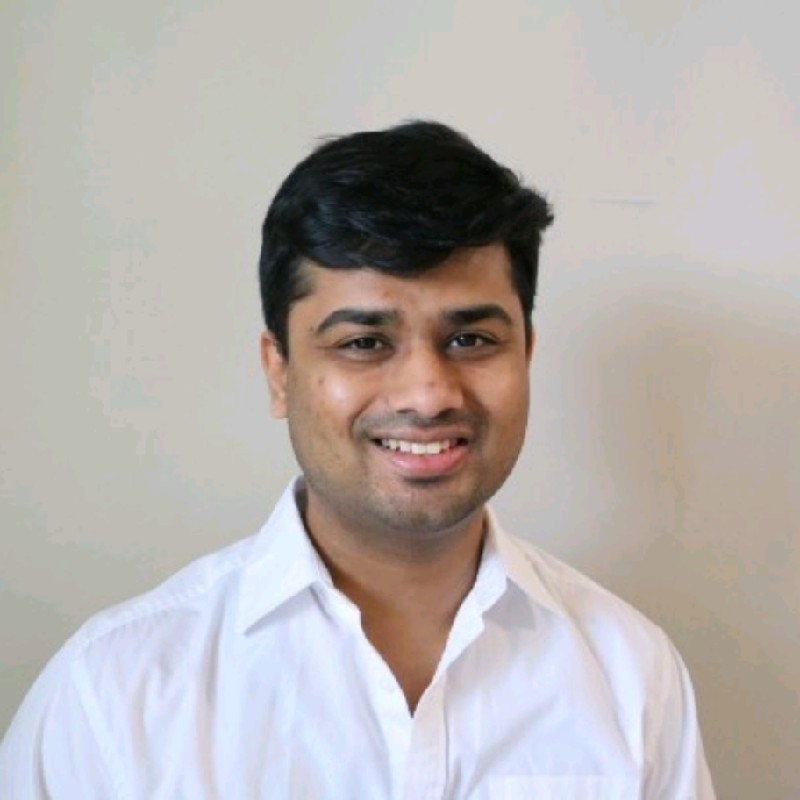}}]{Yashwanth Kumar Nakka} received the B. Tech. in aerospace engineering from the Indian Institute of Space Science and Technology, India, in 2011, and the M. Sc. degree in aerospace engineering from University of Illinois Urbana-Champaign, IL, USA, in 2016 and the M. Sc. degree in space engineering from California Institute of Technology, CA, USA, in 2017. He is currently a PhD. candidate in the department of aerospace, California Institute of Technology. He was an engineer for the GSAT-15 and 16 missions at the Indian Space Research Organization during 2011-2014. His research interests include spacecraft autonomy, motion planning and control under uncertainty, and nonlinear dynamics and control. 

He received the best student paper award at the 2021 American Institute of Aeronautics and Astronautics Guidance, Navigation, and Controls conference and the best paper award at the $11^{\mathrm{th}}$ International Workshop on Satellite Constellations and Formation Flying.
\end{IEEEbiography}
 \vskip -2\baselineskip plus -1fil 
\begin{IEEEbiography}[{\includegraphics[width=1in,height=1.25in,clip,keepaspectratio]{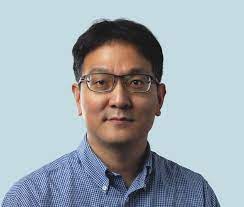}}]{Soon-Jo Chung}(M’06–SM’12) received the B.S.degree (summa cum laude) in aerospace engineering from the Korea Advanced Institute of Science and Technology,  Daejeon, South Korea, in 1998, and the S.M. degree in aeronautics and astronautics and the Sc.D. degree in estimation and control from Massachusetts Institute of Technology, Cambridge, MA, USA, in 2002 and 2007, respectively. He is currently the Bren Professor of Aerospace and Control and Dynamical Systems and a Jet Propulsion Laboratory Research Scientist in the California Institute of Technology, Pasadena,CA, USA. He was with the faculty of the University of Illinois at Urbana-Champaign during 2009–2016. His research interests include spacecraft  and aerial swarms and autonomous aerospace systems, and in particular, on the theory and application of complex nonlinear dynamics, control, estimation, guidance, and navigation of autonomous space and air vehicles.

Dr. Chung was the recipient of the UIUC Engineering Deans Award for Excellence in Research, the Beckman Faculty Fellowship of the UIUC Center for Advanced Study, the U.S. Air Force Office of Scientific Research Young Investigator Award, the National Science Foundation Faculty Early Career Development Award, an Honorable Mention for the 2020 IEEE R-AL Best Paper Award, and three Best Conference Paper Awards from the IEEE and the American Institute of Aeronautics and Astronautics. He is an Associate Editor of IEEE TRANSACTIONS ON AUTOMATIC CONTROL and Journal of Guidance, Control, and Dynamics. He was an Associate Editor of IEEE TRANSACTIONS ON ROBOTICS.
\end{IEEEbiography}

\end{document}